%% file: main.tex
\documentclass[journal]{IEEEtran}
\IEEEoverridecommandlockouts 

\input{custom/packages}

\input{custom/shortcuts}

\input{custom/acronyms}

\input{custom/commands}

\input{custom/during_writing}

\title{
Holistic Fusion: Task- and Setup-Agnostic Robot Localization and State Estimation with Factor Graphs
}

\newif\ifanonymous
\anonymousfalse 
\newcount\revisionmode
\DeclareRobustCommand{\anontext}[2]{%
  \ifanonymous
    \textcolor{orange}{[#2]}%
  \else
    #1%
  \fi
}

\newcommand{\replace}[2]{%
  \ifcase\revisionmode
    #2%
  \or
    {\textcolor{red}{#1}}{\textcolor{ForestGreen}{#2}}%
  \or
    {\textcolor{ForestGreen}{#2}}%
  \or
    {\textcolor{red}{#1}}%
  \fi
}

\newcommand{\replacebox}[1]{%
  \ifcase\revisionmode
    #1%
  \or
    {\setlength{\fboxrule}{1.5pt}\setlength{\fboxsep}{4pt}%
    \fcolorbox{ForestGreen}{white}{%
      \begin{minipage}{\dimexpr\columnwidth-2\fboxsep-2\fboxrule}#1\end{minipage}}}%
  \or
    {\setlength{\fboxrule}{1.5pt}\setlength{\fboxsep}{4pt}%
    \fcolorbox{ForestGreen}{white}{%
      \begin{minipage}{\dimexpr\columnwidth-2\fboxsep-2\fboxrule}#1\end{minipage}}}%
  \or
    #1%
  \fi
}

\mdfdefinestyle{replaceframestyle}{linecolor=ForestGreen,linewidth=1.5pt,innertopmargin=4pt,innerbottommargin=4pt,innerleftmargin=4pt,innerrightmargin=4pt}
\newenvironment{replaceframe}{%
  \ifcase\revisionmode\else\ifnum\revisionmode=3\else\begin{mdframed}[style=replaceframestyle]\fi\fi
}{%
  \ifcase\revisionmode\else\ifnum\revisionmode=3\else\end{mdframed}\fi\fi
}

\author{
\anontext{
Julian Nubert$^{\dagger,\ddagger,\S}$\orcidlink{0000-0001-8949-6134},
\thanks{*This work is supported in part by the Sony Research Grant 2023, the EU Horizon 2020 programme grant agreements No.\ 852044, 101016970, and 101070405, EU Horizon 2021 programme grant agreement No.\ 101070596, the ETH Zurich Research Grant No.\ 21-1 ETH-27, the Swiss National Science Foundation
(SNSF) through project No.\ 227617, the NCCR Digital Fabrication, and the Max Planck ETH Center for Learning Systems. \ This research was partially conducted at the Jet Propulsion Laboratory, California Institute of Technology, under a contract with the National Aeronautics and Space Administration (80NM0018D0004). \ This work was partially supported by Defense Advanced Research Projects Agency (DARPA).}
\thanks{$^{\dagger}$The authors are with the Robotic Systems Lab, ETH Z\"urich, Switzerland.}%
\thanks{$^{\ddagger}$The authors are with the Max Planck Institute for Intelligent Systems, Stuttgart, Germany.}%
\thanks{$^{\S}$The authors are with the NASA Jet Propulsion Laboratory, California Institute of Technology, USA.}%
\thanks{Corresponding Author: Julian Nubert, {\href{mailto:nubertj@gmail.com}{\textcolor{blue}{nubertj@gmail.com}}}}%
Turcan Tuna$^{\dagger}$\orcidlink{0000-0001-8662-4890},
Jonas Frey$^{\dagger,\S}$\orcidlink{0000-0002-7401-2173},\\
Cesar Cadena$^{\dagger}$\orcidlink{0000-0002-2972-6011},
Katherine J. Kuchenbecker$^{\ddagger}$\orcidlink{0000-0002-5004-0313},
Shehryar Khattak$^{\S}$\orcidlink{0000-0002-9304-1455},
Marco Hutter$^{\dagger}$\orcidlink{0000-0002-4285-4990}
}{anonymous authors}
}

\usepackage{fancyhdr}
\newcommand{\mytitle}{\textbf{Accepted version.} To appear in \textit{IEEE Transactions on Robotics, 2026.}.  DOI:
10.1109/TRO.2026.3714645\\
\copyright 2026 IEEE. Personal use of this material is permitted.
Permission from IEEE must be obtained for all other uses, in any current or future media, including reprinting/republishing this material for advertising or promotional purposes, creating new collective works, for resale or redistribution to servers or lists, or reuse of any copyrighted component of this work in other works.}
\begin{document}

\maketitle
\thispagestyle{fancy}
\urlstyle{tt} 

\input{sections/0-Abstract}

\section{Introduction}
\label{sec:intro}
\input{sections/1-Introduction}

\section{Related Work}
\label{sec:related}
\input{sections/2-RelatedWork}

\section{Problem Formulation \& Structure}
\label{sec:problem_formulation}
\input{sections/3-ProblemFormulation}

\section{Methodology}
\label{sec:method}
\input{sections/4-Methodology}

\section{Implementation Details \& Framework Overview}
\label{sec:implementation}
\input{sections/5-ImplementationalDetails}

\section{Experimental Results}
\label{sec:experimental_results}
\input{sections/6-ExperimentalResults}

\section{Discussion}
\label{sec:discussion}
\input{sections/7-Discussion}

\section{Conclusions \& Future Work}
\label{sec:conclusions}
\input{sections/8-Conclusions}

\ifthenelse{\boolean{anonymous}}%
{} 
{ 
\section*{Acknowledgments}
The authors thank their colleagues at ETH Z\"urich and NASA JPL for help conducting the robot experiments and evaluations and for using \ac{GMSF} and \ac{HF} on their robots. 
Special thanks go to Takahiro Miki and the ANYmal Hike team at \ac{RSL}; Nikita Rudin and David Hoeller for the ANYmal Parkour experiments; Patrick Spieler for running the deployments on the JPL RACER vehicle; the entire \ac{HEAP} team at \ac{RSL} and \textit{Gravis Robotics}, Thomas Mantel and the teaching assistants of the ETH Robotic Summer School for their help on the \acp{SMB}; and Mayank Mittal for help generating the renderings.
}

\begingroup
\raggedright
\tiny
\bibliographystyle{IEEEtran}
\bibliography{bibliography}
\endgroup

\ifthenelse{\boolean{anonymous}}%
{} 
{ 
\input{biographies}
}

\end{document}

%% file: custom/packages.tex
\usepackage{amsmath}
\usepackage{amssymb}
\usepackage[nospace]{cite}
\usepackage[dvipsnames,svgnames]{xcolor}
\usepackage{fancyhdr}

\usepackage{adjustbox}
\usepackage{graphicx}
\usepackage{tensor}
\usepackage[utf8]{inputenc}
\usepackage[OT1]{fontenc}
\usepackage[final]{microtype}
\usepackage{booktabs}
\usepackage{siunitx}
\usepackage{acronym}
\usepackage{todonotes}
\usepackage{colortbl}
\usepackage{multirow}
\usepackage{seqsplit}
\usepackage{comment}
\usepackage[bottom]{footmisc}
\usepackage[ruled]{algorithm2e} 
\usepackage[noend]{algpseudocode}
\usepackage{tabularx}
\usepackage{arydshln}
\usepackage{makecell}
\usepackage{ifthen}
\usepackage{tikz}
\usetikzlibrary{arrows.meta,positioning,calc}
\usepackage[framemethod=tikz]{mdframed}

\makeatletter
\let\NAT@parse\undefined
\makeatother
\usepackage[colorlinks=true, citecolor=blue, linkcolor=blue, urlcolor=red]{hyperref}
\usepackage{cleveref}
\usepackage{orcidlink}

%% file: custom/shortcuts.tex
\DeclareMathOperator*{\argmax}{arg\,max}

\newcommand{\SO}{\mathrm{SO}}
\newcommand{\SE}{\mathrm{SE}}
\newcommand{\SOthree}{\SO(3)}
\newcommand{\SEthree}{\SE(3)}
\newcommand{\sethree}{\mathfrak{se}(3)}
\newcommand{\sethreetangent}{\mathbf{\gamma}}

\newcommand{\Real}{\mathbb{R}}

\newcommand{\R}{\mathbf{R}}

\newcommand{\T}{\mathbf{T}}
\newcommand{\transl}{\mathbf{t}}

\newcommand{\residual}{\mathbf{r}}

\newcommand{\rotvel}{\boldsymbol\omega}
\newcommand{\accel}{\mathbf{a}}

\newcommand{\tran}{\mathbf{p}}
\newcommand{\vel}{\mathbf{v}}
\newcommand{\bias}{\mathbf{b}}

\newcommand{\states}{\mathcal{X}}
\newcommand{\state}{\boldsymbol{\mathbf{x}}}
\newcommand{\ContextState}{\mathcal{\theta}}
\newcommand{\Measurements}{\mathcal{Z}}
\newcommand{\Measurement}{\mathbf{z}}
\newcommand{\Measfunction}{\mathbf{h}}

\newcommand{\World}{\mathtt{W}}
\newcommand{\Odom}{\mathtt{O}}
\newcommand{\Imu}{\mathtt{I}}
\newcommand{\Base}{\mathtt{{B}}}
\newcommand{\Reference}{\mathtt{R}}
\newcommand{\Keyframe}{\mathtt{K}}
\newcommand{\Sensor}{\mathtt{S}}
\newcommand{\Map}{\mathtt{M}}
\newcommand{\ENU}{\World_{\text{ENU}}}
\newcommand{\OpenThree}{\Map_{\text{O3D}}}
\newcommand{\Compslam}{\Map_{\text{comp.}}}
\newcommand{\Coinlio}{\Map_{\text{coin.}}}
\newcommand{\Liopard}{\Map_{\text{LIO}}}

\newcommand{\Lidar}{\mathtt{L}}
\newcommand{\Velodyne}{\Lidar_{\text{V}}}
\newcommand{\Ouster}{\Lidar_{\text{O}}}
\newcommand{\Gnss}{\mathtt{G}}
\newcommand{\Leg}{\mathtt{K}}
\newcommand{\Radar}{\mathtt{R}_{\text{a}}}

\newcommand{\WheelAxis}{\mathtt{A}}

\newcommand{\GnssLeft}{\mathtt{G_{\text{L}}}}
\newcommand{\GnssRight}{\mathtt{G_{\text{R}}}}

\newcommand{\Feature}{\mathtt{F}}

\newcommand{\Imumeas}{\mathtt{i}}
\newcommand{\Absolute}{\mathtt{a}}
\newcommand{\Local}{\mathtt{r}}
\newcommand{\Landmark}{\mathtt{f}}

\newcommand{\ba} {\bias^a} 
\newcommand{\bg} {\bias^g} 

\newcommand{\Marginal}{\mathcal{M}}
\newcommand{\adj}{\text{Ad}}



%% file: custom/acronyms.tex
\acrodef{HF}{Holistic Fusion}
\acrodef{GMSF}{Graph MSF}
\acrodef{FG}{Factor graph}
\acrodef{ISAM}{Incremental Smoothing and Mapping}
\acrodef{LM}{Levenberg-Marquardt}
\acrodef{GTSAM}{Georgia Tech Smoothing and Mapping}
\acrodef{FL}{fixed-lag}
\acrodef{LS}{least squares}
\acrodef{GP}{Gaussian process}
\acrodef{BO}{Batch Optimization}
\acrodef{MC}{Markov chain}
\acrodef{KF}{Kalman filter}
\acrodef{EKF}{extended Kalman filter}
\acrodef{UKF}{unscented Kalman filter}
\acrodef{MHE}{moving horizon estimation}
\acrodef{IMU}{inertial measurement unit}
\acrodef{LiDAR}{light detection and ranging}
\acrodef{GNSS}{global navigation satellite system}
\acrodef{RTK}{real-time kinematic}
\acrodef{mocap}{motion capture}
\acrodef{UWB}{ultra-wideband}
\acrodef{RADAR}{radio detection and ranging}
\acrodef{SLAM}{simultaneous localization and mapping}
\acrodef{VIO}{visual inertial odometry}
\acrodef{LIO}{LiDAR Inertial Odometry}
\acrodef{LO}{LiDAR odometry}
\acrodef{LR}{LiDAR registration}
\acrodef{WIO}{Wheel Inertial Odometry}
\acrodef{WO}{Wheel Odometry}
\acrodef{MAP}{maximum a posteriori}
\acrodef{VLM}{vision language model}
\acrodef{MSF}{Multi Sensor Fusion}
\acrodef{TSIF}{two-state implicit filter}
\acrodef{HEAP}{Hydraulic Excavator for an Autonomous Purpose}
\acrodef{SMB}{Super Mega Bot}
\acrodef{RACER}{DARPA Robotic Autonomy in Complex Environments with Resiliency}
\acrodef{ROS}{Robot Operating System}
\acrodef{GT}{ground-truth}
\acrodef{PGT}{post-processed ground-truth}
\acrodef{SOTA}{state-of-the-art}
\acrodef{SE}{State Estimation}
\acrodef{SF}{sensor-fusion}
\acrodef{LI}{linear interpolation}
\acrodef{CT}{continuous-time}
\acrodef{DT}{Discrete Time}
\acrodef{RT}{real-time}
\acrodef{CPU}{central processing unit}
\acrodef{GPU}{Graphics Processing Unit}
\acrodef{GUI}{Graphical User Interface}
\acrodef{RL}{reinforcement learning}
\acrodef{RSL}{Robotic Systems Lab}
\acrodef{SubT}{DARPA Subterranean Challenge}
\acrodef{DOF}{Degree of Freedom}
\acrodef{ML}{machine learning}
\acrodef{ENU}{East-North-Up}
\acrodef{ATE}{absolute translation error}
\acrodef{ARE}{absolute rotation error}
\acrodef{RTE}{relative translation error}
\acrodef{RRE}{relative rotation error}
\acrodef{NIS}{Normalized Innovation Squared}
\acrodef{NOJ}{number of jumps}

%% file: custom/commands.tex
\definecolor{bananamania}{rgb}{0.93, 0.86, 0.51}

\renewcommand{\Comment}[1]{\hspace{0em}\textcolor{blue}{// #1}}

\newcommand{\cmark}{\textcolor{DarkGreen}{\ding{51}}} 
\newcommand{\xmark}{\textcolor{DarkRed}{\ding{55}}} 

\colorlet{CaptionColor}{gray!30}

\makeatletter
\renewcommand{\algocf@makecaption@ruled}[2]{%
  \global\sbox\algocf@capbox{\colorbox{CaptionColor}{%
    \parbox[t]{\columnwidth-2\fboxsep}{\algocf@captiontext{\strut#1}{\strut#2\strut}}}}
}%
\makeatother

\newtheorem{remark}{Remark}

\crefname{remark}{Remark}{Remarks}
\Crefname{remark}{Remark}{Remarks}
\crefname{paragraph}{paragraph}{paragraphs}
\Crefname{paragraph}{Sec.}{Secs.}
\Crefname{chapter}{Chap.}{Chaps.}
\Crefname{section}{Sec.}{Secs.}
\Crefname{subsection}{Sec.}{Secs.}
\Crefname{figure}{Fig.}{Figs.}
\Crefname{table}{Tab.}{Tabs.}
\Crefname{equation}{Eq.}{Eqs.}
\Crefname{appendix}{App.}{Apps.}

%% file: custom/during_writing.tex
\let\labelindent\relax
\usepackage{enumitem,amssymb}
\newlist{todolist}{itemize*}{2}
\setlist[todolist]{label=$\square$}
\usepackage{pifont}

\usepackage{lipsum}

%% file: sections/0-Abstract.tex
\begin{abstract}
Seamless operation of mobile robots in challenging environments requires low-latency \textit{local} motion estimation and accurate \textit{global} localization.
While most sensor-fusion approaches are designed for specific scenarios, this work introduces a flexible open-source solution for task- and setup-agnostic multimodal sensor fusion distinguished by its generality and usability.
\textit{Holistic Fusion} formulates sensor fusion as a combined estimation problem of \textit{i)} the local and global robot state and \textit{ii)} a (theoretically unlimited) number of dynamic variables, including automatic alignment of reference frames; this formulation fits countless real-world applications without conceptual modifications, offering a comprehensive solution beyond hard-coded/task-specific approaches.
The proposed factor-graph formulation enables direct fusion of an arbitrary number of absolute, local, and landmark measurements expressed with respect to different frames by explicitly including them as states in the optimization and modeling their evolution as random walks. 
Moreover, local smoothness and consistency receive particular attention to prevent estimation jumps.
\acl{HF} enables low-latency and smooth online state estimation on typical robot hardware while simultaneously providing low-drift global localization at the \acs{IMU} measurement rate. 
The efficacy of this released framework\footnote{\label{footnote:code}Code: {\scriptsize\anontext{\url{https://github.com/leggedrobotics/holistic_fusion}}{anonymous link}}} is demonstrated in five real-world scenarios on three robotic platforms with distinct task requirements, highlighting the advantages of fusing multiple absolute measurement types.\footnote{\label{footnote:project_page}Project: {\scriptsize\anontext{\url{https://leggedrobotics.github.io/holistic_fusion}}{anonymous link}}}\looseness=-1
\end{abstract}

\begin{IEEEkeywords}
State Estimation, Sensor Fusion, Factor Graph, Reference Frame Alignment, Online, Offline, Local, Global
\end{IEEEkeywords}

%% file: sections/1-Introduction.tex
\pdfpxdimen=\dimexpr 1 in/72\relax
\begin{figure}[ht!]
\includegraphics[width=\columnwidth]{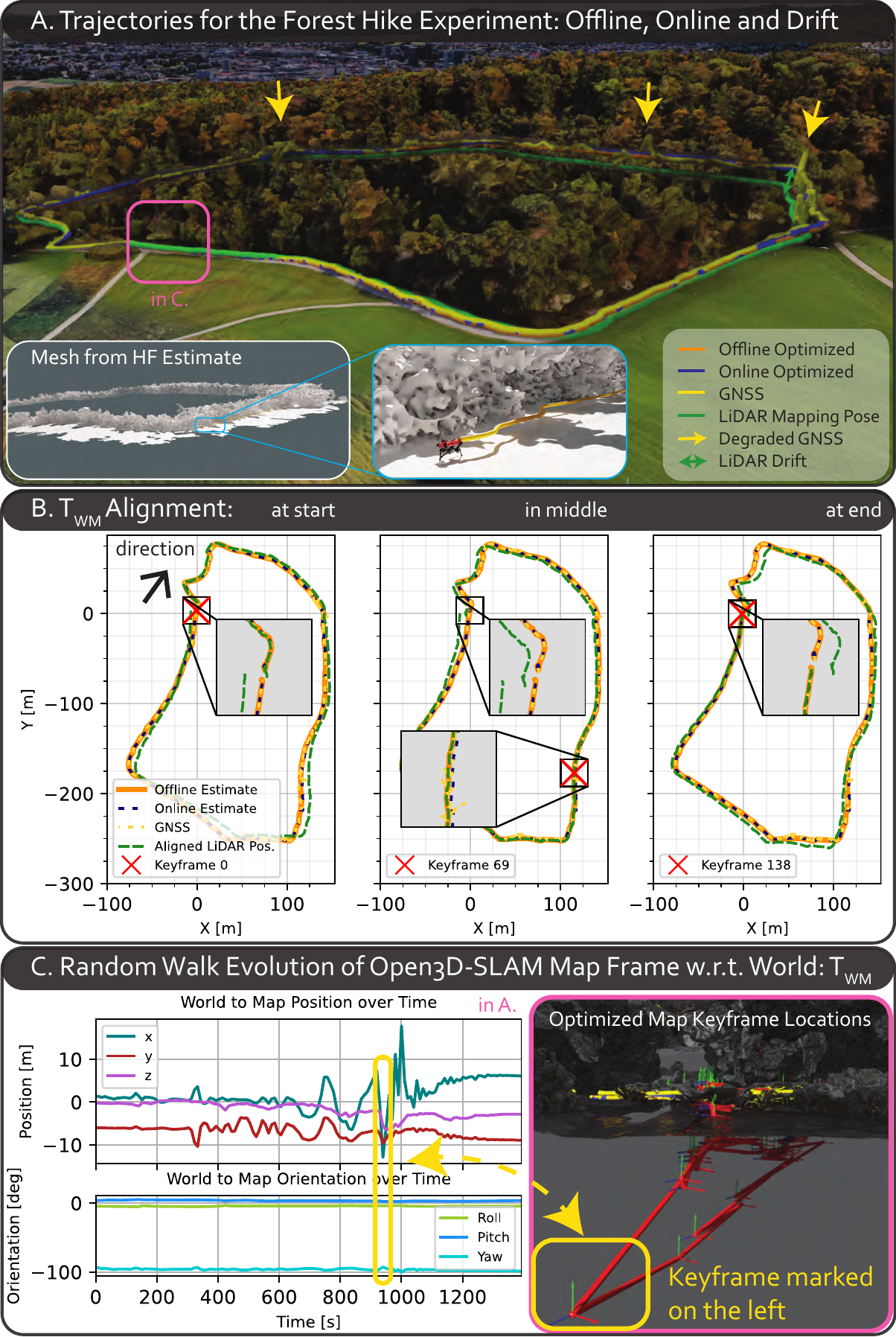}
\centering
\caption{Overview of a fully autonomous hiking experiment using the ANYmal quadrupedal robot in a forest environment. Online and offline motion estimation are conducted by fusing \textit{i)} IMU, \textit{ii)} noisy and degraded \acs{GNSS} measurements, \textit{iii)} the online-aligned (drifting) absolute pose of a \acl{LR} (LR) SLAM system expressed in the map frame, and \textit{iv)} leg odometry. \textbf{A:} Overview of the different trajectories with highlighted drift and \acs{GNSS} loss. \textbf{B:} The aligned \acs{LR} trajectory at three keyframes (indicated by $\times$). At each keyframe, the LR trajectory locally perfectly aligns. \textbf{C:} 2D and 3D visualizations of the random walk are used to illustrate the evolution of the map frame location. Yellow highlights the correspondence between the spikes of the 2D and 3D plots.\looseness=-1}
\label{fig:title_anymal_hike}
\vspace{-3ex}
\end{figure}


\IEEEPARstart{T}{hrough} advances in perception, control, and hardware design, modern mobile robotic systems can perform many complex tasks in the real world. 
From dynamic feedback-loop control and local navigation to object manipulation and global pathfinding, these tasks must often be performed while the robot is simultaneously traversing a challenging environment. 
Although task performance depends on robot localization and motion estimation, the exact requirements differ from task to task and can be highly \textit{context} specific. For example, legged locomotion~\cite{miki2022learning, wang2024learning} and UAV motion control~\cite{bouabdallah2007fullcontrol} require fast and locally consistent state estimates (most importantly, velocities), whereas motion planning~\cite{liusafe24} requires locally accurate poses and geometric maps. In contrast, tasks such as global planning~\cite{dang2020graph} and outdoor construction~\cite{spinelli2024reinforcement} require globally accurate positioning and world representation. \looseness-1

Robots need to be able to perform such tasks in widely varying conditions such as indoor, outdoor, underground~\cite{chung2023into}, and mixed~\cite{helmberger2022hilti} environments. 
These settings frequently pose additional challenges to robot state estimation due to, e.g., poor illumination or texture (affecting cameras)~\cite{khattak2020keyframe}, geometric degeneracy~\cite{nubert2021self,tuna2022x,tuna2024informed}, or disappearing/degrading localization signals (e.g., \ac{GNSS})~\cite{nubert2022graph}.\looseness=-1

\paragraph{Sensor Fusion}
Prior research has demonstrated that the fusion of multiple sensor modalities can improve robot pose estimation reliability\cite{khattak2020complementary}, accuracy~\cite{qin2019general}, and consistency~\cite{sandy2019confusion}, in particular for real-world field deployments such as the \ac{SubT} challenge~\cite{ebadi2023present}. 
Although research on learning-based methods for robot motion estimation~\cite{nubert2021self} and high-rate IMU-based state estimation~\cite{cioffi2023learned} is increasing, the majority of multimodal sensor fusion solutions still rely on traditional techniques such as filtering, often via (extended) \acp{KF}~\cite{mourikis2007multi, bloesch2013stateEKF}, and optimization-based solutions, often via smoothing and factor graphs~\cite{dellaert2017factor,superOdometry}.
Key challenges in the practical fusion of sensor measurements from multiple sources include \textbf{\textit{i)}} varying time delay, \textbf{\textit{ii)}} measurement availability at different rates, and \textbf{\textit{iii)}} disparities across the incoming measurements (type and reference frame). 
While \textbf{\textit{i)}} and \textbf{\textit{ii)}} must be handled with care in practice, a large body of research exists to address these challenges~\cite{superOdometry,lynen2013robust,bloesch2017two}.
Yet, the existing literature barely explores the disparities and environmental contexts of all measurements and allowing for their correct conversion in downstream applications (e.g., querying the optimized high-rate robot state in a low-rate measurement map frame). A simple example is the fusion of \ac{IMU} measurements with multiple (arbitrarily rotated) absolute sensor poses provided by multiple disconnected \ac{SLAM} systems.\looseness=-1

\paragraph{Holistic Fusion}
In response to these significant challenges, we present a flexible formulation called \emph{\ac{HF}} that can (theoretically) handle an arbitrary number of sensor measurements expressed in diverse reference coordinate frames. 
It directly includes the given coordinate frames in addition to the robot states and calibration states as optimization variables in one holistic factor-graph optimization, enabling smooth local estimates while guaranteeing full synchronized expression of the robot state in all present reference frames at \ac{IMU} sample rates.
The varying drift of the measurements expressed in the various reference frames is handled by \textit{explicitly} modeling the evolution of each reference frame location through a random walk (\Cref{fig:title_anymal_hike}).
To facilitate flexible fusion of out-of-order measurements, the \textit{prediction-update-loop} of~\cite{nubert2022graph} is adopted, allowing for the fusion of delayed measurements at arbitrary rate and order.
This design yields three main advantages: 
\textbf{\textit{i)}} The formulation seamlessly generalizes to different setups without modifying the \ac{HF} framework; it can handle the fusion of global and reference frame-based absolute measurements (such as positions), local quantities (such as feature tracks or estimated odometry), landmarks, and raw sensor measurements (such as \acp{IMU} or encoders). Interestingly, the inclusion of multiple absolute measurement types can improve performance over having only a single absolute measurement type.
\textbf{\textit{ii)}} The produced estimates are smoother and more consistent than \textit{either} a direct integration of assumed global quantities (e.g., the pose of an external \ac{SLAM} system), which is often not perfectly aligned with the gravity-aligned world frame, or the formulation of local quantities in the form of binary factors. 
\textbf{\textit{iii)}} The robot state can be seamlessly expressed in all current coordinate frames, for example, in the global 
frame or any map or odometry frame. Hence, the proposed framework also offers a clean and synchronized solution to localization management, overcoming the limitations of existing works.\looseness-1

\paragraph{Contributions}
\label{sec:contributions}
The contributions of this work are:
\begin{enumerate}
    \item A novel state-estimation formulation (\Cref{sec:method_holistic_fusion}) that fits a wide range of real-world scenarios. Beyond the robot state, it includes dynamic context variables, such as reference frames.
    These reference frames are automatically aligned in one holistic optimization, even handling inherent drift through explicit modeling of their evolution as a \textit{random walk} (\Cref{sec:method_drift_modeling}). \textit{Local keyframe alignment} (\Cref{sec:local_keyframe_alignment}) along the path enables robot missions that span large distances.\looseness=-1
    \looseness=-1
    \item The provision of \textit{smooth, not jumping, local estimates} (\Cref{sec:method_smoothness}) by considering the robot velocity in the body frame, in contrast to the state in the world frame.\looseness=-1
    \item A thorough experimental evaluation (\Cref{sec:experimental_results}) on three robot platforms and five different tasks in diverse environments, highlighting \ac{HF}'s wide-ranging applicability.\looseness=-1
    \item A comprehensive software framework\footref{footnote:code} (\Cref{sec:implementation}) with detailed documentation and examples spanning six platforms.\looseness=-1
\end{enumerate}

%% file: sections/2-RelatedWork.tex

\begin{table*}[t]
\centering
\caption{Non-complete overview of state-estimation and sensor-fusion approaches.
{\raisebox{-0pt}{{\small{\cmark}}}} indicates support.
{\raisebox{-2pt}{{\small\xmark}}} indicates limited or no support.}
\vspace{-2ex}
\begin{tabularx}{\textwidth}{>{\raggedright\arraybackslash}l|>{\raggedright\arraybackslash}c|>{\raggedright\arraybackslash}c|>{\raggedright\arraybackslash}c|>{\raggedright\arraybackslash}c|>{\raggedright\arraybackslash}c|>{\raggedright\arraybackslash}c|>{\raggedright\arraybackslash}c|>
{\raggedright\arraybackslash}c|>
{\raggedright\arraybackslash}c|>
{\centering\arraybackslash}X}
    \hline
    \rowcolor{CaptionColor} 
    \textbf{Capability} & \makecell{MSF \\ \cite{lynen2013robust}} & \makecell{TSIF \\ \cite{bloesch2017two}} & \makecell{WOLF \\ \cite{sola-et-al-2022-wolf}} & \makecell{SuperOdometry \\ \cite{superOdometry}} & \makecell{MaRS \\ \cite{brommer2020mars}} & \makecell{OKVIS2 \\ \cite{leutenegger2022okvis2}} & \makecell{maplab 2.0 \\ \cite{schneider2018maplab,cramariuc2022maplab}} & 
    \makecell{MINS \\ \cite{lee2023mins}} & 
    \makecell{gnssFGO \\ \cite{gnssFGO}}& \textbf{\ac{HF}} \\
    \hline

    High-Rate Online Estimation (\textbf{\textit{O1.1}}) & \cmark & \cmark & \cmark & \cmark & \cmark & \cmark & \cmark & \cmark & \cmark & \cmark \\ 
    Dedicated Smooth Odometry (\textbf{\textit{O1.2}})   & \xmark & \xmark & \cmark & \cmark & \xmark & \xmark & \xmark & \cmark & \cmark & \cmark \\ 
    Focus on Usability (\textbf{\textit{O2}})            & \cmark & \cmark & \cmark & \xmark & \cmark & \cmark & \cmark & \cmark & \cmark & \cmark \\ 
    Reference Frame Alignment (\textbf{\textit{O3}})     & \xmark & \xmark & \cmark & \cmark & \cmark & \xmark & \cmark & \cmark & \cmark & \cmark \\ 
    Attitude (Gravity) Alignment (\textbf{\textit{O3}}) & \cmark & \cmark & \xmark & \xmark & \xmark & \xmark & \xmark & \xmark & \xmark & \cmark \\ 
    Extrinsic Calibration (\textbf{\textit{O4}})         & \cmark & \xmark & \cmark & \xmark & \cmark & \cmark & \cmark & \cmark & \xmark & \cmark \\ 
    Focus on Adaptability (\textbf{\textit{O5}})         & \xmark & \xmark & \cmark & \cmark & \cmark & \xmark & \cmark & \xmark & \xmark & \cmark \\ 
    Open-Source Implementation                           & \cmark & \cmark & \cmark & \xmark & \cmark & \cmark & \cmark & \cmark & \cmark & \cmark \\
    Offline State Estimation                                   & \xmark & \xmark & \xmark & \xmark & \xmark & \xmark & \cmark & \xmark & \cmark & \cmark \\ 
    Time Synchronization                                 & \cmark & \xmark & \xmark & \xmark & \xmark & \xmark & \cmark & \cmark & \cmark & \xmark \\ 
    \hline
\end{tabularx}
\label{tab:methods}
\vspace{-3ex}
\end{table*}

\subsubsection{Multi-Sensor Fusion in Robotics}
Multi-sensor fusion for localization and state estimation has been widely studied. Filtering-~\cite{lynen2013robust,bloesch2017two} and optimization-based methods for moving-horizon~\cite{indelman2013information, kilic2019improved} and batch-optimization~\cite{diehl2009efficient, sandy2019confusion} have been proposed to estimate the current robot state. An explicit comparison between both for the case of GNSS \& IMU fusion can be found in~\cite{imuGPS2020}.
In recent years, many different sensors have been investigated to enable accurate robot localization at large scale, including \ac{LiDAR} sensors~\cite{loam2018,jelavic2022open3d}, cameras~\cite{geneva2020openvins}, \ac{RADAR}~\cite{hong2020radarslam}, wheel odometry~\cite{kilic2019improved}, barometers~\cite{lynen2013robust} and GNSS~\cite{imuGPS2020, baehnemann2022under, cao2022gvins}. 
Multi-sensor fusion and state estimation can be divided into two general paradigms: loosely and tightly coupled fusion. In addition, systems can be categorized as \textit{specialized} and application-specific, or \textit{generic} and task-agnostic. 
More tightly coupled systems, such as \ac{LiDAR} inertial odometry~\cite{xu2021fast}, \ac{LiDAR}-visual-inertial odometry~\cite{zhang2015visual} or \ac{LiDAR}-visual-kinematic-inertial odometry~\cite{wisth2022vilens}, have generally shown superior accuracy compared to their loosely coupled counterparts, in particular for non-RTK \ac{GNSS} systems~\cite{falco2017loose}.
In contrast, for applications where robustness is a primary concern, often deployed by the field robotics community, a combination of tightly coupled subsystems and loosely coupled top-level fusion is prevalent~\cite{khattak2020complementary,superOdometry,ebadi2023present}, as seen during the \textit{DARPA SubT} challenge~\cite{chung2023into}, due to increased flexibility and easier handling of outliers or sensor failures.\looseness=-1

\paragraph{Specialized Systems}
Most proposed frameworks are tailored to particular problems or sensor setups. For example, SuperOdometry~\cite{superOdometry}, one of the top-performing odometry systems during the \textit{DARPA SubT} challenge, mixes loose and tight approaches with an IMU-centric architecture. Three different specialized factor graphs are deployed and combined in a central formulation. Similarly, the WildCat~\cite{ramezani2022wildcat} 3D LiDAR-inertial SLAM system achieved the lowest registration error  of the \textit{DARPA SubT} environment. WildCat uses a continuous-time trajectory representation in a sliding-window LiDAR-inertial mapping module with a pose-graph optimization framework to ensure global consistency. 
Moreover, other works investigate the tight fusion of IMU and GNSS using Kalman filter variants~\cite{dong2020tightGnss, falco2017loose,imuGPS2020, lee2023mins} and smoothers~\cite{boche2022visual,imuGPS2020,nubert2022graph,baehnemann2022under, cioffi2020tightly,mascaro2018gomsf,cao2022gvins, gnssFGO}, parts of which are discussed in more detail in the context of mixed-environment operation in~\Cref{sec:mixed_environments}.\looseness=-1

\paragraph{State Estimation for Legged Robots}
A family of works introduced estimators exploiting accurate joint position measurements through the encoder readings of legged robots. 
First, Bloesch et al.\ introduced an \ac{EKF} framework fusing inertial data, leg kinematics, and contact information for robust legged odometry~\cite{bloesch2013stateEKF} and later an \ac{UKF} framework to handle the system's nonlinearities better~\cite{bloesch2013stateUKF}. 
Later, the same authors presented a \ac{TSIF}~\cite{bloesch2017two}, 
reducing the required model accuracy while maintaining real-time performance for challenging motions.
Meanwhile, Camurri et al.~\cite{camurri2017ral} proposed a probabilistic contact estimation method. Building on this estimator, Nobili et al.~\cite{nobili2017rss} addressed drift by introducing additional visual and LiDAR matching constraints, showing the benefit of multimodal state estimation. Following this, Pronto~\cite{camurri2020pronto} demonstrated reliable odometry in rough terrain by leveraging accurate foot-contact detection, high-rate IMU, and joint measurements with low-rate exteroceptive corrections. Utilizing Pronto, Buchanan et al.~\cite{buchanan2022learning} proposed estimating motion displacement with learning-based models through IMU data to reduce dependency on exteroceptive measurements, illustrating how data-driven approaches can aid state estimation on unstructured terrain.
Wisth et al. presented VILENS~\cite{wisth2022vilens}, a visual, inertial, LiDAR, and legged state estimator, enhancing localization robustness by utilizing outlier rejection techniques such as dynamic covariance scaling and fusing multiple complementary sensors using an optimization-based factor-graph backend. Like VILENS, Yang et al.~\cite{yang2023cerberus} proposed a visual-inertial-leg fusion pipeline to address the operation challenges in degraded environments, highlighting the importance of contact outlier rejection. 
Focusing on the accuracy of kinematic measurements, Kim et al.~\cite{kim2022step} introduced STEP (pre-integrated foot velocity) to explicitly model foot kinematics, reducing estimation drift in uncertain foothold conditions. Yoon et al.~\cite{yoon2023invariant} offered consistent state estimates under dynamic motions and challenging terrain for accurate contract estimation utilizing an invariant \ac{KF} framework and explicit slip detection.\looseness=-1 

\paragraph{Low-Rate (\ac{SLAM}) Systems}
Modern SLAM algorithms focus on robustly estimating the robot pose while ensuring map consistency through loop closures. 
For example, Mur-Artal et al.~\cite{mur2015orb} introduced the keypoint-based ORB-SLAM to perform visual SLAM accurately and efficiently. Later, Leutenegger~\cite{leutenegger2022okvis2} proposed OKVIS2, a framework for visual-inertial SLAM with loop closure for global consistency. Moreover, leveraging the differences between sensor modalities, Khattak et al.~\cite{khattak2020complementary} proposed complementary sensor fusion for visual-inertial, thermal-inertial, and LiDAR-based constraints for robot pose estimation and map creation in \ac{GNSS}-denied and degenerate environments. Showing the strength of LiDAR-only methods, \cite{jelavic2022open3d,vizzo2023kiss} leverage solely scan-to-(sub)map registration for non-agile maneuvers.
Utilizing offline batch processing, maplab 2.0~\cite{schneider2018maplab, cramariuc2022maplab} offers a flexible framework for large-scale optimization, mapping, and data re-processing for camera-based measurements by also estimating time-offset, intrinsic, and extrinsic calibrations.  
Similarly, BALM~\cite{liu2021balm} applied the classical bundle adjustment formulation to LiDAR-based constraints instead of camera-based constraints to provide accurate and consistent LiDAR pose and environment representation for long-duration missions.
Relying on the recent advancements in computing resources, many works employ pose-graph optimization, bundle adjustment, and loop closure in a delayed and multi-threaded fashion, allowing refined global consistency~\cite{jelavic2022open3d, leutenegger2022okvis2}. Similarly, utilizing the formulation of BALM~\cite{liu2021balm}, Liu et al.~\cite{liu2024voxel} proposed a versatile and accurate LiDAR mapping and optimization framework, utilizing local and global features in a bundle-adjustment context.\looseness=-1

\paragraph{Generic Multi-Purpose Fusion Systems}
MSF~\cite{lynen2013robust} is a generic filter-based framework for fusing measurements without additional coding effort. Although extrinsic calibration is allowed as part of the estimation, it assumes that all measurements are expressed in the same coordinate frame.
Similarly, \ac{TSIF}~\cite{bloesch2017two} proposes a \ac{KF} for the fusion of measurements without explicit knowledge of the underlying process model by exploiting purely residual-based modeling. Although it achieves a state-of-the-art solution for some applications, e.g., leg-inertial odometry on ANYmal~\cite{hutter2016anymal}, the approach's filter-based nature and formulation effort render its usage in different applications difficult.
In contrast to the two previous methods, WOLF~\cite{sola-et-al-2022-wolf} aims to achieve generic fusion using nonlinear optimization in the form of a factor graph. Similar to~\cite{lynen2013robust}, WOLF is usable with little to no custom code, enabling fast prototyping and development. Yet, it is unclear whether the framework can handle multiple absolute measurements.
The method proposed in this work can be understood as a general multi-purpose solution, since a (theoretically) unlimited number of poses, positions, velocities, and headings expressed in arbitrary coordinate frames can be fused in one holistic optimization.\looseness=-1

\paragraph{Learning-Based Solutions}
In recent years, \ac{ML} approaches have played an increasing role in the estimation community to tackle parts of the estimation pipeline that have been computationally expensive or difficult to model with classical techniques, such as solving the point-cloud registration problem~\cite{li2019net,nubert2021self} or detecting environmental degeneracy~\cite{nubert2022learning}.
Moreover, recently, learning-based solutions have been used to overcome the ill-posed nature of classically underconstrained estimation problems, e.g., for the case of IMU-only motion estimation on drones~\cite{cioffi2023learned}, for displacement learning in combination with a stochastic cloning-based \ac{EKF} for human motion estimation~\cite{liu2020tlio} or leg odometry~\cite{buchanan2022learning}, or for improved \ac{IMU}-based dead reckoning~\cite{brossard2020ai}.
However, while a fascinating field of research, learning-based approaches are still not widely applied in real-world robotic applications beyond local odometry estimation due to limitations in absolute accuracy and generalization to out-of-distribution scenarios.\looseness=-1

\subsubsection{Practical Considerations}
\paragraph{Delayed \& Out-of-Order Measurements}
To deal with delays, \ac{KF} implementations often either augment the state vector~\cite{gopalakrishnan2011ekf} or recalculate the filter by updating the measurement sequence stored in a buffer~\cite{lynen2013robust}. These steps increase implementation effort and computational complexity compared to the standard \ac{EKF}~\cite{julier1997new}. Due to their history, optimization-based methods can naturally deal with delayed measurements  simply by inserting a measurement factor at the corresponding past timestamp. 
To benefit from the flexibility of graph-based methods while still being able to provide motion estimates at the IMU frequency, this work builds on the optimization-based prediction update loop of Nubert et al.~\cite{nubert2022graph}.
Moreover, the problem of measurements arriving at arbitrary timestamps is usually addressed through \textit{i)} explicit hardware triggering~\cite{baehnemann2022under}, \textit{ii)} additional software engineering~\cite{lynen2013robust}, or by \textit{iii)} interpolating~\cite{talbot2024continuous} or pre-integrating to the correct timestamps~\cite{forster2016manifold}.\looseness=-1

\paragraph{Operation in Mixed Environments}
\label{sec:mixed_environments}
One major challenge when fusing various measurements in real-world applications is managing the different local or global coordinate systems in which the measurements are expressed. 
In~\cite{boche2022visual}, the authors proposed a visual-inertial and GNSS solution to tackle drift-free state estimation outdoors. They showed the importance of robust frame alignment after the GNSS dropout and during initialization. Similarly, \cite{cao2022gvins} integrated raw Doppler shift measurements and a coarse-to-fine Earth-centered anchor optimization scheme to integrate global measurements into the estimation problem correctly. Similarly, to be able to operate in a globally accurate manner despite \ac{GNSS} dropout,~\cite{nubert2022graph} proposed the use of a dual-factor graph formulation to cover the scenarios when \ac{GNSS} is present or not.
While measurements are usually expressed in specific reference frames, the robotic state estimation community has not addressed explicit drift modeling of these, hindering the usage of multiple (global and local) reference frames in a single optimization.
As a response, \acl{HF} seamlessly handles operation in mixed environments, as all measurements are considered in a \textit{semi-global} context, and the coordinate frame transformations are introduced as states in the optimization, including explicitly modeled drift.\looseness-1

%% file: sections/3-ProblemFormulation.tex

\pdfpxdimen=\dimexpr 1 in/72\relax
\begin{figure}[t]
\includegraphics[width=\columnwidth]{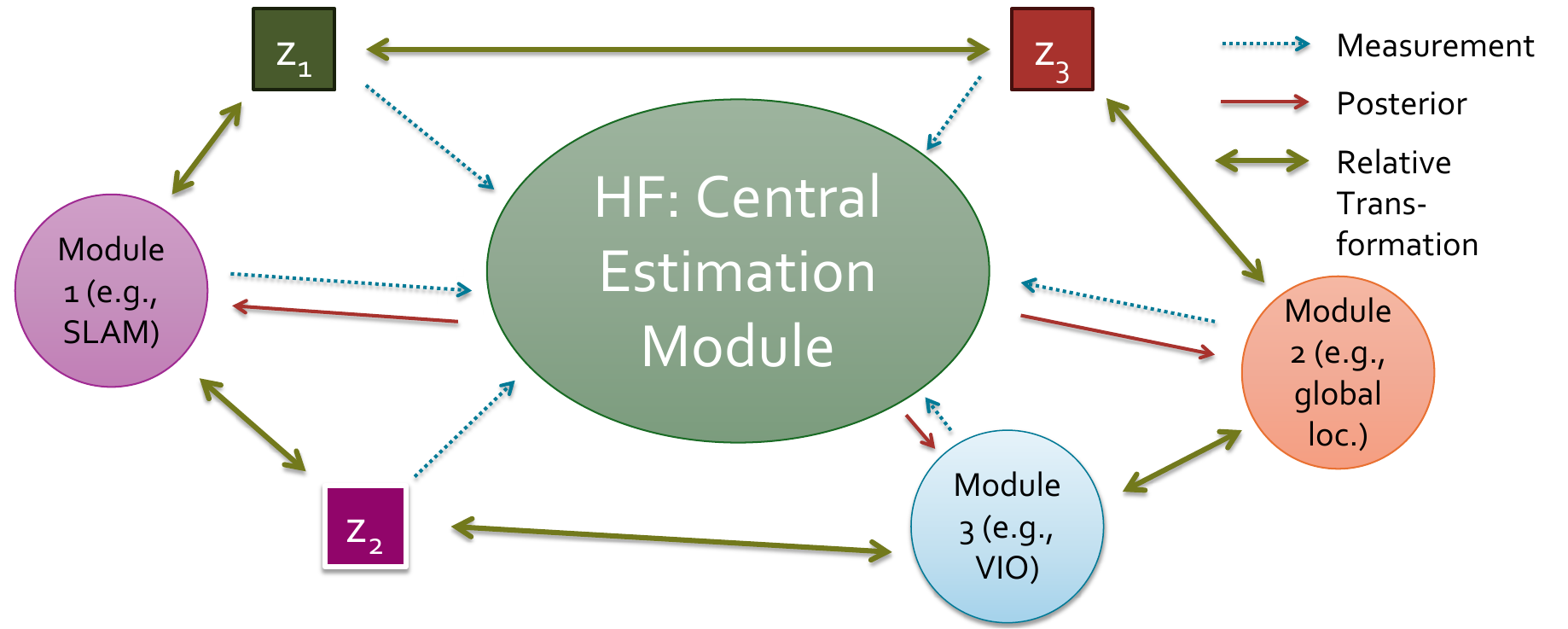}
\centering
\caption{Illustration of the intended role, capabilities and information streams of \ac{HF}. Not only is \ac{HF} acting as a central fusion module, but it also \textit{i)} estimates (relative) transformations between the reference frames of the measurements or external modules, and it \textit{ii)} allows for either high- or low-rate beliefs to be propagated back to the modules, which can use them as priors. An example of an external module is a \ac{SLAM} system, as used in all experiments in \Cref{sec:experimental_results}.}
\label{fig:holistic_fusion_overview_scope}
\vspace{-3ex}
\end{figure}

\begin{figure*}[t]
\includegraphics[width=\textwidth]{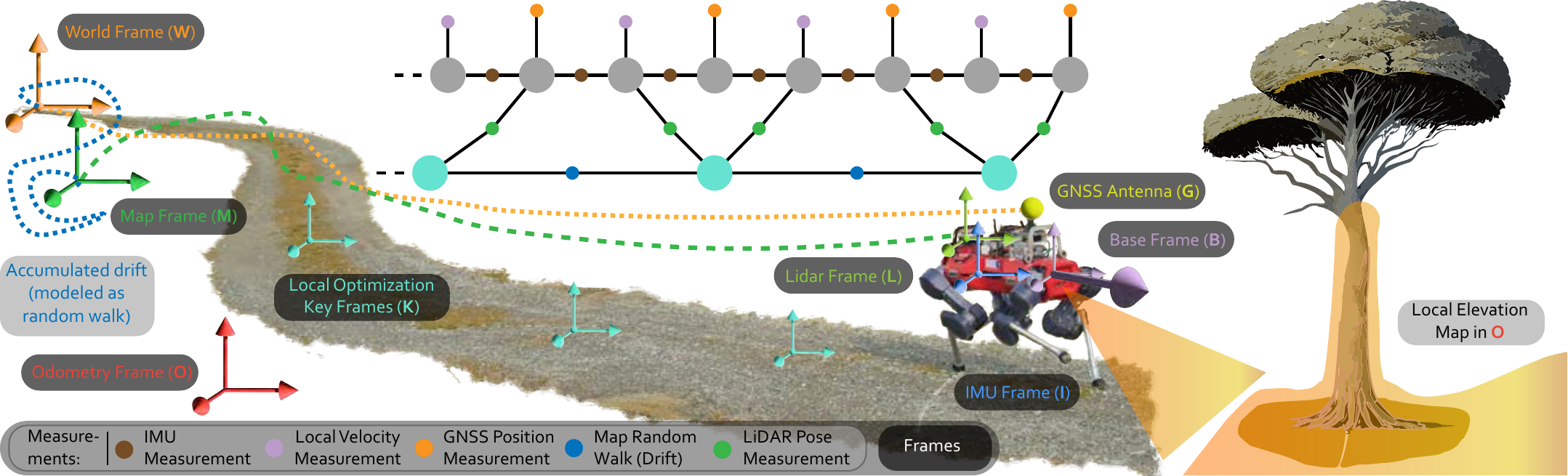}
    \centering
    \caption{An illustrative scenario of \acs{HF}. During the mission, different measurements are fused directly without preprocessing (in the shown example: \acs{IMU}, \textcolor{orange}{\acs{GNSS}}, \textcolor{ForestGreen}{\acs{LR}}, \textcolor{Orchid}{local velocity}). To fuse global and non-global absolute measurements, which drift against each other (\textcolor{NavyBlue}{blue path}), \ac{HF} explicitly estimates the shift between the reference coordinate frames, aligns the measurements, and models them as a random walk. This alignment is not conducted at the global origin but at \textcolor{TealBlue}{local keyframes}. The graph at the top shows a simplified version of the resulting graph with factors and states depicted in the same colors.\looseness=-1}
\label{fig:holistic_fusion_illustration}
\vspace{-2ex}
\end{figure*}

\subsection{Problem Formulation}

\subsubsection{Goal}
\ac{HF} acts as a central state-estimation module, estimating the robot state from raw measurements and the processed outputs of other modules, as shown in \Cref{fig:holistic_fusion_overview_scope}.
\ac{HF} also aims to \textit{i)} optimize the reference frames of all fused measurements and modules and \textit{ii)} provide high-rate estimates to be fed back to the corresponding submodules.

The particular \textbf{objectives} for the state-estimation problem in mixed and large-scale environments are the following:
\begin{enumerate}
    \item \textbf{\textit{O1}}: Estimation of the real-time robot state at the current time $t_k$. This, in particular, includes the following points:
    \textbf{\textit{O1.1}}: Low-latency and \ac{IMU}-rate estimation to enable real-time low-level control.
    \textbf{\textit{O1.2}}: Locally smooth and consistent estimates in $\Odom$ (\Cref{sec:core_coordinate_frames}) suitable for control and local navigation.
    \textbf{\textit{O1.3}}: Globally accurate estimates in $\World$ (\Cref{sec:core_coordinate_frames}) for global navigation and wayfinding.
    \item  \textbf{\textit{O2}}: Flexible fusion of any number and type of delayed, out-of-order measurements without prior engineering.
    \item  \textbf{\textit{O3}}: Determination of the relationship between different reference frames, i.e., automatically aligning all frames, constituting a synchronized localization manager (robot state can be expressed in any reference frame at full rate).
    \item  \textbf{\textit{O4}}: Extrinsic calibration (online and offline) between each sensor frame $\Sensor_i$ (\Cref{sec:sensor_frames}) and its calibrated $\Sensor_{i, \text{corr.}}$.
    \item  \textbf{\textit{O5}}: Easy use of the framework for new setups/tasks; factor generation for new measurements should be simple.\looseness=-1
\end{enumerate}

\subsubsection{Illustration of \acs{HF}}
Traditionally, the primary goal of state estimation in robotics lies in estimating the robot's motion (in $\World$), given a set of measurements. 
A simplified example of a real-world outdoor robot mission setup (cf. \Cref{sec:exp_anymal_hike}) is shown in \Cref{fig:holistic_fusion_illustration}. The robot motion should be determined given absolute \ac{GNSS} measurements, map-based \ac{LR} measurements, local leg odometry, and IMU measurements. 
Contrary to previous work, \ac{HF} allows for the direct fusion of two absolute measurements: \textit{i)} global \ac{GNSS} position and \textit{ii)} \ac{LR} $\SEthree$-poses expressed in the map frame.
In particular, \ac{HF} explicitly aligns the \ac{LR} $\SEthree$ measurements provided in the \ac{LiDAR} sensor frame (green path) with the $\Real^3$ \ac{GNSS} measurements provided in the \ac{GNSS} antenna frame (orange path). 
To avoid increasing disparity due to the drifting \ac{LiDAR} estimates, \ac{HF} explicitly models the relationship between the map and global coordinate frames as a random walk, allowing the map frame to shift relative to the world frame over time.
This choice eliminates the need to convert the \ac{LR} poses either manually to relative measurements, e.g.,~\cite{superOdometry}, or to set the alignment between reference frames independently of the core optimization, e.g.,~\cite{nubert2022graph}.
The estimation in the world frame $\World$ (orange coordinate frame in \Cref{fig:holistic_fusion_illustration}) aims to be as accurate as possible but might experience jumps in the belief, whereas the estimate in the odom frame $\Odom$ (red coordinate frame in \Cref{fig:holistic_fusion_illustration}) tries to be as smooth as possible. This prioritization is analogous to the typical map $\rightarrow$ odom relationship used in robotics, often combined with a localization manager.
The difference from a selection of previous estimators and frameworks is further illustrated in \Cref{tab:methods}. While frameworks exist for each objective \textit{\textbf{O1}}--\textit{\textbf{O5}}, \ac{HF} is the most versatile public framework to date, with a clear focus on practicality and usability for real-world robotic tasks.\looseness=-1

\subsection{Problem Structure}

\subsubsection{Frame Definitions}

\paragraph{Core Coordinate Frames}
\label{sec:core_coordinate_frames}
The core frames of every holistic fusion estimation problem are the fixed world frame $(\World)$, the odometry frame $(\Odom)$, and the \ac{IMU} frame $(\Imu)$ of the central \ac{IMU}. Multiple IMU sensors can be used, but one is designated as the central \ac{IMU}, which serves as the core sensor for the \ac{HF} estimation framework. Moreover, the robot is assumed to have a base frame $(\Base)$, which may or may not coincide with $\Imu$.\looseness=-1

\paragraph{(Robot-Specific) Sensor Frames}
\label{sec:sensor_frames}
Every measurement is assumed to be expressed w.r.t. a sensor frame $(\Sensor_i)~\forall~i \in \{1, \dots, N_{F_S}\}$, where $N_{F_S}$ is the number of sensors. Here, the coordinate system $\Sensor_i$ describes the coordinate origin of the $i$-th sensor, with $\Sensor_0$ coinciding with $\Imu$, for the central \ac{IMU} sensor. This work assumes that all sensor frames are rigidly connected to $\Imu$ through a (potentially unknown) transformation $T_{\Imu,\Sensor_i}$.
The sensor frames established for the ANYmal~\cite{hutter2016anymal} quadrupedal robot are illustrated in \Cref{fig:holistic_fusion_illustration}.\looseness=-1

\begin{table}[t]
\centering
\caption{Overview of the coordinate frames of the three robotic systems investigated in \Cref{sec:experimental_results}. Core frames are always present in \ac{HF}.\looseness=-1}
\vspace{-2ex}
\begin{tabularx}{\columnwidth}{>{\raggedright\arraybackslash}c|>{\raggedright\arraybackslash}c|>{\raggedright\arraybackslash}c|>{\centering\arraybackslash}X}
    \hline
    \rowcolor{CaptionColor} 
    $\Reference_i$ & $\Sensor_i$ & \textbf{Symbol} & \textbf{Name} \\
    \hline
    \rowcolor{gray!12} 
    \multicolumn{4}{c}{Core (present in every \ac{HF} estimation problem)} \\
    \hline
    \cmark &        & $\World$ & Main fixed world frame \\
    \cmark &        & $\Odom$  & Drifting odometry frame \\
           & \cmark & $\Imu$   & Central \ac{IMU} frame \\
           & \cmark & $\Base$  & Robot base frame \\
    \hline
    \rowcolor{gray!12} 
    \multicolumn{4}{c}{ANYmal~\cite{hutter2016anymal}, quadrupedal robot, indoor \& outdoor, \Cref{sec:exp_anymal}} \\
    \hline
    \cmark &        & $\ENU$       & Fixed GNSS \ac{ENU} frame \\
    \cmark &        & $\OpenThree$ & Open3D-SLAM~\cite{jelavic2022open3d} map frame \\
           & \cmark & $\Gnss$      & Single GNSS antenna frame \\
           & \cmark & $\Velodyne$     & Velodyne LiDAR frame \\
           & \cmark & $\Leg$       & Leg odometry frame \\
    \hline
    \rowcolor{gray!12} 
    \multicolumn{4}{c}{\anontext{RACER}{anonymous platform}, offroad vehicle, highly dynamic, \Cref{sec:exp_racer}} \\
    \hline
    \cmark &        & $\ENU$       & Fixed \ac{GNSS} \ac{ENU} frame \\
    \cmark &        & $\Liopard$   & LIO~\cite{rose} map frame \\
           & \cmark & $\Velodyne$  & Velodyne \ac{LiDAR} frame \\
           & \cmark & $\Radar$     & \ac{RADAR} frame \\
           & \cmark & $\WheelAxis$ & Wheel axis center \\
           & \cmark & $\Gnss$      & \ac{GNSS} antenna \\
    \hline
    \rowcolor{gray!12} 
    \multicolumn{4}{c}{\anontext{\acs{HEAP}~\cite{heap2021}, walking}{anonymous platform}, excavator, environmental degeneracy, \Cref{sec:exp_heap}} \\
    \hline
    \cmark &        & $\ENU$       & Fixed GNSS \ac{ENU} frame \\
    \cmark &        & $\Compslam$  & CompSLAM~\cite{khattak2020complementary} map frame \\
    \cmark &        & $\Coinlio$   & Feature-based CoinLIO~\cite{pfreundschuh2024coin} map frame \\
           & \cmark & $\GnssLeft$  & Left GNSS antenna frame \\
           & \cmark & $\GnssRight$ & Right GNSS antenna frame \\
           & \cmark & $\Ouster$    & Ouster LiDAR frame \\
    \hline
\end{tabularx}
\label{tab:frames_overview}
\vspace{-3ex}
\end{table}

\paragraph{(Setup-Specific) Reference Frames}
\label{sec:ref_frames}
In contrast\replace{Si}{} to relative measurements (e.g., odometry), absolute or landmark measurements are expressed w.r.t. a reference frame $(\Reference_i)~\forall~i \in \{1, \dots, N_{F_R}\}$, where $N_{F_R}$ denotes the number of reference coordinate frames. In the holistic fusion formulation, two reference frames $(\Reference_{\{0, 1\}})$ that always exist are $\Reference_0 \equiv \World, \Reference_1 \equiv \Odom$. 
Additionally, more reference frames can be present, such as the map frame of a (drifting) mapping or localization framework, the odometry frame of an (even more drifting) odometry solution, or additional fixed frames from a non-drifting localization system such as \ac{GNSS}, fixed \ac{UWB} markers, or \ac{mocap}. 
\Cref{fig:holistic_fusion_illustration} shows a simplified example of possible world and map reference frames.
All coordinate frames (reference and sensor) used in this work are listed in \Cref{tab:frames_overview}.\looseness=-1

\subsubsection{Measurements}
\label{sec:problem_form_measurements}
To reliably estimate the state of the robot, a variety of measurements can be used, e.g., originating from (wheel and joint) encoders, \acp{LiDAR}, \ac{RADAR}, cameras, \ac{UWB} or \ac{GNSS} antennas.
Like other factor-graph and filter-based estimation formulations, HF can generally handle arbitrary measurements.
\replace{}{Note that throughout this work, the term \textit{measurement} (denoted $\Measurement$ or $\Measurements$ for a set) refers exclusively to actual observations returned by a sensor or external module, and not to the constraints (factors) in the factor graph; the latter additionally involve the measurement function $\Measfunction(\state)$ and potentially additional state variables as detailed in \Cref{sec:method_holistic_fusion}.}
In particular, the following four categories are explicitly supported out of the box: IMU measurements, absolute measurements, landmark measurements, and local/relative measurements. These categories cover most use cases encountered in practical robotic state-estimation problems, allowing the fusion of non-gravity-aligned and drifting measurements expressed in arbitrary sensor frames with respect to arbitrary reference frames.
\looseness=-1

\paragraph{\acs{IMU} Measurements}
Due to their high rate and affordability, this work assumes that at least one \ac{IMU} sensor is available. 
While multiple \ac{IMU} sensors are theoretically supported, one of the \acp{IMU} must be designated as the \textit{core} measurement, defining both the core frame for estimation and the state-creation rate.
The corresponding measurements of the \ac{IMU} are:\looseness=-1
\begin{equation}
    ^{\Imumeas} \Measurement = {}_\Imu \Measurement_{\World \Imu} = 
    \begin{bmatrix}
        _{\Imu} \accel_{\World \Imu} \\
        _{\Imu} \rotvel_{\World \Imu}
    \end{bmatrix}
    ~\text{with}~
    \accel \in \Real^3,
    \rotvel \in \Real^3.
\end{equation}
Here, ${}_\Imu \Measurement_{\World \Imu}$ refers to the motion of the \ac{IMU} frame $\Imu$ w.r.t. the world frame $\World$ expressed in the local IMU frame $\Imu$ (cf. the ``Core'' frames in \Cref{tab:frames_overview}), and $\accel$ and $\rotvel$ are the measured acceleration and angular velocity.
All \ac{IMU} measurements until timestep $k$ are denoted as $^\Imumeas \Measurements_k \triangleq \{ ^\Imumeas \Measurement \}_{i \in ^\Imumeas \mathcal{K}_{N_{\Imumeas}}}$, where $^\Imumeas \mathcal{K}_{N_{\Imumeas}}$ is the set of all \ac{IMU} measurement times until time $t_k$.\looseness=-1

\paragraph{Absolute Measurements}
\label{sec:absolute_measurements}
This category expresses certain \textit{absolute} quantities w.r.t. a reference frame:
\begin{equation}
    ^\Absolute \Measurement \doteq {}_{\Reference_i} \Measurement_{\Reference_i \Sensor_i}.
\end{equation}
Here, ${}_{\Reference_i} \Measurement_{\Reference_i \Sensor_i}$ expresses an absolute measurement of a sensor $\Sensor_i$ w.r.t. a reference frame $\Reference_i$, expressed in the same reference frame.
Examples include measured GNSS positions, $_{\ENU}\transl_{\ENU \Gnss} \in \Real^3$, poses coming from a \ac{LiDAR} mapping framework, $\T_{\Map \Lidar} \in \SEthree$, or linear velocities given in an external fixed frame $_{\Reference_i}\vel_{\Reference_i \Sensor_i}$.
These exemplary reference and sensor frames are also highlighted in \Cref{tab:frames_overview}.
All absolute measurements until timestep $k$ are denoted as 
$^\Absolute \Measurements_k \triangleq \{ ^\Absolute \Measurement \}_{i \in ^\Absolute \mathcal{K}_{N_{\Absolute}}}$.
The flexible handling of multiple absolute measurements expressed in multiple (different) reference frames is among the core strengths of the proposed approach, as it circumvents the typical conversion of absolute measurements (except the ones expressed in \replace{world}{the global world frame $\World$}) to local quantities, as \replace{e.g., }{}done in works such as~\cite{superOdometry}.\looseness=-1

\paragraph{Feature Landmark Measurements}
\label{sec:meas_landmark}
\ac{HF} also allows landmark measurements. For example, those could be measured camera features for \ac{VIO} or foot contact points for legged robots. 
They are defined as:\looseness=-1
\begin{equation}
    ^\Landmark \Measurement \doteq {}_{\Sensor_i} \Measurement_{\Sensor_i \Feature_m}.
\end{equation}
Here, ${}_{\Sensor_i} \Measurement_{\Sensor_i \Feature_m}$ is the landmark feature location $\Feature_m$ measured w.r.t. the sensor frame $\Sensor_i$ and expressed in the local sensor frame $\Sensor_i$. More specifically, $\Feature_m$ denotes the $m$-th landmark feature coordinate, which could be a position (e.g., for camera point features) or $\SEthree$ pose (e.g., for footstep poses).
In \Cref{tab:frames_overview}, the \replace{}{corresponding} sensor frame for the leg odometry landmark \replace{}{in the given example} is denoted as $\Leg$.
All feature landmark measurements until timestep $k$ are denoted as $^\Landmark \Measurements_k \triangleq \{ ^\Landmark \Measurement \}_{i \in ^\Landmark \mathcal{K}_{N_{\Landmark}}}$.\looseness=-1

\paragraph{Local \& Relative Measurements}
Another vital class is relative/local measurements\replace{ that}{, which} can read as either
\looseness=-1
\begin{equation}
    ^\Local \Measurement \doteq {}_{\Sensor_i} \Measurement_{\World \Sensor_i},
\quad\text{or}\quad
    ^\Local \Measurement \doteq {}_{\Sensor_{i,k}} \Measurement_{\Sensor_{i,k} \Sensor_{i, k+1}}.
\end{equation}
Here, ${}_{\Sensor_i} \Measurement_{\World \Sensor_i}$ is a measurement in the sensor frame $\Sensor_i$ w.r.t. the world frame $\World$, expressed in the local sensor frame $\Sensor_i$; an example is the velocity of a wheel encoder in the world (cf. \Cref{tab:frames_overview} for the offroad vehicle).
Moreover, ${}_{\Sensor_{i,k}} \Measurement_{\Sensor_{i,k} \Sensor_{i, k+1}}$ is a measurement in the sensor frame $\Sensor_{i, k+1}$ at time $k+1$ w.r.t. to the sensor frame $\Sensor_{i,k}$ at time $k$, expressed in the local sensor frame $\Sensor_{i,k}$ at time $k$; an example is the delta pose of an \ac{LR} odometry system.
These measurements do not require any reference frame alignment or dynamic variables. All local measurements until timestep $k$ are denoted as $^\Local \Measurements_k \triangleq \{ ^\Local \Measurement \}_{i \in ^\Local \mathcal{K}_{N_{\Local}}}$.\looseness=-1

\paragraph{Full Measurement Observation}
Using the measurement types defined in the previous sections, the complete measurement observation vector is then denoted as\looseness=-1
\begin{equation}
    \Measurements \doteq \{ {}^\Imumeas \Measurements, {}^\Absolute \Measurements, {}^\Landmark \Measurements, {}^\Local \Measurements \}.
\end{equation}

%% file: sections/4-Methodology.tex

\subsection{Preliminaries}
\subsubsection{\acs{MAP} Estimation}

For a set of optimization variables $\states$ and a set of provided measurements $\Measurements$, the \ac{MAP} estimation formulation is defined as
\begin{equation}
\label{equ:MAP}
    \states^\star = \argmax_{\states} p(\states|\Measurements) = \argmax p(\Measurements|  \states)p(\state_0).
\end{equation}
By iteratively using Bayes' rule (illustrated in \Cref{equ:MAP} for $\state_0$) within the optimization horizon, the joint probability distribution can be fully factorized into the likelihood and prior terms.
The solution $\states^\star$ maximizing the joint probability distribution in \Cref{equ:MAP} minimizes its negative log-likelihood.
By assuming Gaussian noise for each of the measurements $\Measurement_i \sim \mathcal{N}(h(\state),\,\sigma^{2})$, the problem can be rewritten as a least-squares optimization with residuals $\boldsymbol{r}_{\Measurement}(h(\state))$.\looseness=-1

\subsubsection{Factor Graph-Based Estimation}
\acp{FG}~\cite{dellaert2017factor} can be seen as a convenient and visualizable formalization of such optimization problems by creating bipartite graphs consisting of variables and (measurement) constraints.
While allowing for simpler formalization and creation of the optimization problem, \acp{FG} also have the advantage that efficient incremental optimization algorithms such as ISAM and iSAM2~\cite{kaess2012isam2} have been proposed, \replace{}{enabling the online optimization of complex optimization problems on constrained hardware}.
Compared to most \ac{FG}-based estimation solutions, the graph structure of \ac{HF} is more complex, offloading some of the normally hardcoded logic and conversions to the optimization.\looseness=-1

\begin{figure*}[t]
\includegraphics[width=\textwidth]{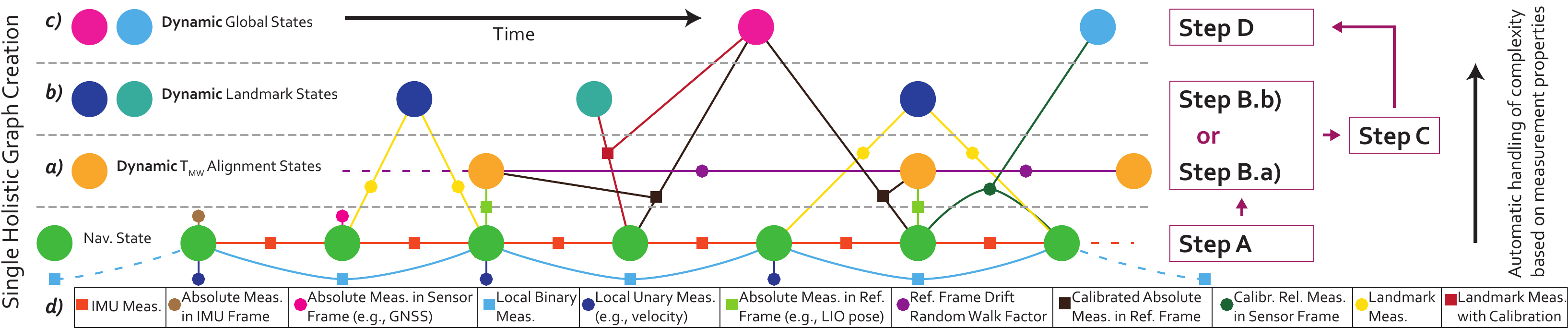}
\centering
\caption{Structural overview of the factor graph design of \ac{HF} proposed in this work. As depicted, IMU measurements drive the creation of the robot state. All other states are created dynamically (setup-specific) based on the provided measurements and framework configuration: \textit{\textbf{a)}} Reference frame alignment states. \textit{\textbf{b)}} Landmark states. \textit{\textbf{c)}} Global states. \textit{\textbf{d)}} Example measurement types depicted as factors in the graph. \textbf{Right side:} Steps from \Cref{alg:method_graph_creation}.}
\label{fig:graph_overview}
\vspace{-3ex}
\end{figure*}

\subsection{Holistic State Variables}
\label{sec:method_holistic_state_variables}
One major distinction of \ac{HF} compared to other sensor-fusion frameworks~\cite{nubert2022graph, mascaro2018gomsf, lynen2013robust} is the fact that the estimated state does not contain only fixed-size state information about the robot motion but also introduces a dynamic (setup-specific) set of context variables that are holistically optimized. 
In particular, these dynamically created states are allocated as needed and consist of \textbf{\textit{i)}} global time-invariant states, \textbf{\textit{ii)}} time-variant (i.e., drifting) or time-invariant (non-drifting) reference frame alignment states, and \textbf{\textit{iii)}} landmark states.
This design stands in contrast to existing sensor fusion formulations, which include only the robot state and IMU biases, and potentially landmark locations and calibration states, in their state vector, limiting the measurement models supported in practice to those that include these states.
The overall set of variables is given as\looseness=-1
\begin{equation}
    \mathbf{\Theta} \doteq \{ ^I \states_{N_I},  \ContextState \},~\text{with}~\ContextState \doteq \{ ^G \states_{N_G}, ^R \states_{N_R} ,^L \states_{N_L} \}.
\label{eq:state-definition}
\end{equation}
Here, $^I \states$ denotes the set of robot navigation states,  and $\ContextState$ the (dynamic) state variables consisting of global, reference frame alignment, and landmark states.
\looseness=-1

\subsubsection{Fixed-Size Robot Navigation States}
\label{sec:robot_state}
The robot navigation state variables are defined as
\begin{equation}
    ^I \states_{N_I} \dot{=} \{^I \state_i\}_{i \in ^I \mathcal{K}_{N_I}} = \{^I \state_1, \dots, ^I \state_{N_I}\},
\end{equation}
with $N_I$ being the number of navigation states.
Further, \ac{HF} follows the common navigation state definition~\cite{forster2016manifold,nubert2022graph}, given as\looseness=-1
\begin{equation}
\label{equ:imu_state}
    ^I \state_i \doteq \left[ \R_{\World \Imu, i}, \tensor[_\World] \tran{_{\World \Imu, i}}, \tensor[_\World] \vel{_{\World \Imu, i}}, \tensor[_\Imu] \bias{_i^g}, \tensor[_\Imu] \bias{_i^a} \right] \in \SOthree \times \Real^{12},
\end{equation}
which describes the motion of the \ac{IMU} sensor in $\World$. Here, $\R \in \SOthree$ defines the orientation, $\tran \in \Real^{3}$ the position, $\vel \in \Real^{3}$ the linear velocity, and $\bg \in \Real^{3}$ and $\ba \in \Real^{3}$ the gyroscope and accelerometer IMU biases expressed in $\Imu$, respectively.
Note that $\World$ is the world frame as defined in \Cref{sec:core_coordinate_frames}, which is always gravity aligned. It might or might not correspond to a gravity-aligned reference frame, e.g., \acl{ENU} (ENU) in the case of directly fusing GNSS measurements as absolute measurements. Since this estimate can jump as new knowledge arrives, the odom frame $\Odom$, as also introduced in \Cref{sec:core_coordinate_frames} and further detailed in \Cref{sec:method_smoothness}, is its other gravity-aligned counterpart, consistently delivering smooth estimates.\looseness=-1

\subsubsection{Dynamic States}
A set of dynamic states exists along with the constant-size robot navigation state. These dynamic states are more difficult to handle in practice, as their number is unknown beforehand. The number and types of dynamic states depend on the specific setup, the number of sensors used, the update frequency, and the problem formulation. All of these aspects are assumed to be unknown beforehand. An overview of the structure of dynamic states is given in \Cref{fig:graph_overview}, generalizing the example presented in \Cref{fig:holistic_fusion_illustration}.\looseness=-1

\paragraph{Dynamic Global Time-Invariant States}
In this category, each state is assumed to be \textit{global}, i.e., available and constant over the entire time horizon.
The full set of global states is\looseness=-1
\begin{equation}
    ^G \states_{N_G} \dot{=} \{^G \state_i\}_{i \in ^G \mathcal{K}_{N_G}} = \{^G \state_1, \dots, ^G \state_{N_G}\},
\end{equation}
with $N_G$ the previously unknown number of global dynamic states (usually relatively small in practice).
These states can be helpful when modeling the context of other optimization variables. In this work, they correspond mainly to sensor calibration variables, which can also vary in type, e.g., $^G \state \in \SEthree$ for pose-measuring sensors (e.g., \acp{LiDAR}) and $^G \state \in \Real^{3}$ for position-measuring sensors.
Assuming constant calibrations over the estimation horizon is sufficient in this work, as no tight visual-feature fusion is included in the experiments, which would require accurate sub-millimeter and sub-degree calibration (e.g., due to vibration or temperature). In future work, calibrations and other global factors could also support an evolution model as part of the reference frame alignment.

\paragraph{Dynamic Reference-Frame Alignment States}

A central component of \ac{HF} is the set of \textit{reference-frame alignment} states, which align measurements with the robot state $^I \state_j$, considering reference frames $\Reference_i$.
The set of alignment states is given as\looseness=-1
\begin{equation}
\label{equ:reference_frame_alignment_state}
    ^R \states_{N_R} \dot{=} \{^R \state_i\}_{i \in ^R \mathcal{K}_{N_R}} = \{^R \state_1, \dots, ^R \state_{N_R}\},
\end{equation}
where $N_R$ is the number of reference-frame state variables. Note that in the most general case, $N_R \neq N_{F_R}$ (cf. \Cref{sec:ref_frames}), i.e., the number of optimization variables that represent the reference frames is usually higher ($N_R \geq N_{F_R}$) than the number of reference frames.
\replace{In this work, all alignments are described through rigid transformations, i.e., $^R \state_i \in \SEthree~\forall~i$.}{In this work, all reference-frame alignment states are rigid transformations from the world frame $\World$ to the respective \replace{}{\textit{i)} discretized} reference frame $\Reference_{i,k}$, i.e.,
$^R \state_{j} \doteq \T_{\World \Reference_{i,k}} \in \SEthree~\forall~i,$ (cf. \Cref{sec:method_drift_modeling}) or \replace{}{\textit{ii)} discretized} keyframe $\Keyframe_{i,k}$, i.e., $^R \state_{j} \doteq \T_{\World \Keyframe_{i,k}} \in \SEthree~\forall~i$ (for better numerical stability, cf. \Cref{sec:local_keyframe_alignment}).}
\looseness=-1

\paragraph{Dynamic Landmark States}

Finally, \textit{landmark} states $^\Landmark \Measurement$ are introduced to the optimization as:\looseness=-1
\begin{equation}
    ^L \states_{N_L} \dot{=} \{^L \state_i\}_{i \in ^L \mathcal{K}_{N_L}} = \{^L \state_1, \dots, ^L \state_{N_L}\}.
\end{equation}
Here, $N_L$ is precisely the number of feature landmark coordinates (cf. \Cref{sec:meas_landmark}); a new state variable is created not for every landmark measurement but for any new feature location.\looseness=-1

\subsection{Holistic Fusion}
\label{sec:method_holistic_fusion}

\subsubsection{\acs{MAP} Estimation for Holistic Fusion}
\ac{HF} allows one to add measurements expressed in different reference frames and estimates the additional state variables required.
While for filtering-based solutions, only the last state of the \ac{MC} is estimated through recursive updates, for \ac{FG}-based approaches, all states in the time window are estimated via optimization-based smoothing.
In our \ac{HF} formulation, this can be written as a \ac{MAP} estimation (\Cref{equ:MAP}):\looseness=-1
\begin{equation}
    \mathbf{\Theta}^\star = \argmax_{\mathbf{\Theta}} p(^I \states, ^L \states, ^G \states, ^R \states | \Measurements).
\end{equation}
A \ac{FG} factorization is used~\cite{dellaert2017factor} for simplifying the prior expression and solving it using \ac{LS} by minimizing the negative log-likelihood of the Gaussian error distributions.\replace{}{ In practice, solving this \ac{MAP} problem amounts to minimizing a sum of squared residuals $\residual(\Measurement,\Measfunction(\state))$, where $\Measurement$ denotes the measurement and $\Measfunction(\state)$ the corresponding measurement function. In the following, we introduce a structured algorithmic approach for constructing $\Measfunction(\state)$ as a general template; the explicit residual definitions and factor implementations for each measurement type are provided in \Cref{sec:method_implemented_measurement_types}.}\looseness=-1

\subsubsection{Graph Structure and Graph Creation}
One of the main contributions of \ac{HF} is its structured approach to creating the underlying \ac{FG} to account for all the objectives introduced before. 
While \Cref{fig:holistic_fusion_illustration} is an illustration for a specific use case, \Cref{fig:graph_overview} provides a more generic overview of the underlying graph structure and the steps required for the graph creation.

\paragraph{Hierarchical Graph Structure}
Typical \ac{FG} instances have a central stream of robot navigation states (green states in \Cref{fig:graph_overview}), which are connected through binary (often \ac{IMU}) measurements and are constrained by additional binary (e.g., odometry) or unary (e.g., 6D pose) measurements. In most existing systems, only a single \textit{global} frame is chosen, often coinciding with the reference frame of \ac{GNSS} and barometer measurements as done, e.g., in SuperOdometry~\cite{superOdometry}. While visual and \ac{LR} estimates are also fused in, they are added only as additional binary odometry measurements in the central \ac{FG}~\cite{superOdometry}.
While enabling the fusion of multiple measurements, this method delivers reliable local odometry estimates but lacks the functionality of \textit{localizing} the robot in each of the individual reference frames.
Localization might be needed to perform tasks in frames that are not $\World$, such as path following or traversability estimation in a local map frame.
Earlier works circumvented the problem by introducing explicit formulations for context-dependent \textit{localization} scenarios (e.g., indoors vs.\ outdoors), e.g., by using a dual-graph formulation, introducing two reference frames~\cite{nubert2022graph} for localization and state estimation in environments with and without GNSS. 
In contrast, \ac{HF} allows for the direct fusion of sensor measurements from different reference coordinate frames, requiring a unique graph structure.\looseness=-1

A simplified version of the deployed graph structure is shown in \Cref{fig:graph_overview}. In addition to the main state thread, three different layers of dynamic states are introduced to allow for the fusion of arbitrary factors (including landmarks) measured in arbitrary sensor frames w.r.t. to arbitrary reference frames while allowing for extrinsic calibration:
The \textbf{\textit{Dynamic Alignment States}} layer is at the core of \ac{HF}, as it consists of dynamically allocated transformation variables that align the various reference frames as part of the optimization. An essential aspect of this innovation is the explicit modeling of drift, as explained in \Cref{sec:method_drift_modeling}.\looseness=-1
The \textbf{\textit{Dynamic Landmark States}} layer consists of landmark measurements created dynamically if needed, e.g., for leg odometry or feature location factors. 
The \textbf{\textit{Dynamic Global States}} represent the dynamic global variables, e.g., required for extrinsic calibration.\looseness=-1

One crucial aspect is properly handling these dynamic variables separately during online estimation and offline optimization, as discussed in \Cref{sec:method_online_se} and \Cref{sec:method_offline_se}, respectively.\looseness=-1

\paragraph{Graph Creation}
The creation of the graph is performed dynamically, depending on the properties of the added measurement.
The pseudo-code for this creation is shown in Algorithm~\ref{alg:method_graph_creation}\replace{.}{, and a visual summary of the transformation pipeline is provided in \Cref{fig:algorithm_summary}.}
The unique attribute of this structure is that it fits all three (non-\ac{IMU}) measurement types as introduced in \Cref{sec:problem_form_measurements}: \textbf{\textit{i)}} absolute measurements (of type $_{\Reference_i} \Measurement_{\Reference_i \Sensor_i}$), \textbf{\textit{ii)}} landmark measurements (${}_{\Sensor_i} \Measurement_{\Sensor_i \Feature_m}$), and \textbf{\textit{iii)}} local measurements ($_{\Sensor_i} \Measurement_{\World \Sensor_i}$).
The separation into \textbf{Steps A}, \textbf{B}, \textbf{C}, and \textbf{D} allows for an easy implementation of complicated measurement functions $\Measfunction(\state)$.
In particular, each new measurement must implement the interface functions \textcolor{purple}{\footnotesize\texttt{\seqsplit{transformFromIToS()}}}, \textcolor{purple}{\footnotesize\texttt{\seqsplit{transformFromSToScorr()}}}, and either \textcolor{purple}{\footnotesize\texttt{\seqsplit{transformStateFromWToR()}}} or \textcolor{purple}{\footnotesize\texttt{\seqsplit{transformLandmarkFromWToI()}}}.
The interface classes directly handle common functionalities, including dynamic variable allocation for $(^G \state, ^R \state, ^L \state)$.\looseness=-1

\begin{algorithm}[t]
\SetAlgoLined
\SetAlgoSkip{}
\replacebox{%
\caption{Creation of measurement function $\Measfunction(\state)$ using \textcolor{Green}{robot nav. state} and \textcolor{orange}{dynamic states}, which are optimized variables in the factor graph; the resulting transformed states are shown in black. The required functionalities for each new measurement are colored in \textcolor{purple}{dark pink}.\looseness=-1}
\begin{algorithmic}[1]
    \small
    \State \Comment{Main function for creation of h(x)}
    \State \textbf{Function} \texttt{createHolisticHx()}
    \State \hspace{0.2cm} \textbf{Step A:} Generate \ac{IMU} state in $\World$: \textcolor{Green}{$^I {_\World \state _{\World \Imu}}$}
    \State \hspace{0.4cm} \textbf{call} \texttt{getNearestImuStateInWorld()}
    \State \hspace{0.2cm} \Comment{If absolute measurement and not in world}
    \State \hspace{0.2cm} \textbf{Step B.a):} Create \textcolor{orange}{$^R \state = \T_{\World \Reference}$} and transform to $\Reference$: $^I {_\Reference \state_{\Reference \Imu}}$
    \State \hspace{0.4cm} \textbf{if} \texttt{measType == abs.} \textbf{\&\&} \texttt{refFrame != $\World$}
    \State \hspace{0.6cm} \textbf{call} \texttt{getOrCreateAlignmentState()}
    \State \hspace{0.6cm} \textbf{call} \textcolor{purple}{\texttt{transformStateFromWToR()}}
    \State \hspace{0.2cm} \Comment{If landmark measurement}
    \State \hspace{0.2cm} \textbf{Step B.b):} Create \textcolor{orange}{$^L {_\World \state_{\World \Feature}}$} and transform to $^L{_\Imu \state_{\Imu \Feature}}$
    \State \hspace{0.4cm} \textbf{if} \texttt{measType == landmark}
        \State \hspace{0.6cm} \textbf{call} \texttt{createLandmarkState()}
        \State \hspace{0.6cm} \textbf{call} \textcolor{purple}{\texttt{transformLandmarkFromWToI()}}
    \State \hspace{0.2cm} \Comment{If sensor frame not IMU}
    \State \hspace{0.2cm} \textbf{Step C:} Transform $\state$ to $\Sensor$: ${^I {_\Reference \state_{\Reference \Sensor}}}$ (\textbf{a}) or ${^L{_\Sensor \state_{\Sensor \Feature}}}$ (\textbf{b})
    \State \hspace{0.4cm} \textbf{if} \texttt{sensorFrame != $\Imu$}
        \State \hspace{0.6cm} \textbf{call} \textcolor{purple}{\texttt{transformFromIToS()}}
    \State \hspace{0.2cm} \Comment{If extrinsic calibration}
    \State \hspace{0.2cm} \textbf{Step D:} Create \textcolor{orange}{$^G {_\Sensor \state_{\Sensor \Sensor_{\text{corr.}}}}$} and transform to \\ \hspace{1.5cm} ${^I {_\Reference \state_{\Reference \Sensor_{\text{corr.}}}}}$ (\textbf{a}) or ${^L{_{\Sensor_{\text{corr.}}} \state_{\Sensor_{\text{corr.}} \Feature}}}$ (\textbf{b})
    \State \hspace{0.4cm} \textbf{if} \texttt{extrinsicCalibration == true}
        \State \hspace{0.6cm} \textbf{call} \texttt{createGlobalState()}
        \State \hspace{0.6cm} \textbf{call} \textcolor{purple}{\texttt{transformFromSToScorr()}}
\end{algorithmic}
}
\label{alg:method_graph_creation}
\end{algorithm}

\begin{figure}[t]
\vspace{-3ex}
\replacebox{%
\centering
\resizebox{\columnwidth}{!}{%
\begin{tikzpicture}[
    node distance=0.4cm and 0.6cm,
    state/.style={inner sep=2pt},
    steplbl/.style={draw, circle, inner sep=1pt, font=\scriptsize\bfseries},
    arr/.style={-{Stealth[length=2mm]}, thick},
    lbl/.style={font=\scriptsize, above, midway}
]
\node[font=\small\bfseries] (slabel) {Absolute Meas.:};
\node[state, right=0.5cm of slabel] (s1) {\textcolor{Green}{${}^I{}_\World \state_{\World \Imu}$}};
\node[steplbl, above=0.05cm of s1] {A};
\draw[arr] (s1.east) -- ++(0.8,0) node[lbl] {\scriptsize\textcolor{orange}{${}^R\state{=}\T_{\World \Reference}$}} node[state, right] (s2) {${}^I{}_\Reference \state_{\Reference \Imu}$};
\node[steplbl, above=0.05cm of s2] {B.a};
\draw[arr] (s2.east) -- ++(0.9,0) node[lbl] {\scriptsize $\T_{\Sensor\Imu}$} node[state, right] (s3) {${}^I{}_\Reference \state_{\Reference \Sensor}$};
\node[steplbl, above=0.05cm of s3] {C};
\draw[arr] (s3.east) -- ++(0.9,0) node[lbl] {\scriptsize\textcolor{orange}{${}^G{}_\Sensor \state_{\Sensor \Sensor_{\text{corr.}}}$}} node[state, right] (s4) {${}^I{}_\Reference \state_{\Reference \Sensor_{\text{corr.}}}$};
\node[steplbl, above=0.05cm of s4] {D};
\node[font=\small\bfseries, below=0.9cm of slabel] (llabel) {Landmark Meas.:};
\node[state, right=0.5cm of llabel] (l1) {\textcolor{Green}{${}^I{}_\World \state_{\World \Imu}$}};
\node[steplbl, above=0.05cm of l1] {A};
\draw[arr] (l1.east) -- ++(0.8,0) node[lbl] {\scriptsize\textcolor{orange}{${}^L{}_\World\state_{\World\Feature}$}} node[state, right] (l2) {${}^L{}_\Imu \state_{\Imu \Feature}$};
\node[steplbl, above=0.05cm of l2] {B.b};
\draw[arr] (l2.east) -- ++(0.9,0) node[lbl] {\scriptsize $\T_{\Sensor\Imu}$} node[state, right] (l3) {${}^L{}_\Sensor \state_{\Sensor \Feature}$};
\node[steplbl, above=0.05cm of l3] {C};
\draw[arr] (l3.east) -- ++(0.9,0) node[lbl] {\scriptsize\textcolor{orange}{${}^G{}_\Sensor \state_{\Sensor \Sensor_{\text{corr.}}}$}} node[state, right] (l4) {${}^L{}_{{\Sensor_{\text{corr.}}}} \state_{\Sensor_{\text{corr.}} \Feature}$};
\node[steplbl, above=0.05cm of l4] {D};
\end{tikzpicture}%
}
\caption{Visual summary of the state transformation steps in Algorithm~\ref{alg:method_graph_creation}. \textbf{Top row:} absolute measurement pipeline. \textbf{Bottom row:} landmark measurement pipeline. Circled letters denote the corresponding algorithm steps.\looseness=-1}
\label{fig:algorithm_summary}
}
\vspace{-3ex}
\end{figure}

\begin{replaceframe}
\subsubsection{Implemented Measurement Types}
\label{sec:method_implemented_measurement_types}
Following the problem formulation in \Cref{sec:problem_form_measurements}, this work supports \textbf{\textit{i)}} \ac{IMU} factors following~\cite{forster2016manifold}, \textbf{\textit{ii)}} holistic factors following \Cref{alg:method_graph_creation}, and \textbf{\textit{iii)}} standard (analytic) \ac{GTSAM} factors directly constraining the \ac{IMU} state in $\World$ (\Cref{equ:imu_state}). 
Each measurement factor defines an $n$-dimensional error residual $\residual(\Measurement,\Measfunction(\state)) \in \Real^n$ in the tangent space\replace{. Each measurement factor has to define an $n$-dimensional error residual $\residual \in \Real^n$ in the tangent space. 
This residual $\residual(\Measurement,\Measfunction(\state))$ is a function of the measurement (i.e., actual observation) $\Measurement$, and the measurement function $\Measfunction(\state)$. Here, $\Measurement$ and $\Measfunction(\state)$ can be on a manifold, but $\residual$ is always a vector in the tangent space.
Given}{, where $\Measurement$ and $\Measfunction(\state)$ can be on a manifold, but $\residual$ is always a vector in the tangent space.
Given} the measurements from \Cref{sec:problem_form_measurements} and assuming Gaussian noise, the full optimization objective is\looseness=-1
\begin{equation}
\label{equ:residual_objective}
\begin{split}
    r =
    \sum_{i \in ^\Imumeas \mathcal{K}_k} & \left( \|\ ^\Imumeas \residual_{i}\|^2_{\Sigma_{\Imumeas,i}} \right) +
    \sum_{i \in ^\Absolute \mathcal{K}_{N_{\Absolute}}} \left( \|\ ^\Absolute \residual_{i}\|^2_{\Sigma_{\Absolute,i}} \right) + \\
    & \sum_{i \in ^\Landmark \mathcal{K}_{N_{\Landmark}}} \left( \|\ ^\Landmark \residual_{i}\|^2_{\Sigma_{\Landmark,i}} \right) +
    \sum_{i \in ^\Local \mathcal{K}_{N_{\Local}}} \left( \|\ ^\Local \residual_{i}\|^2_{\Sigma_{\Local,i}} \right).
\end{split}
\end{equation}
In practice, robust norms (e.g., Huber, Cauchy, and Tukey) can easily replace the \textit{L2}-norms of \Cref{equ:residual_objective}. 
The following \replace{subsections detail}{paragraphs detail} the implemented factors used in this work.

\paragraph{IMU Measurement Factors}
\ac{HF} uses the common \ac{IMU} factor from~\cite{forster2016manifold}, where the measured \ac{IMU} acceleration and angular velocity constrain both the neighboring \ac{IMU} navigation states and biases (\Cref{equ:imu_state}).
The corresponding residual is\looseness=-1
\begin{equation}
    ^\Imumeas \residual_{i} \dot{=} \left[\residual_{\Delta\R_i}^\top, \residual_{\Delta v_i}^\top, \residual_{\Delta p_i}^\top \right],
\end{equation}
as defined in \cite[Equation (45)]{forster2016manifold}.
\replace{There is no need to express this factor as a holistic factor, as \textit{i)} the sensor frame is $\Imu$ itself, which is also the reference for all calibrations of other sensors, and \textit{ii)} the reference frame is always the global inertial frame. }{This factor does not require holistic treatment, as $\Imu$ is both the sensor frame and the calibration reference, and the reference frame is always the global inertial frame.} \looseness=-1

\paragraph{Holistic (Expression) Factors}
\replace{The holistic expression factors closely follow the strategy of \Cref{alg:method_graph_creation}, making sure that \textbf{\textit{i)}} the alignment of the reference frame, \textbf{\textit{ii)}} the potential incorporation of created landmark states, \textbf{\textit{iii)}} the (sometimes non-trivial) transformation to the sensor frame, and \textbf{\textit{iv)}} the extrinsic calibration are all implemented for the given measurement type. }{The holistic expression factors follow Algorithm~\ref{alg:method_graph_creation}, implementing \textbf{\textit{i)}} reference-frame alignment, \textbf{\textit{ii)}} landmark state incorporation, \textbf{\textit{iii)}} sensor-frame transformation, and \textbf{\textit{iv)}} extrinsic calibration for each measurement type.}
Each factor generates a measurement function $\Measfunction(\state)$ from a subset of the four corresponding state types: $(^I \state, ^L \state, ^G \state, ^R \state)$.
\replace{Note that, without loss of generality, instead of using the true reference frame location $T_{\World\Reference}$, in the following we can use the keyframe location $T_{\World \Keyframe}$ after manually subtracting the keyframe position from the measurements corresponding to that keyframe, cf. \Cref{sec:local_keyframe_alignment}. }{Without loss of generality, we use the keyframe location $T_{\World \Keyframe}$ instead of $T_{\World\Reference}$ by subtracting the keyframe position from the corresponding measurements, cf. \Cref{sec:local_keyframe_alignment}.}
The optimization variables are highlighted in the following measurement functions in \textcolor{BlueViolet}{blue-violet}. 
\replace{Details on the precise implementation of each holistic measurement factor can be found in the accompanying code. }{Implementation details are available in the accompanying code.}\footref{footnote:code}\looseness=-1

\paragraph{Absolute $\SEthree$ Pose Factor}
\replace{This factor allows the integration of absolute pose factors expressed w.r.t. to a reference frame.}{This factor integrates absolute pose measurements w.r.t. a reference frame. }
Using the random-walk modeling of the reference-frame alignment in \Cref{equ:method_ref_frame_random_walk_gaussian} and \Cref{equ:method_ref_frame_random_walk_keyframe_between_mean}, the corresponding measurement function $\mathbf{h}(\state)$ is given as:
\begin{equation}
\label{equ:method_absolute_pose_factor}
    \Measfunction(\state) = 
    \Tilde{\T}_{\Reference \Sensor_{\text{corr}}}(\state) =
    \overbrace{\textcolor{BlueViolet}{\T_{\Reference \World}}}^{\text{Step\,B.a)}}
    \overbrace{\textcolor{BlueViolet}{\T_{\World \Imu}}}^{\text{Step\,A}}
    \overbrace{\T_{\Imu \Sensor}}^{\text{Step\,C}}
    \overbrace{\textcolor{BlueViolet}{\T_{\Sensor \Sensor_{\text{corr}}}}}^{\text{Step\,D}} \in \SEthree,
\end{equation}
\replace{which corresponds to the general form of absolute measurements as defined in \Cref{sec:absolute_measurements}. Here, $\Reference$ corresponds to an arbitrary reference frame, and $\Sensor_{\text{corr}}$ corresponds to the true (corrected) but often unknown sensor location (in contrast to the modeled sensor frame $\Sensor$). }{following the general form from \Cref{sec:absolute_measurements}, where $\Reference$ is an arbitrary reference frame and $\Sensor_{\text{corr}}$ the true (corrected) sensor location.}
The residual is defined as $^{\Absolute} \residual = (\log(\Measfunction(\state)^{-1} \Measurement))_\vee \in \Real^6$.
Here, $\log(\cdot)$ is the logarithm map: $\SEthree \rightarrow \sethree$, and $[\cdot]_\vee$ is the projection from the Lie algebra to the tangent space $\sethree \rightarrow \Real^{6}$. 
\replace{Important components of the factor creation not shown in \Cref{equ:method_absolute_pose_factor} are \textbf{\textit{i)}} the creation of the dynamic variables, \textbf{\textit{ii)}} sensible initialization of the variable values, and \textbf{\textit{iii)}} the random walk implementation in the \ac{FG}. }{Key components not shown in \Cref{equ:method_absolute_pose_factor} include dynamic variable creation, sensible initialization, and the random walk implementation.}\looseness=-1

\begin{remark}
\label{rem:ref_frame_alignment}
\replace{The reference frame alignment is of use not only for the case where multiple absolute measurements expressed in different reference frames are fused, but even in the case where a single measurement is present but not gravity aligned. For example, consider the case of fusing the output of a non-gravity aligned localization system (e.g., Open3d-SLAM~\cite{jelavic2022open3d}) with IMU measurements, which would not work reliably if one cannot align the non-gravity-aligned reference frame with the gravity-aligned world frame $\World$ (needed for the \ac{IMU}). }{Reference-frame alignment is useful not only when fusing measurements from multiple reference frames, but also for a single non-gravity-aligned source. For example, fusing a non-gravity-aligned localization system (e.g., Open3d-SLAM~\cite{jelavic2022open3d}) with \ac{IMU} measurements requires aligning its reference frame with the gravity-aligned world frame~$\World$.}\looseness=-1
\end{remark}

\paragraph{Absolute Position Factor}
\replace{This factor allows absolute position measurements to be added relative to an arbitrary reference frame. 
Examples in this work include (single or dual) \ac{GNSS} measurements. }{This factor integrates absolute position measurements relative to an arbitrary reference frame, e.g., (single or dual) \ac{GNSS} measurements.}
The factor is defined as\looseness=-1
\begin{equation}
\label{equ:method_absolute_position_factor}
\begin{split}
    \Measfunction(\state) =
    _\Reference & \Tilde{\transl}_{\Reference \Sensor_{\text{corr}}} (\state)
    =
    \overbrace{\textcolor{BlueViolet}{\T_{\Reference \World}}}^{\text{Step\,B.a)}}
    \overbrace{\textcolor{BlueViolet}{_\World \transl _{\World \Imu}}}^{\text{Step\,A}} + \\
    & \overbrace{\textcolor{BlueViolet}{\R_{\Reference \World}}}^{\text{Step\,B.a)}}
    \overbrace{\textcolor{BlueViolet}{\R_{\World \Imu}}}^{\text{Step\,A}}
    \left(
    \overbrace{_\Imu \transl_{\Imu \Sensor}}^{\text{Step\,C}} +
    \overbrace{\R_{\Imu \Sensor}~\textcolor{BlueViolet}{_\Sensor \transl_{\Sensor \Sensor_{\text{corr}}}}}^{\text{Step\,D}}
    \right)
    \in \Real^3,
\end{split}
\end{equation}
with residual $^{\Absolute} \residual = \Measfunction(\state) - \Measurement \in \Real^3$

\paragraph{3D Landmark Factor}
\label{sec:local_landmark}
\replace{This factor allows the addition of three-dimensional position landmark measurements, located in $\World$ but measured in and to $\Base$.
Examples are measured foothold positions of a walking robot or features measured by a \ac{LiDAR} sensor. }{This factor integrates 3D position landmark measurements in $\World$, measured in $\Base$, e.g., foothold positions or \ac{LiDAR} features.}
The measurement function is defined as\looseness=-1
\begin{equation}
\label{equ:method_landmark_factor}
    \Measfunction(\state) = 
    {}_{\Sensor_{\text{corr}}} \Tilde{\transl}_{\Sensor_{\text{corr}} \Feature}(\state) =
    \overbrace{\textcolor{BlueViolet}{\T_{\Sensor \Sensor_{\text{corr}}}^{\textcolor{black}{-1}}}}^{\text{Step\,D}}
    \overbrace{\T_{\Sensor \Imu}}^{\text{Step\,C}}
    \overbrace{\textcolor{BlueViolet}{\T_{\World \Imu}^{\textcolor{black}{-1}}}}^{\text{Step\,A}}
    \overbrace{\textcolor{BlueViolet}{_\World \transl_{\World \Feature}}}^{\text{Step\,B.b)}}
    \in \Real^3,
\end{equation}
and the 3D residual for each landmark is defined as before.\looseness=-1

\paragraph{Local Linear Velocity Factor}
\replace{This factor allows the integration of absolute local velocities of arbitrary sensor frames $\Sensor_i$ in $\World$.
An example is the wheel velocity, which (no-slip assumption) is always perpendicular to the wheel axis, or \ac{RADAR} as shown in~\cite{nissov2024robust}. }{This factor integrates local velocity measurements of arbitrary sensor frames $\Sensor_i$ in $\World$, e.g., wheel odometry (no-slip assumption) or \ac{RADAR}~\cite{nissov2024robust}.}
The corresponding measurement model is:\looseness=-1
\begin{equation}
\label{equ:method_linear_velocity_factor}
\begin{split}
    \Measfunction(\state) = 
    _{\Sensor_{\text{corr}}} & \Tilde{\vel}_{\World \Sensor_{\text{corr}}} (\state) = 
    \overbrace{\textcolor{BlueViolet}{\R_{\Sensor \Sensor_{\text{corr}}}^{\textcolor{black}{-1}}}}^{\text{Step\,D}}
    \left( 
    \overbrace{\R_{\Sensor \Imu}}^{\text{Step\,C}}
    \left(
    \overbrace{\textcolor{BlueViolet}{\R_{\World \Imu}^{\textcolor{black}{-1}} {_\World \vel_{\World \Imu}}}}^{\text{Step\,A}}
    \right.
    \right.
    \\
    &
    \left.
    \left.
    + 
    \overbrace{_\Imu \rotvel_{\World \Imu} \times _\Imu \transl_{\Imu \Sensor}}^{\text{Step\,C}}
    \right) +
    \overbrace{_\Sensor \rotvel_{\World \Sensor} \times \textcolor{BlueViolet}{_\Sensor \transl_{\Sensor \Sensor_{\text{corr}}}}}^{\text{Step\,D}}
    \right) \in \Real^3,
\end{split}
\end{equation}
with a simple 3D residual in the vector space as before.

\paragraph{Standard \acs{IMU}-Constraining Factors}
\ac{HF} also allows the integration of regular (non-holistic) factors as proposed in other works. Note that by default, these factors do not support calibration or alignment of the reference frames. 
\replace{Moreover, our implementation assumes that the measurement is already converted to the $\Imu$ frame, which, e.g., for a position measurement of an arbitrary sensor frame, is problematic, as the orientation in $\World$ should also be included in the factor to constrain the robot's orientation properly (the lever arm between $\Imu$ and $\Sensor$ can help to constrain the yaw angle). }{Moreover, these factors assume measurements are already in the $\Imu$ frame, which is problematic for position measurements of other sensor frames, as the lever arm between $\Imu$ and $\Sensor$ helps constrain orientation.}\looseness=-1

The following factors are used in this work as a reference/baseline. Details can be found in the accompanying code.\footref{footnote:code}

\paragraph{Relative $\SEthree$ Pose Between Factor}
\replace{This is the typical pose between factors used in most existing \ac{SLAM} or \ac{SF} solutions. }{This is the standard pose-between factor used in most \ac{SLAM} and \ac{SF} systems.}
The measurement function is\looseness=-1
\begin{equation}
    \Measfunction(\state) = \textcolor{BlueViolet}{\T_{\World \Imu_k}^{\textcolor{black}{-1}} \T_{\World \Imu_{k+1}}} \in \SEthree, 
\end{equation}
as available in the \ac{GTSAM} library~\cite{dellaert2017factor}.

\paragraph{Absolute $\SEthree$ Pose Factor}
This is the typical $\SEthree$ prior factor as available in \ac{GTSAM}:$\Measfunction(\state) = \textcolor{BlueViolet}{\T_{\World \Imu}}$.
\end{replaceframe}

\subsubsection{Reference-Frame Drift Modeling}
\label{sec:method_drift_modeling}
The alignment of two trajectories, represented as either $\SEthree$ poses or $\Real^3$ positions, is commonly done in the robotics literature, most commonly in the form of \textit{Umeyama} alignment\cite{umeyama1991least}.
This method estimates a single rigid transformation $\T \in \SEthree$, and optionally a scale parameter $\mathbf{s}$.
While this technique is valuable for comparing two trajectories, as done in the assessment of trajectory quality (cf. \Cref{sec:experimental_results}) or alignment of two non-drifting measured trajectories of the \textit{same} sensor frame, for sensor fusion it is not suitable, primarily due to drift; an example of \textit{Umeyama} alignment for this work's field deployments is shown in \Cref{fig:exp_hike_alignment}, where no single rigid transformation exists that aligns the two trajectories.
In response, this work introduces random-walk modeling of the present reference frames, as shown in \Cref{fig:graph_overview}. 
Modeling the reference frame evolution as a random walk is inspired by previous work in estimation, which successfully modeled quantities such as \ac{IMU} biases as a random walk. This model is simple to implement and successful. Inherently, the random walk formulation depends primarily on time; i.e., the longer the duration, the more the variable (in this case, the reference frame transformation) is allowed to change.
An alternative could be a distance-based noise model, which, however, would be significantly harder to implement.
Introduced in \Cref{equ:reference_frame_alignment_state}, each reference-frame alignment is represented as a transformation $\T_{\World \Reference_{i,k}} \in \SEthree$ for reference frame $i$. Here, $k$ denotes the sample time of the reference frame.\looseness=-1 

\paragraph{Change of $\SEthree$}
The derivatives of $\SEthree$ transformations are expressed using a screw theory formulation: 
\begin{equation}
    \dot{\T}_{\World \Reference_i} = \T_{\World \Reference_i} [ \sethreetangent_i]_\wedge,
    \quad\text{with}~\T = 
    \begin{bmatrix}
    \R & \tran \\
    \mathbf{0}_3^\top & 1
    \end{bmatrix}.
\end{equation}
Here, $\sethreetangent_i = [\rotvel_i^\top, \vel_i^\top]^\top \in \Real^6$ is the local group velocity vector of the reference frame $\Reference_i$. The \textit{wedge} operator $\wedge$ describes the map from the tangent space to the corresponding Lie algebra: $[\cdot]_\wedge \colon \Real^6 \rightarrow \sethree$. 
For small changes over a time interval $\Delta t$ at timestep $k$, this can be propagated to the manifold as:\looseness=-1
\begin{equation}
\label{equ:dt_random_walk}
    \T_{\World \Reference_{i,k+1}} \approx \T_{\World \Reference_{i, k}} \Delta \T_{\sethreetangent_{i}}, ~\text{with}~ \Delta \T_{\sethreetangent_{i}} = \text{exp} ([\Delta t \sethreetangent_{i}]_\wedge).
\end{equation}
Here, $\text{exp}()$ is the exponential map, which for $\SEthree$ can be computed efficiently using the manifold retraction.\looseness=-1

\paragraph{\ac{DT} Random Walk}
\ac{HF} models the reference frame evolution as a multivariate \ac{DT} random walk, i.e.,\looseness=-1
\begin{equation}
\label{}
    \sethreetangent_{i} \sim \mathcal{N}(\mathbf{0},\mathbf{\Sigma}_i^{2}), ~ \text{with}~\text{diag}(\mathbf{\Sigma}_i) = [\sigma_{i, 1}, \dots, \sigma_{i, 6}]^\top.
\end{equation}
While for most realistic measurements $\sigma_{i,j} > 0$ to properly consider the drift of real-world measurements, $\sigma_{i,j} = 0$ can be used for non-drifting measurements or directions. 
In the \ac{FG}, this constraint can be added as a  zero-mean $\SEthree$ \textit{between} factor.\looseness-1

\begin{remark}
\label{rem:loop_closures}
As shown in \Cref{equ:dt_random_walk}, the evolution of the reference frame is assumed to continuously drift over time without large jumps from one keyframe to the other. Jumps such as loop closures can, however, still be handled in the \ac{HF} formulation by simply adding an external ``correction'' measurement in $\World$ or another static/slowly drifting reference frame. The optimization will then move the drifting map frame (e.g., from the \ac{LR} registration) to align with this correction, for both the online marginalization and offline bundle adjustment.\looseness-1
\end{remark}

\begin{remark}
\label{rem:random_walk_tuning}
In practice the random walk should be set as loosely as necessary (to account for the occurring drift), but as strictly as possible, to serve as a valid motion prior needed for some estimation problems, e.g., when either no global measurements are available at times, or when specific motions are required over a time window to render the estimation problem observable, e.g., when fusing only position measurements.\looseness=-1
\end{remark}

\subsubsection{Local Keyframe-Based Reference-Frame Alignment}
\label{sec:local_keyframe_alignment}
\Cref{equ:dt_random_walk} models the evolution of the true reference frame w.r.t. the $\World$ frame.
While it is important to analyze the actual drift occurring for each measurement, modeling the optimization variable as $\T_{\World \Reference_{i, k}}$ introduces numerical issues when traveling large distances. With increasing distance, the rotation lever arm gets larger, leading to proportionally increased sensitivity w.r.t. to the origin distance. 
As a result, in \ac{HF}, a new keyframe is generated whenever a new reference-frame alignment variable is created (after $\Delta t$). The measurements associated with a reference frame are then manually transformed to this new keyframe location \replace{$\Keyframe$}{$\Keyframe_{i, k+1}$} via \replace{$\T_{\Keyframe_{i, k+1}}\Reference_i$}{$\T_{\Keyframe_{i, k+1} \Reference_i}$}, allowing for local adaptation along the trajectory in case of occurring drift. 
Moreover, the resulting transformation between the previous reference alignment frame remains as uncertain as before but gets a non-zero mean in position:\looseness=-1
\begin{equation}
\label{equ:method_ref_frame_random_walk_gaussian}
    \sethreetangent_{i} \sim \mathcal{N}(\mathbf{\sethreetangent_{i, \mu}}, \mathbf{\Sigma}_i^{2}), ~ \text{with}~\text{diag}(\Sigma_i) = [\sigma_{i, 1}, \dots, \sigma_{i, 6}]^\top.
\end{equation}
Here, $\text{exp}([\sethreetangent_{i, \mu}]_\wedge) = \T_{\Keyframe_{i, k} \Keyframe_{i, k+1}}$ is the known shift in keyframe location solely based on the corresponding measurements (independent of the optimization output) with zero rotation, i.e.,\looseness=-1 
\begin{equation}
\label{equ:method_ref_frame_random_walk_keyframe_between_mean}
    \T_{\Keyframe_{i, k} \Keyframe_{i, k+1}} = 
    \begin{bmatrix}
    \mathbf{I} & {}_{\Keyframe_{i,k}} \transl_{\Keyframe_{i,k} \Keyframe_{i,k+1}} \\
    \mathbf{0}_3^\top & 1
    \end{bmatrix}.
\end{equation}
The known keyframe position expressed in the corresponding reference frame ${}_{\Reference_i} \transl_{\Reference_i \Keyframe_{i,k+1}}$ is subtracted from the current measurements $\Measurement$ to allow for local alignment (with a small lever arm).
The optimization variable in this new, numerically more stable setting is then $\T_{\World \Keyframe_{i,k}}$ with non-zero expected value in position for the $k$-th created alignment variable.
Finally, to backtrace the overall reference frame drift, this estimate can be transformed to $\T_{\World \Reference_{i,k}}$ by subtracting the keyframe position:\looseness=-1
\begin{equation}
    \T_{\World \Reference_{i,k}} = \T_{\World \Keyframe_{i,k}} \T_{\Keyframe_{i,k} \Reference_{i,k}}.
\end{equation}
This parametrization as a new keyframe is not just a design choice but a crucial component to make this automatic alignment work in practice, as shown later in \Cref{sec:exp_local_keyframes}.\looseness=-1


\subsubsection{Automatic Calibration}
\label{sec:method_calibration}
While not the main focus of this work, \ac{HF} also supports automatic extrinsic calibration. Each calibration variable is assumed to be constant throughout the horizon and is modeled as a global variable $^G \state_i$ (e.g., $\in \SEthree$).\looseness=-1

\subsubsection{Uncertainty in \acs{HF}}
A notion of uncertainty is essential for online and offline operations in practice. 
\ac{HF} computes the uncertainty of all variables (robot state and dynamic variables) using marginal covariances.
The adjoint map $\adj_{\World \Imu}$ of $\T_{\World \Imu}$ is used to map these covariances from the tangent space ${{}_\Imu \Marginal}$ to $\World$:\looseness=-1
\begin{equation}
    _\World \Marginal = \adj_{\World \Imu} ~ {{}_\Imu \Marginal} ~ \adj_{\World \Imu}^\top.
\end{equation}

\subsection{Online State Estimation}
\label{sec:method_online_se}

\replace{Performing online estimation is significantly more challenging than offline, primarily due to delayed and out-of-order measurements and because the total estimation of the latest timestamp $k$ must be causal. }{Online estimation is more challenging than offline due to delayed and out-of-order measurements and the causality requirement at the latest timestamp $k$.} 
\looseness=-1

\pdfpxdimen=\dimexpr 1 in/72\relax
\begin{figure}[t]
\includegraphics[width=\columnwidth]{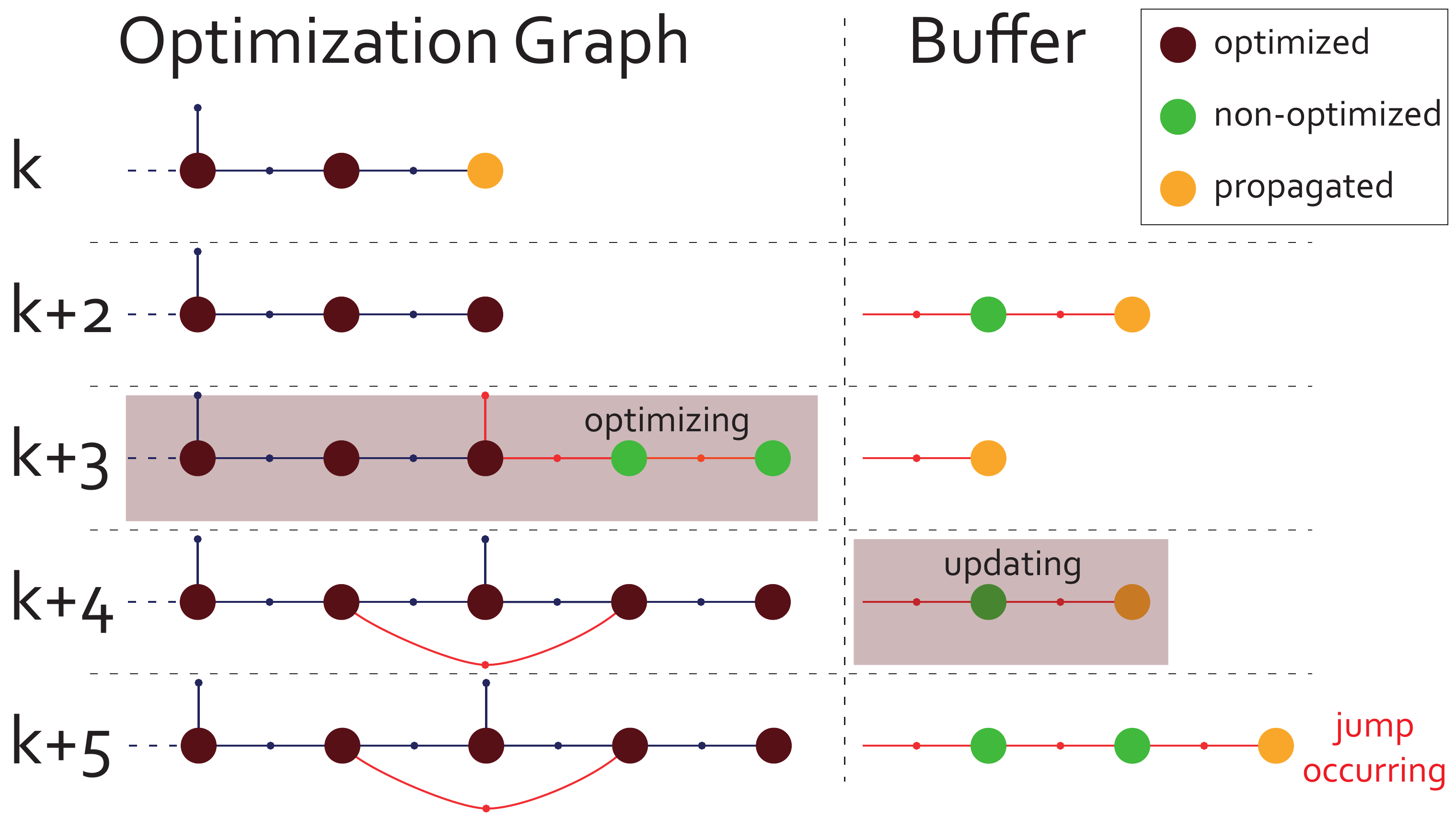}
\centering
\caption{Illustration of the high-rate state propagation and asynchronous optimization-based measurement updates, similar to the scheme in~\cite{nubert2022graph}.}
\vspace{-3ex}
\label{fig:online_se_high_rate}
\end{figure}

\subsubsection{High Estimation Rates in Presence of Delays}
\label{sec:online_se_high_rates}
Building on the high-rate prediction–update framework of~\cite{nubert2022graph}, our approach focuses on providing accurate state estimates at the latest timestamp, which is essential for closed-loop control, high-rate mapping, undistortion, and path planning. \ac{HF} runs \ac{IMU}-based state propagation at full rate and performs slower fixed-window optimization when new non-\ac{IMU} measurements arrive, similar in spirit to \ac{MHE} but using full \ac{MAP} estimation with probabilistic marginalization. As illustrated in \Cref{fig:online_se_high_rate}, nonlinear optimization runs in the background while measurements are integrated asynchronously, and the latest belief is always re-propagated using buffered \ac{IMU} data.\looseness=-1

\subsubsection{Out-of-Order Measurements}
\label{sec:online_se_out_of_order}
Most modern \ac{FG} methods create states at low rate and rely on \ac{IMU} pre-integration~\cite{forster2016manifold}, which complicates \ac{RT} operation when measurements arrive out of order, since rewiring the graph and inserting new timestamped states becomes costly. \ac{HF} avoids this issue by maintaining dense states and performing \ac{IMU} pre-integration only when no further measurements are expected, which is feasible given the short \ac{RT} horizon and enables reliable performance on laptop-grade \acp{CPU}. Future improvements could apply \ac{LI} near measurement timestamps to reduce the required state rate, or transition to a \ac{CT} backend using \aclp{GP} or spline-based representations~\cite{talbot2024continuous}.
\looseness=-1

\subsubsection{Odometry Estimation: Smoothness and Consistency}
\label{sec:method_smoothness}
\begin{figure}[t]
\includegraphics[width=\columnwidth]{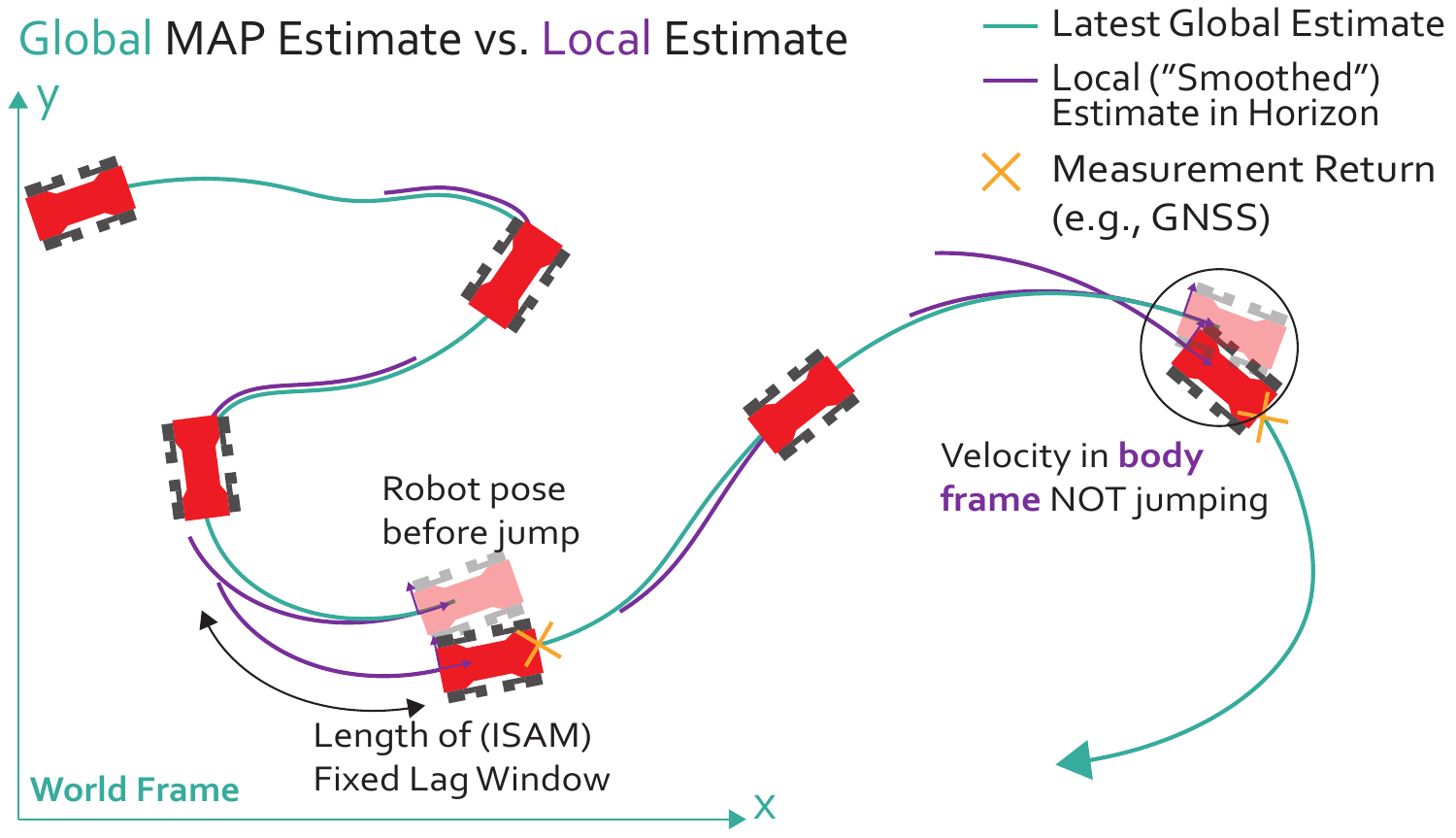}
\centering
\caption{Motivation and illustration of the smooth odometry generation by integrating the body frame state estimated in the current smoother window.}
\label{fig:method_online_se_smoothness}
\vspace{-3ex}
\end{figure}

\replace{With newly arriving information, the belief of the current robot state in $\World$ can change. Quite realistically, this can significantly affect the robot's pose and velocity, e.g., when \ac{GNSS} information returns after a long time of absence, leading to a significant jump in the robot's estimated position (and heading).
In many applications, e.g., localization and pathfinding, having this up-to-date best knowledge of the actual robot state is desirable. However, in scenarios like tracking control or point-cloud undistortion, the smoothness of the provided estimate is critical.
\Cref{fig:method_online_se_smoothness} (twice) illustrates this scenario of the robot pose experiencing a significant update during online operation. }{New information can cause significant jumps in the robot's estimated pose and velocity in $\World$, e.g., when \ac{GNSS} returns after a prolonged absence.
While such up-to-date estimates are desirable for localization, tracking control and point-cloud undistortion require smooth estimates, as illustrated in \Cref{fig:method_online_se_smoothness}.} 
\replace{To provide a smooth estimate of the robot at any time, \ac{HF} proposes using a specific strength of an optimization-based smoothing approach. While the current belief of the robot can change significantly with new information, this will consistently happen in the smoother window; none of the estimates will \textit{locally} show jumps due to \ac{IMU} and optional kinematic smoothness constraints. }{\ac{HF} exploits the optimization-based smoother: while the belief can change significantly with new information, \ac{IMU} and kinematic constraints ensure that no estimate \textit{locally} shows jumps within the smoother window.}
Hence, to avoid \textit{any} jumps, the current (pose and linear velocity) odometry belief is converted to the body frame ($\T_{\Imu_{k} \Imu_{k+1}}$), incremented there, and then mapped back to the odometry frame $\Odom$ using\looseness=-1
\begin{equation}
    \T_{\Odom \Imu_{k+1}} = \T_{\Odom \Imu_{k}} \T_{\Imu_{k} \Imu_{k+1}};\quad
    _\Odom \vel_{\Odom \Imu_{k+1}} = \R_{\Odom \Imu_{k+1}} ~ {_\Imu \vel_{\Odom \Imu_{k+1}}}.
\end{equation}
Here, the main challenge is to extract the local update $\T_{\Imu_{k} \Imu_{k+1}}$ from the optimized robot state variables $(\T_{\World \Imu},~{}_\World \vel_{\World \Imu})$.
During the prediction phase (cf. \Cref{sec:online_se_high_rates}) of $\T_{\Odom \Imu_{k}}$, this is done the same way as for the propagated state in $\World$ by performing the integration of the \ac{IMU} measurements expressed in the body frame; refer to~\cite[Equation 30]{forster2016manifold} for details.
\replace{However, the velocity and position's single- and double-integration characteristics will lead to significant drift over time. 
To address this issue, with every newly arriving update $(\T_{\World \Imu}, _\World \vel_{\World \Imu})$, $_\Imu \vel_{\Odom \Imu_{k+1}}$ is overwritten by $_\Imu \vel_{\World \Imu_{k+1}}$, which is smooth, as illustrated in \Cref{fig:method_online_se_smoothness}, to track the actual physical velocity of the robot as closely as possible, assuming \textit{zero velocity} between $\Odom$ and $\World$. }{However, single- and double-integration drift accumulates over time. To counter this effect, with every update $(\T_{\World \Imu}, _\World \vel_{\World \Imu})$, $_\Imu \vel_{\Odom \Imu_{k+1}}$ is overwritten by $_\Imu \vel_{\World \Imu_{k+1}}$ (smooth, cf. \Cref{fig:method_online_se_smoothness}), assuming \textit{zero velocity} between $\Odom$ and $\World$.}
\replace{Moreover, the roll and pitch components of $\T_{\Odom \Imu}$ are overwritten by those of $\T_{\World \Imu}$ to maintain full observability and fulfill the requirements of modules depending on gravity (e.g., a locomotion controller).
From the last optimized state, the velocity is propagated according to \Cref{fig:online_se_high_rate} to the current timestamp and then integrated in real time to obtain the new state of the robot, leading to smooth trajectories in $\T_{\Odom \Imu}$.
The jumps, smoothness, and consistency that occur in $\World$ and $\Odom$ are highlighted in \Cref{fig:exp_anymal_hike_elevation_mapping} for the ANYmal hike experiment. }{Moreover, roll and pitch of $\T_{\Odom \Imu}$ are overwritten by those of $\T_{\World \Imu}$ for gravity-dependent modules (e.g., locomotion control).
From the last optimized state, velocity is propagated per \Cref{fig:online_se_high_rate} and integrated in real time, yielding smooth trajectories in $\T_{\Odom \Imu}$ (cf. \Cref{fig:exp_anymal_hike_elevation_mapping}).}\looseness=-1
\pdfpxdimen=\dimexpr 1 in/72\relax

\subsubsection{Asynchronous Online Optimization}

\paragraph{Computational Complexity}
\replace{Optimization is performed within a time window to keep the computational complexity tractable during online operation. 
The default \ac{GTSAM}~\cite{dellaert2017factor} \ac{FL} smoother is used with iSAM2~\cite{kaess2012isam2} as an incremental solver. 
This selection enables optimization using complete marginalization of old variables, which, by default, is well-suited for odometry estimation. }{\ac{HF} uses the \ac{GTSAM}~\cite{dellaert2017factor} \ac{FL} smoother with iSAM2~\cite{kaess2012isam2}, performing windowed optimization with complete marginalization of old variables.}\looseness=-1

\paragraph{\ac{FL} Smoother Effects on Variables}
\label{sec:smoother_effects}
\replace{However, as visible in \Cref{fig:graph_overview}, many dynamic state variables in \ac{HF} are either global or (slowly) changing reference frames modeled as a random walk.
To ensure that this paradigm also works for the \ac{RT} fixed-lag smoother in case of dis- and reappearing measurements, which would normally lead to the loss of existing variables, the graph states are handled as follows (increasing complexity). }{However, many dynamic variables in \ac{HF} are global or slowly changing reference frames (cf. \Cref{fig:graph_overview}). To handle dis- and reappearing measurements in the \ac{RT} \ac{FL} smoother without losing variables, the states are managed as follows (increasing complexity).}\looseness=-1

\paragraph*{Landmark state variables}
\replace{Currently, this class of variables is the simplest, as no feature re-detection is provided. As an example, for leg odometry, this means that once the contact is broken, the foothold remains part of the (\ac{RT}) optimization until it is marginalized without special treatment. }{These are simplest, as no feature re-detection is provided; e.g., footholds remain in the graph until marginalized without special treatment.}\looseness=-1
\paragraph*{Global state variables}
\replace{This information should not be forgotten for global variables once the corresponding measurements leave the smoother time window. Hence, each global variable is stored in a memory buffer with its last estimated belief and uncertainty. Once the measurement is marginalized, the corresponding variable is labeled \textit{inactive}. If, after a while, measurements constraining this global variable return, the corresponding variable is \textit{activated} again, and a virtual \textit{prior factor} of its previous belief and uncertainty is added to the \ac{RT} graph. }{Global variables must persist beyond the smoother window. Each is stored with its last belief and uncertainty and labeled \textit{inactive} upon marginalization. When corresponding measurements reappear, the variable is \textit{activated} with a virtual \textit{prior factor} encoding its previous estimate.}\looseness=-1
\paragraph*{Reference-frame state variables}
\replace{The situation is even more complicated for reference state variables, which can change over time. Hence, a single variable to be (de)activated is insufficient. 
If a reference frame-constraining measurement reappears, there is a differentiation: }{Reference-frame variables change over time, so simple (de)activation is insufficient. When a constraining measurement reappears:}
\textbf{\textit{i)}} if the corresponding reference frame is not too old ($\leq \Delta t$ in \Cref{sec:local_keyframe_alignment}), the variable is activated again. A prior belief and uncertainty are added to the \ac{RT} graph for global variables similar to before. 
\textit{ii)} Otherwise, the last variable is reactivated and added to the \ac{RT} graph, but additionally, a new variable is created, and a random-walk factor with mean $\T_{\Keyframe_{j, k} \Keyframe_{j, k+1}}$ and scaled uncertainty is added to both the \ac{RT} graph and the offline smoother.\looseness=-1

\begin{figure}[t]
\includegraphics[width=\columnwidth]{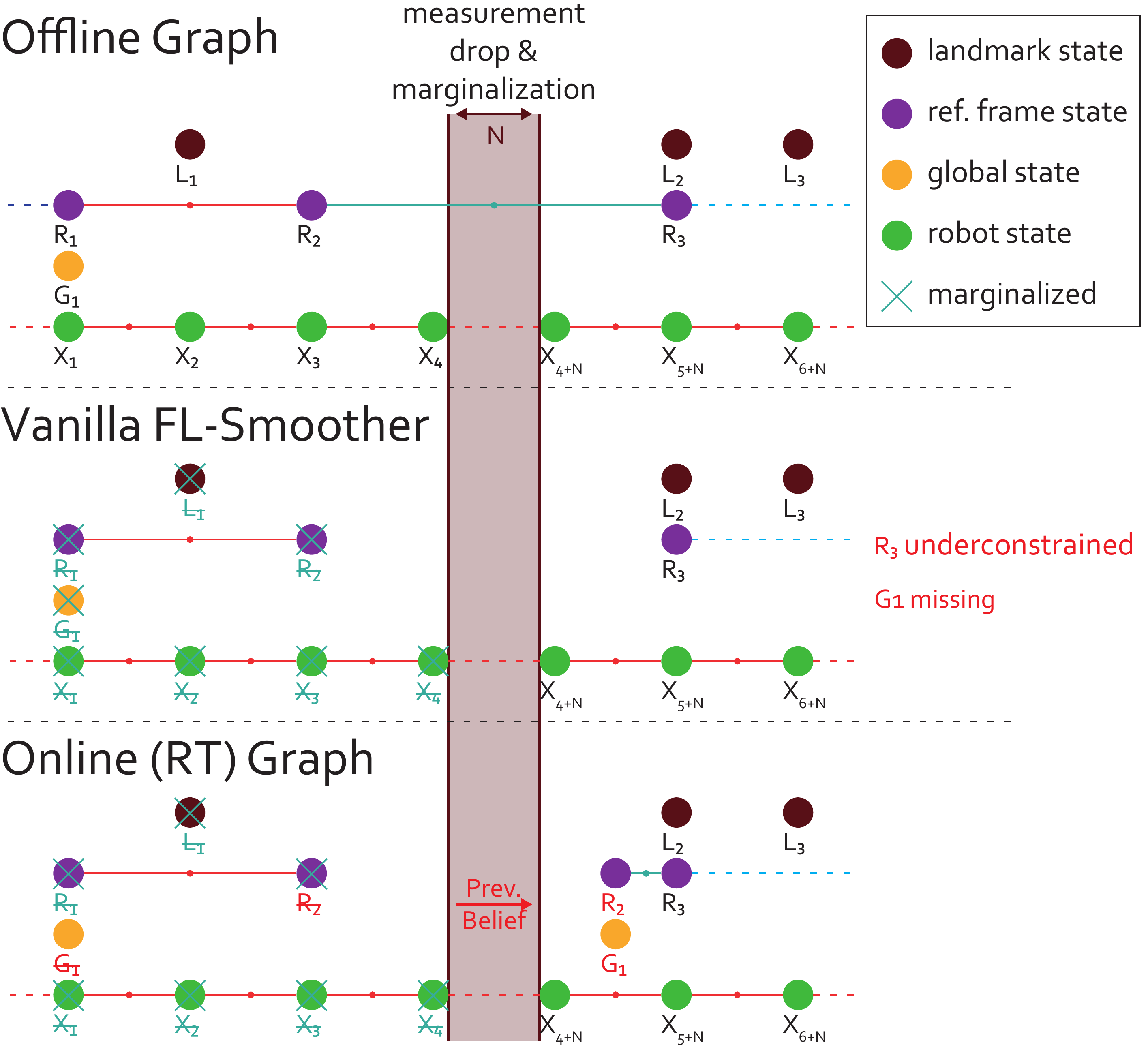}
\centering
\caption{Effect of the limited \ac{FL} window of the smoother leading to variables being marginalized out. The handling of the different dynamic state variables in \ac{HF} and the differences between its online and offline graphs are highlighted.\looseness=-1}
\label{fig:online_se_fixed_lag}
\vspace{-3ex}
\end{figure}

This process allows for seamless operation with dis- and reappearing holistic measurements and is illustrated in \Cref{fig:online_se_fixed_lag}.\looseness=-1

\subsubsection{Observability of Dynamic Variables}
\label{sec:observability}
\replace{In practice, when creating a new dynamic variable, such as for reference frame alignments, some of these variables may not be fully observable directly after creation. A prior with high uncertainty is added to the online graph to prevent the linearized system from becoming indeterminate. A similar prior factor is also added at the beginning of the mission to constrain the initial state in $\World$. }{Newly created dynamic variables (e.g., reference-frame alignments) may not be immediately observable. A high-uncertainty prior is added to prevent indeterminacy, similarly to the initial state prior in $\World$.}\looseness=-1

\subsection{Offline Batch Estimation}
\label{sec:method_offline_se}
\replace{Estimating robot states in an offline setting is much simpler than online. Beyond the holistic fusion idea introduced in \Cref{sec:method_holistic_fusion}, the main difference between \ac{HF}'s offline estimation and earlier works is the much higher number of robot states. }{Offline estimation is simpler than online. Beyond the holistic fusion of \Cref{sec:method_holistic_fusion}, \ac{HF}'s main distinction from prior work is a much higher number of robot states.}\looseness=-1

\pdfpxdimen=\dimexpr 1 in/72\relax

\subsubsection{High-Quality Initialization}
\replace{Yet, this high number of states does not introduce much computational overhead, as each variable is automatically initialized through the belief of the online graph, allowing it to converge in a few iterations despite a complicated graph structure.
The general offline graph structure is the same as the online graph (except for the additional constraints of \Cref{sec:smoother_effects}, \Cref{sec:observability}). Thus, the online estimate can be used without any modifications as an initial guess for the offline smoother.
The resulting optimization times for offline estimation are reported in \Cref{tab:exp_real_world_mission_data}. }{Despite the high state count, each variable is initialized from the online belief, enabling convergence in few iterations. The offline graph mirrors the online structure (without the additional constraints of \Cref{sec:smoother_effects}, \Cref{sec:observability}), so the online estimate directly serves as an initial guess (cf. \Cref{tab:exp_real_world_mission_data}).}\looseness=-1

\subsubsection{Easier Graph Creation and Fewer Assumptions}
\replace{The offline graph is much easier to build, as no variables are marginalized out, and hence, no new virtual prior factors are required to reintroduce the lost information.
Even more importantly, the offline graph is mathematically more tractable, as no prior factors on the state variables are required to make the problem fully observable. Instead, reference-frame alignment or calibration becomes observable without additional helper constraints. }{Without marginalization, no virtual priors are needed to reintroduce lost information. The offline graph is also mathematically simpler: reference-frame alignment and calibration become observable without auxiliary constraints.}\looseness=-1

\subsubsection{Pseudo GT Generation}
\replace{One advantage of the graph formulation of \ac{HF} is the high rate of robot states created. This design allows for an offline-optimized trajectory at (potentially) full \ac{IMU} frequency. The resulting trajectory is smooth and consistent (e.g., \Cref{fig:exp_anymal_indoor_eval} and the supplementary video).
Due to the high rate and easy accessibility of this solution, it can serve well as a fast and accessible way to create post-mission pseudo \ac{GT} trajectories, to check the online performance, or as a target for learning-based estimators. }{The high state rate yields offline-optimized trajectories at up to full \ac{IMU} frequency, producing smooth and consistent results (cf. \Cref{fig:exp_anymal_indoor_eval} and the supplementary video). This output readily serves as post-mission pseudo \ac{GT} for evaluating online performance or training learning-based estimators.}\looseness=-1

%% file: sections/5-ImplementationalDetails.tex

The implementation and open-sourcing of \ac{HF} are at the core of this work. The corresponding framework is released for the benefit of the robotics community.\footref{footnote:code}
The \ac{HF} framework was designed with flexibility and usability in mind to fulfill the needs of most real-world mobile-robot applications.
Although the framework was initially built on the dual-graph estimation of \acl{GMSF}~\cite{nubert2022graph}, \ac{HF} constitutes a complete generalization by eliminating all customized design choices that had been made for construction robots. Thus, \ac{HF} is a generic framework suitable for widely varying robotic applications and systems.\looseness=-1

\begin{figure}[t]
\includegraphics[width=\columnwidth]{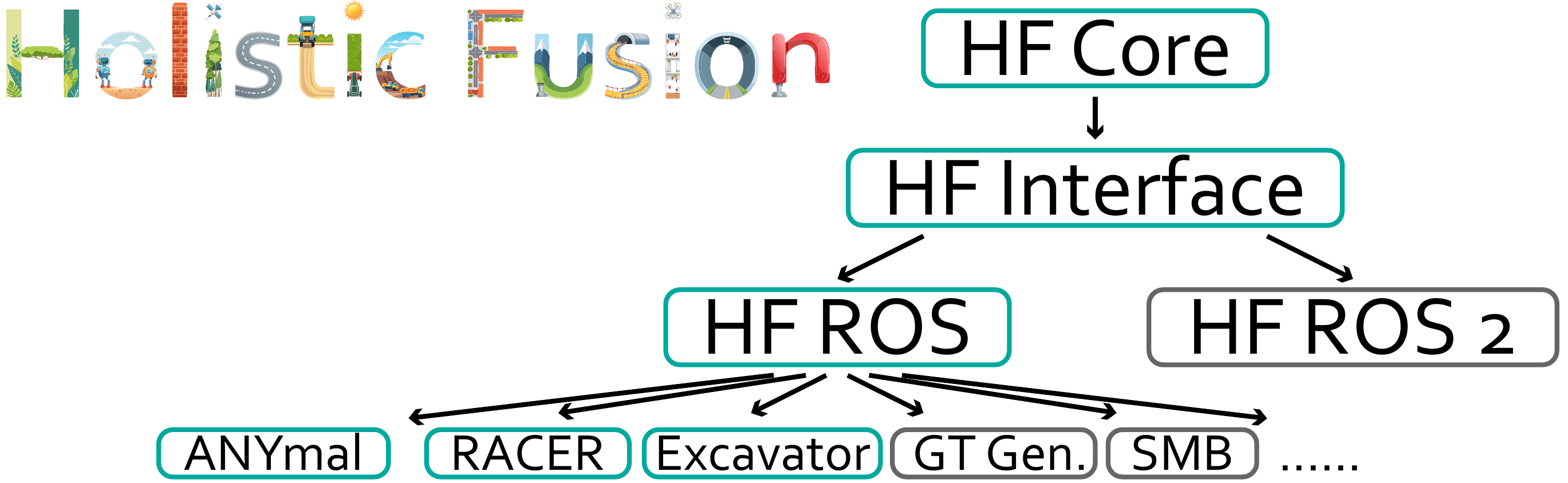}
\centering
\caption{High-level overview of the software structure and examples. HF Core is fully generic and templated, while HF Interface provides concrete implementations. HF ROS and ROS2 are corresponding middleware wrappers.\looseness=-1}
\label{fig:implementation_overview}
\vspace{-3ex}
\end{figure}

All code is written in \textit{C++} with main dependencies on the \ac{GTSAM} and \textit{Eigen} libraries. The backend implementation of \ac{HF} is independent of any robotic middleware such as \acs{ROS} or \acs{ROS}2.
To enable communication with other modules, a \acs{ROS} wrapper with additional functionalities such as callback, logging, visualization, and message advertisement is provided, which has been extensively tested on the robots shown in \Cref{fig:exp_robotic_platforms}. 
The \ac{HF} framework is documented and in-line commented, and it provides an auto-generated \textit{Doxygen}.\footnote{\label{footnote:doxygen}{\scriptsize\anontext{\url{https://leggedrobotics.github.io/holistic_fusion/doxy}}{anonymous url}}}
A \textit{Read the Docs}\footnote{\label{footnote:read_the_docs}{\scriptsize\anontext{\url{https://leggedrobotics.github.io/holistic_fusion/docs}}{anonymous url}}} is available to guide the user through the examples, the software architecture, and the main tuning parameters.
\Cref{fig:implementation_overview} provides an overview of the latest \ac{HF} software packages, with the ones evaluated in this work highlighted in turquoise.
\pdfpxdimen=\dimexpr 1 in/72\relax
More information can be found in the online documentation.\footref{footnote:read_the_docs}\looseness=-1

%% file: sections/6-ExperimentalResults.tex

This section provides experimental results, analyses, comparisons, and ablation studies on three robotic platforms.
For all platforms shown, \ac{HF} is the default localization and state-estimation solution. 
By automatically aligning all coordinate frames, \ac{HF} is a fully synchronized localization manager. 
By introducing the notion of drift, it can fuse various (absolute) measurements into a single optimization. 
The simple and fast online and offline optimization capabilities with states up to \ac{IMU} rate make the framework suitable for effective (online and) offline estimation, as well as \ac{PGT} generation.
These aspects are evaluated qualitatively and quantitatively in the following subsections.\looseness=-1

\subsection{Robotic Platforms}
\begin{figure}[t]
\includegraphics[width=\columnwidth]{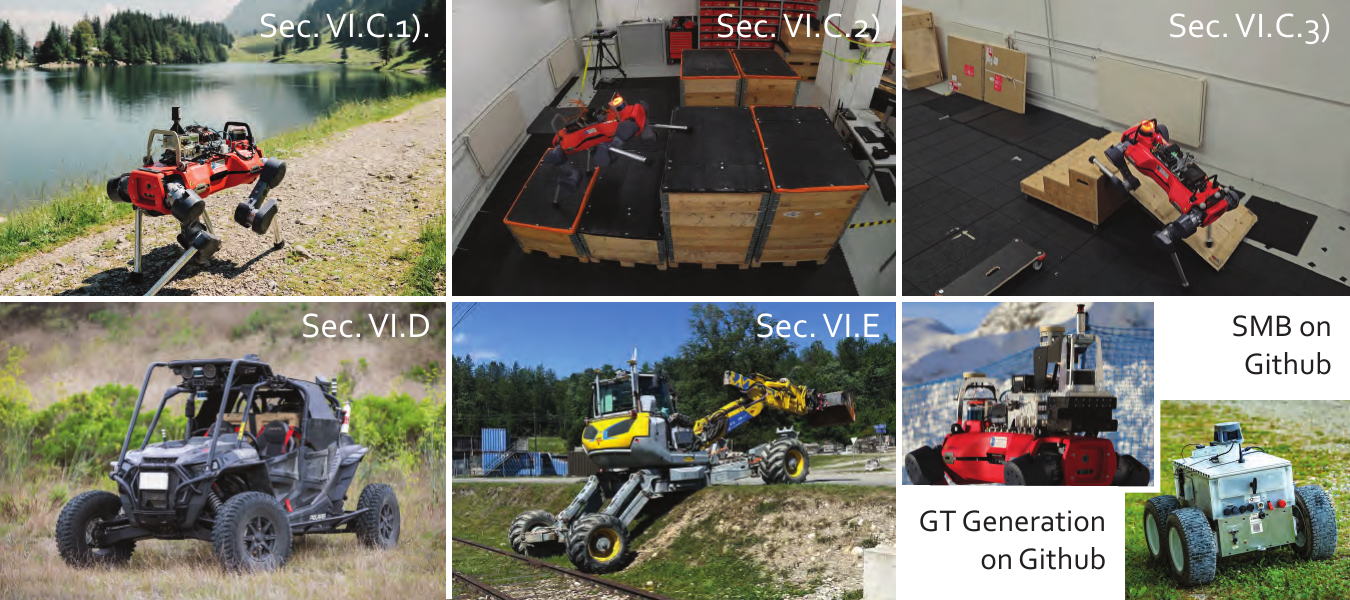}
\centering
\caption{Overview of the evaluated robotic platforms. Due to its advanced capabilities and sensor measurement setup, three ANYmal datasets (\Cref{sec:exp_anymal}.1/2/3) are investigated. Moreover, the \anontext{\ac{RACER}}{anonymous} vehicle (\Cref{sec:exp_racer}) is a case study for high-speed off-road driving and demanding tracking control. \protect\anontext{\ac{HEAP}}{excavator} (\Cref{sec:exp_heap}) operates in mixed environments and requires high accuracy for global position and orientation in the presence of geometric degeneracy. \ac{SMB}, a robot used for education, and Boxi~\cite{frey2025boxi}, an integrated sensor box for GT generation, are not  in this work but are available  in the open-source examples.\looseness=-1}
\label{fig:exp_robotic_platforms}
\vspace{-3ex}
\end{figure}

\acl{HF}'s suitability is demonstrated on three robotic platforms in five scenarios covering different sensor setups. 
A complete overview is shown in \Cref{fig:exp_robotic_platforms}, with the particularities of each system highlighted in \Cref{tab:exp_platform_particularities}. 
Each robot serves a different purpose and has a different sensor suite, from inspection and surveillance (ANYmal) to off-road traversal (\anontext{\ac{RACER}}{anonymous vehicle}) and construction (\anontext{\ac{HEAP}}{anonymous excavator}).
Note that the measurements in all investigated experiments can arrive delayed and out of order, without prior knowledge of their characteristics.\looseness=-1
\begin{table}[b]
\vspace{-3ex}
\caption{Particularities of each of the robotic platforms \& experiments.}
\vspace{-2ex}
\centering
\begin{tabularx}{\columnwidth}{>{\raggedright\arraybackslash}c|>{\raggedright\arraybackslash}c|>{\centering\arraybackslash}X}
    \hline
    \rowcolor{CaptionColor} 
    Platform & Type & Particularities \\
    \hline
    All & Sensors & \textbf{\ac{IMU}}, \textbf{\ac{LiDAR}}
    \\
    \hline
    \multirow{4}{*}{\makecell{ANYmal~\cite{hutter2016anymal}, \\ \Cref{sec:exp_anymal}}} 
        & Sensors & \textbf{leg-odometry}, single \ac{GNSS} antenna \\ \cline{2-3}
        & Motions & \textbf{dynamic}, \textbf{vertical} (climbing), \textbf{high-acceleration} stomping, multi-contact, \textbf{leg slip}, \textbf{high distances} \\ \cline{2-3}
        & Environments & indoors \& outdoors, \textbf{mixed} \\
    \hline
    \multirow{4}{*}{\makecell{\anontext{RACER~\cite{frey2024roadrunner}}{vehicle}, \\ \Cref{sec:exp_racer}}}
        & Sensors & \textbf{3 \ac{LiDAR}} sensors, \textbf{\ac{RADAR}}, \ac{GNSS}, single \textbf{wheel} encoder \\ \cline{2-3}
        & Motions & \textbf{highly dynamic}, wheel slip, \textbf{unpaved ground}, \textbf{few geometric features} \\ \cline{2-3}
        & Environments & outdoors \\
    \hline
    \multirow{4}{*}{\makecell{\anontext{\ac{HEAP}~\cite{heap2021}}{excavator}, \\ \Cref{sec:exp_heap}}} 
        & Sensors & \textbf{2 \ac{GNSS}} antennas \\ \cline{2-3}
        & Motions & \textbf{multiple-hour/days long} operations, tracking/control \textbf{in world frame} \\ \cline{2-3}
        & Environments & outdoors, \textbf{covered} by building structures \\
    \hline
\end{tabularx}
\label{tab:exp_platform_particularities}
\vspace{-2.5ex}
\end{table}
\pdfpxdimen=\dimexpr 1 in/72\relax

\subsection{Evaluation}
As all missions presented are real-world robotic applications under demanding circumstances, \ac{GT} is obtained in different ways across experiments.
For the ANYmal parkour~\cite{hoeller2024anymal} (\Cref{sec:exp_anymal_parkour}) and controlled indoor experiments (\Cref{sec:exp_anymal_indoor_mocap}), \ac{GT} is obtained directly from a \textit{Qualisys} \ac{mocap} system to investigate the smoothness and quality of the estimated trajectories.
%
\begin{table}[b]
\vspace{-3ex}
\caption{Mission overview and offline optimization complexity.}
\vspace{-2ex}
\centering
\begin{tabularx}{\columnwidth}{>{\raggedright\arraybackslash}c|>{\raggedright\arraybackslash}c|>{\raggedright\arraybackslash}c|>{\centering\arraybackslash}X|>{\centering\arraybackslash}X}
    \hline
    \rowcolor{CaptionColor} 
    Mission & ID & Length [\si{\minute}] & \# Opt. Variables & Offl. Opt. Time [\si{\second}] \\
    \hline
    \multirow{2}{*}{\makecell{ANYmal Hike \\ (\Cref{sec:exp_anymal_hike})}}
            & 1    & 23.6                  & 167,020           & 56.7                               \\
            & 2  & 32.5                  & 238,216           & 64.1                          \\
    \hline
    \makecell{ANYmal Parkour \\ (\Cref{sec:exp_anymal_parkour})}
              & 1  & 0.48                    & 3,217                & 0.66                               \\
    \hline
    \multirow{5}{*}{\makecell{ANYmal Indoor \\ (\Cref{sec:exp_anymal_indoor_mocap})}}
              & 1  & 1.03                  & 6,943             & 1.49                            \\
              & 2  & 0.99                  & 6,529             & 1.43                            \\
              & 3  & 0.93                  & 6,109             & 1.36                            \\
              & 4  & 0.96                  & 6,301             & 1.38                            \\
              & 5  & 0.76                  & 4,906             & 1.05                            \\
    \hline
    \anontext{\ac{RACER}}{Vehicle} (\Cref{sec:exp_racer})
              & 1  & 10.2                    & 66,250             & 24.04                            \\
    \hline
    \multirow{2}{*}{\makecell{\anontext{\acs{HEAP}}{Excavator} \\ (\Cref{sec:exp_heap})}}
              & 1  & 12.9                    & 115,728           & 43.2                              \\
              & 2  & 17.9                    & 161,217           & 62.4                              \\
    \hline
\end{tabularx}
\label{tab:exp_real_world_mission_data}
\end{table}
For the ANYmal hike (\Cref{sec:exp_anymal_hike}), \anontext{\ac{RACER}}{vehicle} (\Cref{sec:exp_racer}), and \anontext{\ac{HEAP}}{excavator} (\Cref{sec:exp_heap}), we use offline batch optimization with \ac{GNSS} measurements to compute globally accurate trajectories, denoted \ac{PGT}. Experiments with available real \ac{GT} (motion capture) show that the offline-optimized trajectories are smooth, high-rate, and more accurate than the real-time solutions, making them a fair comparison baseline.
We evaluate the components of \ac{HF} via an ablation study and numerical comparison against \ac{GT} or \ac{PGT}. Mission durations and offline optimization times are listed in \Cref{tab:exp_real_world_mission_data}; the quality of the offline estimates is also illustrated in the accompanying video.
For computing \ac{ATE}, \ac{ARE}, \ac{RTE}, and \ac{RRE}, we use the EVO library~\cite{grupp2017evo}. All evaluations run on a PC with an Intel i9 13900K \ac{CPU}.\looseness=-1

\subsection{ANYmal -- Agile Locomotion in Mixed Environments}
\label{sec:exp_anymal}
This section presents three real-world \ac{SOTA} robot missions: 
\textbf{\textit{i)}} a recently conducted fully autonomous hike in the \anontext{Swiss Alps}{mountains} using a \ac{VLM}-based architecture for planning (\Cref{sec:exp_anymal_hike}), 
\textbf{\textit{ii)}} the highly dynamic ANYmal parkour~\cite{hoeller2024anymal} (\Cref{sec:exp_anymal_parkour}), and \textbf{\textit{iii)}} an indoor dataset with high-rate \ac{GT} (\Cref{sec:exp_anymal_indoor_mocap}).
While \textbf{\textit{iii)}} is remote-controlled, \textbf{\textit{i)}} and \textbf{\textit{ii)}} are \textit{fully autonomous robot} missions that used \ac{HF} during deployment.\looseness=-1

\pdfpxdimen=\dimexpr 1 in/72\relax
\begin{figure}[t]
\includegraphics[width=\columnwidth]{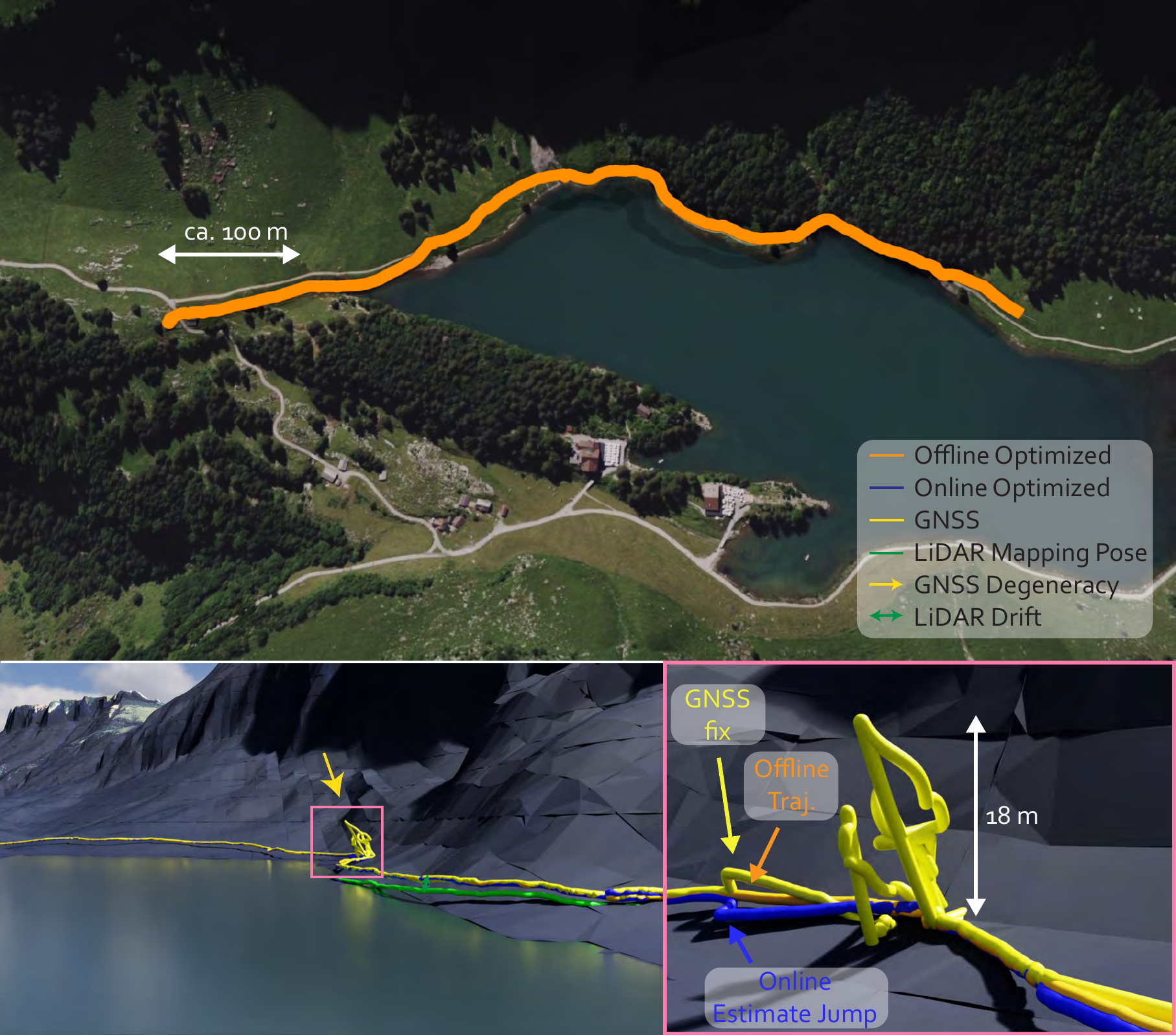}
\centering
\caption{Overview of the real-world \protect\anontext{\textbf{Seealpsee}}{mountain place} experiment. 
\textbf{Bottom row:} The \ac{GNSS} signal is unstable due to vegetation/elevation. The online estimate experiences a small jump after the GNSS return. The offline-optimized trajectory is smooth and unaffected by GNSS drops.\looseness=-1}
\label{fig:exp_anymal_hike_seealpsee}
\vspace{-3ex}
\end{figure}

\subsubsection{Autonomous Hiking}
\label{sec:exp_anymal_hike}
First, two fully autonomous hiking missions are evaluated: \textit{\textbf{a)}} a \SI{23.6}{\minute} long mission in a forest close to \anontext{Zurich, Switzerland}{forest place}, and \textit{\textbf{b)}} a \SI{32.5}{\minute} long mission in \anontext{the Swiss Alps at Seealpsee, Switzerland}{mountain place}. 
The mission setup and sensor suite are illustrated in \Cref{fig:holistic_fusion_illustration}, and the top left of \Cref{fig:exp_robotic_platforms} shows a photo of the \anontext{Seealpsee}{mountain place} deployment. 
Besides LiDAR and camera, the robot has leg joint encoders and a single \ac{RTK} \ac{GNSS} antenna (cf. \Cref{tab:exp_platform_particularities}). The estimator fuses \ac{GNSS}, \ac{IMU}, and leg-kinematic measurements with \ac{LiDAR} scan-to-map registration poses of a degeneracy-aware~\cite{tuna2022x, tuna2024informed} variant of Open3D \ac{SLAM}~\cite{jelavic2022open3d}.
Both hikes are conducted fully autonomously using a \ac{VLM}-based planner.\looseness=-1
\begin{table}[b]
\vspace{-3ex}
\centering
\caption{Global estimation quality comparison for the \textbf{hike} experiments. HF - Odom corresponds to the odometry introduced in \Cref{sec:method_smoothness}. \textbf{MINS:} The \textbf{Mountain} experiment \textbf{diverged} (``div.") mid-way; hence we also report results without it (``split").}
\vspace{-2ex}
\begin{tabularx}{\columnwidth}{l|>{\centering\arraybackslash}X >{\centering\arraybackslash}X|>{\centering\arraybackslash}X >{\centering\arraybackslash}X}
    \hline
    \rowcolor{CaptionColor} 
    Method & \multicolumn{2}{c|}{ATE [\si{\meter}]} & \multicolumn{2}{c}{ARE [\si{\deg}]} \\
    \rowcolor{CaptionColor} 
                                          & Forest & Mountain & Forest & Mountain \\
    \hline
    ANYmal \acs{TSIF}~\cite{bloesch2017two} - Odom     & 33.38 & 18.22    & 15.42 & 9.09 \\
    Open3D SLAM~\cite{jelavic2022open3d} - \acs{LR} & 1.46  & 1.25     & 2.53  & 4.47 \\
    \hline
    MINS~\cite{lee2023mins} - World (div.)    & 0.44 & 4.53    & 2.14 & 29.86 \\
    MINS~\cite{lee2023mins} - World (split)    & 0.44 & 0.23    & 2.14 & 4.10 \\
    \hline
    HF \ac{GNSS}+\ac{IMU} - World & 0.53 & 1.03 & 6.47 & 11.25 \\ 
    HF \ac{GNSS}+\ac{IMU} - Odom & 101.55 & 37.84 & 104.17 & 21.61 \\
    \hline
    \ac{HF} - World (\acs{LR}-between)        & 0.52  & 1.12     & 5.56  & 13.00 \\
    \ac{HF} - Odom (\acs{LR}-between)         & 66.00 & 109.95    & 29.17 & 41.79 \\
    \hline
    \ac{HF} - World                                    & \textbf{0.42}  & 0.38     & 1.42  & 2.63 \\
    \ac{HF} - Odom                                     & 22.91 & 24.97    & 11.55 & 13.62 \\
    \hline
    \ac{HF} - World (\ac{GNSS} filtered) & 0.49 & \textbf{0.12} & \textbf{1.23} & \textbf{1.22} \\
    \ac{HF} - Odom (\ac{GNSS} filtered) & 15.94 & 10.69    & 8.87  & 6.11 \\
    \hline
\end{tabularx}
\label{table:exp_anymal_hike_absolute_errors}
\end{table}

\paragraph{Global Estimation Quality}
A reliable and robust estimation in $\World$ is necessary to follow global waypoints despite poor or missing \ac{GNSS} due to vegetation or multipath effects, especially in mountain regions.
\Cref{fig:title_anymal_hike} highlights the \ac{GNSS} degeneracy and estimated trajectories for the forest deployment, and \Cref{fig:exp_anymal_hike_seealpsee} does the same for \anontext{Seealpsee}{mountain place}.
\Cref{table:exp_anymal_hike_absolute_errors} reports the \ac{ATE} and \ac{ARE} for both deployments, showing that fusing both \ac{GNSS} and absolute \ac{LR} is needed for global accuracy given unstable \ac{GNSS} and drifting \ac{LiDAR} map registration.
At times, \ac{GNSS} performs poorly, and the absolute \ac{LR} measurements are not attitude-aligned, as the map used does not have a notion of gravity, highlighting the need for automatic alignment if fused as an absolute measurement. 
Interestingly, fusing the \ac{LR} poses as absolute measurements (\textit{HF - World} and \textit{HF - World (GNSS filtered)}) outperforms adding them as between factors (\textit{HF - World (LR-between)}). Here, \textit{GNSS filtered} refers to fusing-in the \ac{GNSS} measurement only if below a covariance threshold, \SI{1}{\meter} in this case.
While the absolute errors for \textit{HF - Odom} are (expectedly) larger than \textit{HF - World}, they are still much smaller than \textit{\ac{TSIF} - Odom}, suggesting a more minor overall drift despite being an entirely local quantity.
Moreover, we compare against MINS~\cite{lee2023mins}, a global \ac{GNSS}/\ac{IMU}/\ac{LiDAR} fusion framework. While it works well in the Forest setting, simultaneous \ac{LiDAR} and \ac{GNSS} dropout during the Mountain Hike required splitting the dataset (\textit{split} in \Cref{table:exp_anymal_hike_absolute_errors}). \ac{HF} outperforms MINS in this challenging case, highlighting its \ac{SOTA} global-fusion performance.\looseness=-1

\pdfpxdimen=\dimexpr 1 in/72\relax
\begin{figure}[t]
\includegraphics[width=\columnwidth]{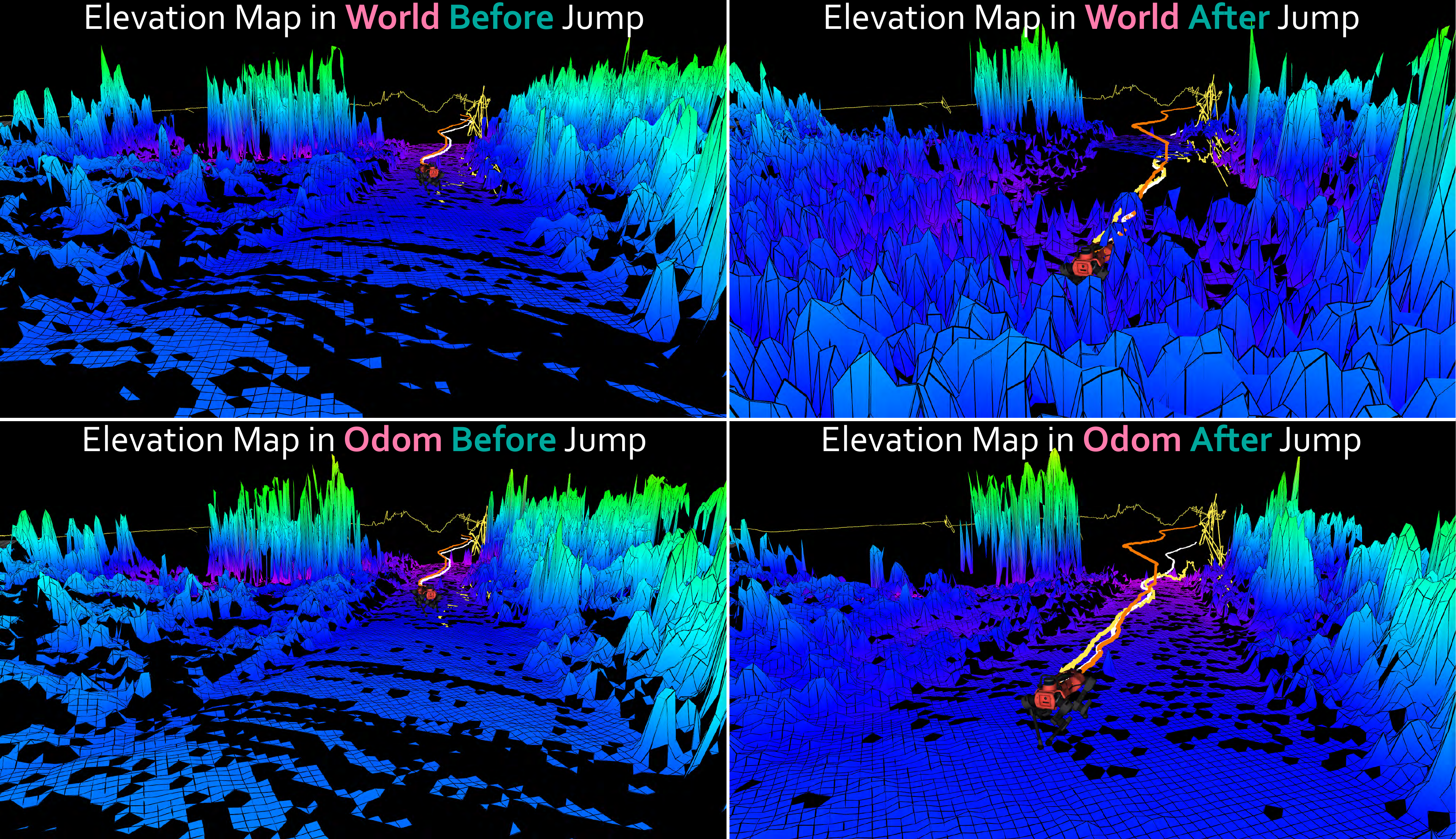}
\centering
\caption{Qualitative result for a local elevation map expressed in $\Odom$ vs. $\World$ for the \textbf{forest hike dataset}. The return of \ac{GNSS} leads to an update jump of the estimate in $\World$, creating a corrupted elevation map. The odometry estimate in $\Odom$ does not jump (\Cref{sec:method_smoothness}), rendering it suitable for local mapping and navigation.\looseness=-1}
\label{fig:exp_anymal_hike_elevation_mapping}
\vspace{-3ex}
\end{figure}
\begin{table}[b]
\centering
\vspace{-3ex}
\caption{Average local estimation quality and smoothness in terms of \ac{RTE}, \ac{RRE}, \ac{NOJ} and jitter for the two ANYmal \textbf{hike} experiments. \textbf{MINS:} Evaluating full forest and split version of mountain experiment (parts 1 and 2) due to divergence (cf.~\Cref{table:exp_anymal_hike_absolute_errors}).}
\vspace{-2ex}
\begin{tabularx}{\columnwidth}{l|>{\raggedright\arraybackslash}c|>{\raggedright\arraybackslash}c|>{\centering\arraybackslash}X|>{\centering\arraybackslash}X}
    \hline
    \rowcolor{CaptionColor} 
    Method & \makecell[tc]{RTE \\ $[\%]$} & \makecell[tc]{RRE \\ $[\si{\degree}/\si{\meter}]$} & \makecell[tc]{\ac{NOJ} \\ $[\#]$} & \makecell[tl]{Jitter \\ {$[\si{\meter}/\si{\second}^3]$}} \\
    \hline
    ANYmal \acs{TSIF}~\cite{bloesch2017two} - Odom & 5.01     & \textbf{0.57}     & 45     & 48,400   \\
    Open3D-SLAM~\cite{jelavic2022open3d} - \acs{LR} ($\leq$10\si{\hertz})                   & 6.28     & 3.66     & 1        & -   \\
    \hline
    MINS~\cite{lee2023mins} - World (split)     & 4.52 & 0.62    & 1,364 & 70,619 \\
    \hline
    HF - World (\acs{LR}-between)       & 4.84     & 4.22     & 5,727     & 245,777  \\
    HF - Odom (\acs{LR}-between)        & 4.09     & 2.73     & 1,898     & 5,094   \\
    \hline
    \ac{HF} - World                                  & 4.07     & 0.97     & 2,482     & 59,382 \\
    \ac{HF} - Odom                                   & 3.13     & 0.71     & \textbf{0}         & \textbf{3,470} \\
    \hline
    \ac{HF} - World (\ac{GNSS} filtered)  & \textbf{2.62} & 0.81 & 631 & 50,053   \\
    \ac{HF} - Odom (\ac{GNSS} filtered)   & 3.10 & 0.65 & \textbf{0}   & 3,890   \\
    \hline
\end{tabularx}
\label{table:exp_anymal_hike_local_errors}
\end{table}

\paragraph{Local Estimation Quality and Consistency}
Locally precise, smooth, and consistent estimates without jumps are essential for control, path-following, or local elevation mapping tasks.
The \ac{RTE}, \ac{RRE}, \ac{NOJ}, and jerk (third derivative of translation estimate) are reported in \Cref{table:exp_anymal_hike_local_errors}. 
The \ac{RTE} and \ac{RRE} are defined as the average drift of all pairs of \SI{1}{\meter} traversed distance. Jerk, given as the third time derivative of position, is a commonly reported metric in human body pose modeling and graphics~\cite{huang2018deep} to measure discontinuities. At the same time, a jump in \ac{NOJ} is defined as a motion of more than \SI{10}{\centi\meter} between two estimates (at \SI{400}{\hertz} \ac{IMU} rate).
It can be seen that the \textit{\ac{HF} - Odom} estimate significantly reduces the amount of jerk and \ac{NOJ}, when compared to \ac{HF} - World and also \ac{TSIF}, which is the default leg odometry estimator on ANYmal and also an entirely local odometry estimation solution. \ac{HF} also outperforms MINS in all quantities except \ac{RRE} by a significant margin, again highlighting superior performance on the evaluated datasets. Finally, a qualitative result of local elevation mapping in $\World$ or $\Odom$ during \ac{GNSS} absence and return is shown in \Cref{fig:exp_anymal_hike_elevation_mapping}.
In contrast to $\World$, the jump is fully suppressed in $\Odom$, leading to a smooth and consistent map suitable for locomotion~\cite{miki2022learning}.
As a result, \ac{HF} should be seen not only as a global localization system but also as a local odometry solution that provides smooth estimates, ideally suited for control, which are superior to TSIF and can hence fully replace it both globally and locally.\looseness=-1
\pdfpxdimen=\dimexpr 1 in/72\relax
\begin{figure}[t]
\includegraphics[width=\columnwidth]{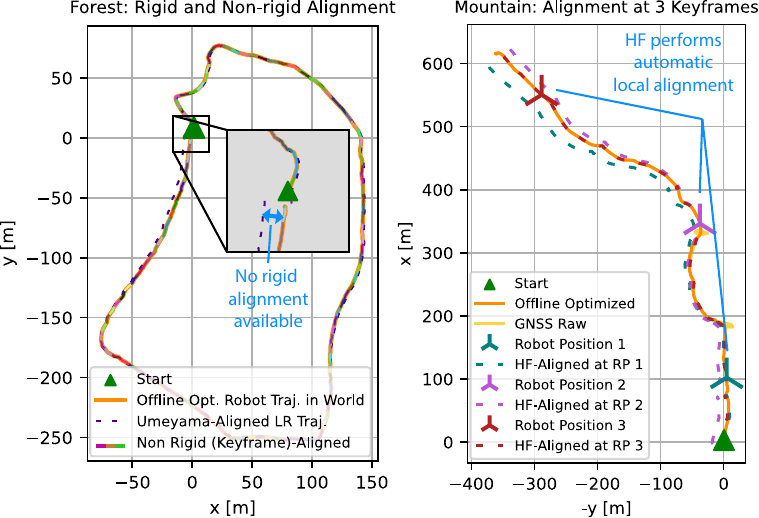}
\centering
\caption{Alignment of the \ac{LR} trajectory and the fused estimate. \textbf{Left:} Rigid (Umeyama)-aligned trajectory vs. local non-rigid alignment of \ac{HF} (color corresponding to the corresponding keyframe) for the \textbf{forest} experiment. Here, Umeyama is not able to align the \ac{LR} with the fused estimate near the end due to the occurrence of drift. \textbf{Right:} HF alignment visualized at three keyframes for the \textbf{mountain hike}. The two trajectories align perfectly locally around their corresponding keyframes.\looseness=-1}
\label{fig:exp_hike_alignment}
\vspace{-2ex}
\end{figure}
\pdfpxdimen=\dimexpr 1 in/72\relax
\begin{figure}[b]
\vspace{-3ex}
\includegraphics[width=\columnwidth]{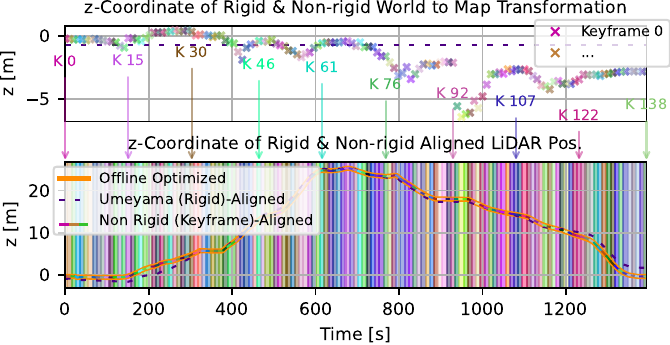}
\centering
\caption{Umeyama alignment (dashed) and non-rigid local alignment around keyframes (colored) of the \ac{LR}- and the offline optimized trajectories for the \protect\anontext{\textbf{Hoenggerberg forest}}{forest place} experiment. \textbf{Top:} z-coordinate of the global Umeyama alignment and each keyframe coordinate. \textbf{Bottom:} z-coordinate of the aligned drifting \ac{LR} poses using the alignments from the top. It can be seen that a single rigid alignment is not even sufficient for proper alignment in translation.\looseness=-1}
\label{fig:exp_hike_non_rigid_alignment_hoengg}
\end{figure}

\begin{figure}[t]
\includegraphics[width=\columnwidth]{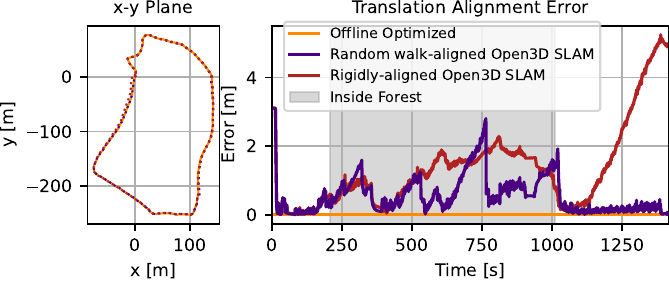}
\centering
\caption{\textbf{Left:} Visualization of the online and offline-estimated trajectories for the \textbf{forest} experiment using random walk and \textbf{no} random walk. \textbf{Right:} the translation component of the alignment error between the aligned \ac{LR} pose and the offline optimized trajectory. Even in the presence of good \ac{GNSS} signal outside the forest, the \textit{no random walk} case (red) is not able to properly align the \ac{LR} pose and the fused trajectory due to the accumulated drift in the \ac{LiDAR} map.\looseness=-1}
\label{fig:exp_anymal_hike_no_random_walk}
\end{figure}

\begin{figure}[b]
\vspace{-3ex}
\includegraphics[width=\columnwidth]{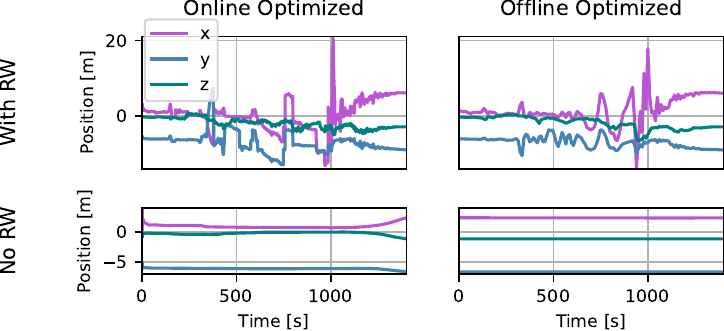}
\centering
\caption{\textbf{Left:} Change in ${}_{\World}\transl_{\World \Map}$ during online estimation with and without random walk modeling for the \textbf{forest} experiment. \textbf{Right:} Similar behavior of the estimated drift for the offline batch optimization. Takeaway 1: The drift can be estimated well, even in the online case. Takeaway 2: For the case of \textit{zero random walk}, the drift is kept roughly constant even in the online case due to the marginalization and belief propagation (in contrast to \ac{MHE}).\looseness=-1}
\label{fig:exp_anymal_hike_online_offline_random_walk_no_random_walk}
\end{figure}

\paragraph{Reference-Frame Alignment and Drift}
\label{sec:exp_anymal_ref_frame_drift}
As shown in \Cref{fig:title_anymal_hike}-A for the forest dataset, there are two issues with a direct fusion of the absolute \ac{LR} poses: \textbf{\textit{i)}} they are not aligned with \ac{GNSS}, \textbf{\textit{ii)}} they drift without a global reference, changing the trajectory shape. 
While \textbf{\textit{i)}} can be partially addressed via Umeyama alignment, shown in \Cref{fig:exp_hike_alignment} for \anontext{Seealpsee}{mountain place}, it \textit{cannot} handle inter-frame drift. 
As shown in \Cref{fig:title_anymal_hike}-B and \Cref{fig:exp_hike_alignment} (right) in three different keyframes each, the proposed framework aligns the reference frames online at each moment in time and models the drift as a random walk. 
The estimated drift of the Open3D \ac{SLAM} map frame w.r.t. $\World$ is visualized in \Cref{fig:title_anymal_hike}-C. 
Interestingly, since this alignment estimation is continuously performed, each trajectory snippet of the \ac{LR} trajectory can locally be aligned, yielding non-rigid alignment as shown in the colored trajectories (colored by keyframe) of \Cref{fig:exp_hike_alignment} (left) and \Cref{fig:exp_hike_non_rigid_alignment_hoengg}.
Without allowing the map frame $\OpenThree$ to drift against $\World$, emulated by setting the random walk to $0$, the \ac{GNSS} and the \ac{LR} trajectories \textit{cannot} be aligned, as seen near the mission end in \Cref{fig:exp_anymal_hike_no_random_walk}.
\pdfpxdimen=\dimexpr 1 in/72\relax
Effectively, this results in an estimate somewhere in between \ac{LR} and \ac{GNSS} trajectories. 
This behavior occurs both for the offline and the online estimate due to proper marginalization of the previous alignment belief. This can be seen in \Cref{fig:exp_anymal_hike_online_offline_random_walk_no_random_walk}, where the estimated alignment behaves similarly for online and offline optimization.\looseness=-1
\pdfpxdimen=\dimexpr 1 in/72\relax

\begin{figure}[t]
\replacebox{%
\includegraphics[width=\columnwidth]{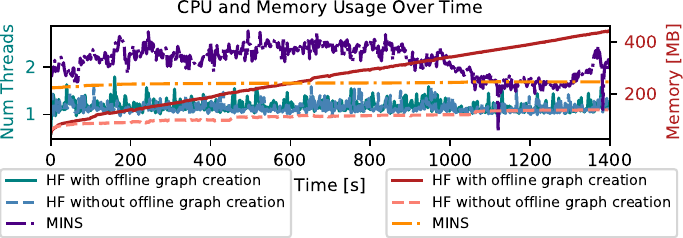}
\centering
\caption{\replace{\ac{CPU} memory usage over time for \textit{i)} no offline graph creation, and \textit{ii)} with offline graph creation. As expected, creating the offline graph results in growing memory usage while the average CPU load remains constant.}{\ac{CPU} and memory usage over time for \ac{HF} (\textit{i)} without and \textit{ii)} with offline graph creation) and MINS~\cite{lee2023mins}. The computational load of MINS is higher and less steady than that of \ac{HF}. Note that this comparison is not fully fair, as MINS includes LiDAR odometry internally, whereas \ac{HF} runs it as a separate process.}\looseness=-1}
\label{fig:exp_anymal_hike_memory_over_time}
} 
\vspace{-2ex}
\end{figure}

\paragraph{Computational Complexity}
Solving asynchronous optimization on ANYmal is challenging and historically motivated filtering-based approaches.
This section reports on the computational requirements for running the estimator for the presented scenario.
\Cref{tab:exp_anymal_hike_computational_complexity} lists mean and std. of \textit{i)} latency, \textit{ii)} async optimization time, \textit{iii)} \ac{ATE}, and \textit{iv)} \ac{ARE} for different state-creation rates.
Independent of the state-creation rate, the latency remains small in the lower \si{\micro\second} range. As expected, asynchronous optimization time rises with more variables in the smoothing window.
Note that the state is always propagated on the robot at full \ac{IMU} frequency independent of the state-creation rate. Although almost no difference can be observed in \ac{ATE} and \ac{ARE} between \SI{40}{\hertz} and \SI{100}{\hertz}, \SI{10}{\hertz} leads to an increase in translation and orientation error.
Moreover, the memory and \ac{CPU} load over time for the entire forest mission are shown in \Cref{fig:exp_anymal_hike_memory_over_time} for the cases of with and without offline graph creation in the background\replace{.}{, alongside the resource consumption of MINS~\cite{lee2023mins} on the same dataset. Note that this comparison is not fully fair, as MINS includes LiDAR odometry internally, whereas \ac{HF} runs LiDAR odometry as a separate module.}
\pdfpxdimen=\dimexpr 1 in/72\relax
\begin{table}[t]
\caption{Tradeoff of the computational complexity and accuracy vs.\ the state-creation rate for the \textbf{ANYmal hike} experiment.}
\centering
\begin{tabularx}{\columnwidth}{
    l|
    >{\raggedright\arraybackslash}c 
    >{\raggedright\arraybackslash}c|
    >{\raggedright\arraybackslash}c 
    >{\raggedright\arraybackslash}c|
    >{\centering\arraybackslash}X 
    >{\centering\arraybackslash}X|
    >{\centering\arraybackslash}X 
    >{\centering\arraybackslash}X}
    \hline
    \rowcolor{CaptionColor} 
    Rate & \multicolumn{2}{c|}{Latency $[$\si{\micro\second}$]$} & \multicolumn{2}{c|}{Opt. Time $[$\si{\milli\second}$]$} & \multicolumn{2}{c|}{ATE [\si{\meter}]} & \multicolumn{2}{c}{ARE [\si{\degree}]} \\
    \rowcolor{CaptionColor} 
    $[\si{\hertz}]$ & $\mu$ & $\sigma$ & $\mu$  & $\sigma$ & $\mu$  & $\sigma$ & $\mu$  & $\sigma$ \\
    \hline
    10         & \textbf{22.41} & \textbf{13.62} & \textbf{1.29} & \textbf{0.25} & 0.384   & \textbf{0.258} & 1.715 & 1.511 \\
    40         & 25.31          & 17.07          & 3.86          & 0.77          & \textbf{0.373}   & 0.259 & 1.562 & 1.429 \\
    100        & 29.63          & 19.99          & 8.96          & 1.35          & 0.374   & 0.260 & \textbf{1.557} & \textbf{1.424}  \\
    \hline
\end{tabularx}
\label{tab:exp_anymal_hike_computational_complexity}
\vspace{-2ex}
\end{table}
As expected, the allocated memory remains near-constant for the online graph-only case, but it grows in size at the state-creation rate for the offline smoother case, as the entire history needs to be stored.
Conversely, the \ac{CPU} load remains constant for both scenarios throughout the entire mission.\looseness=-1

\subsubsection{ANYmal Parkour}
\label{sec:exp_anymal_parkour}
\begin{figure}[t]
\includegraphics[width=\columnwidth]{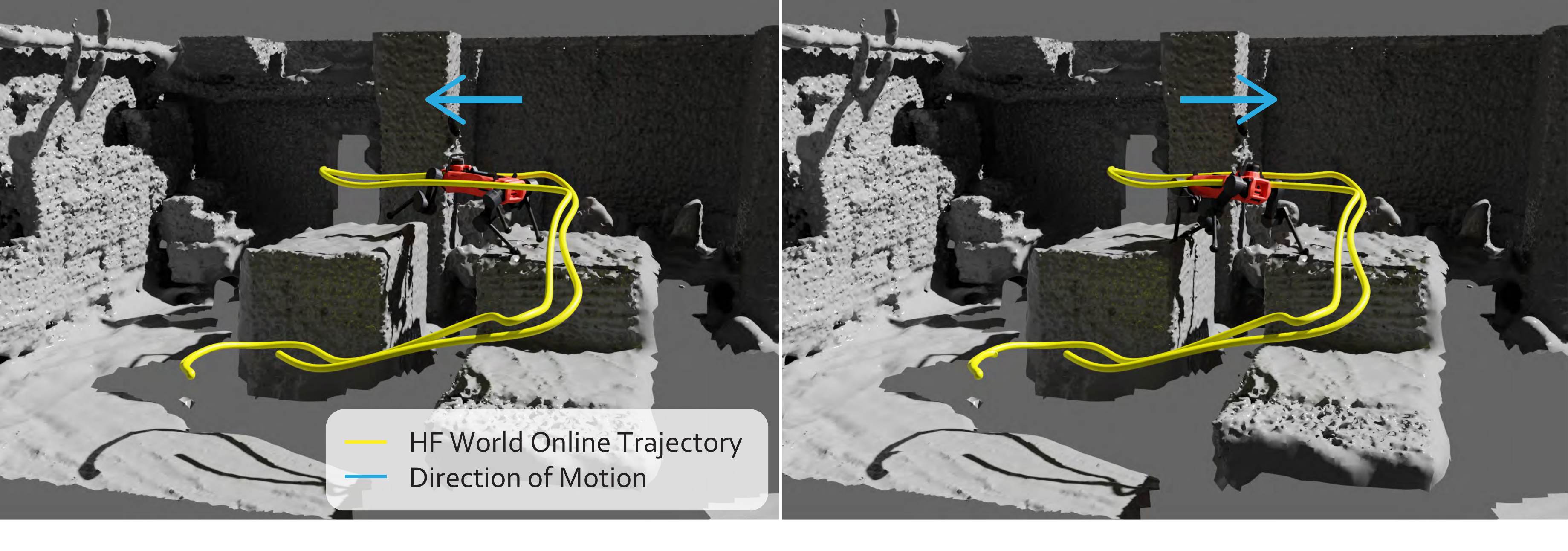}
\includegraphics[width=\columnwidth]{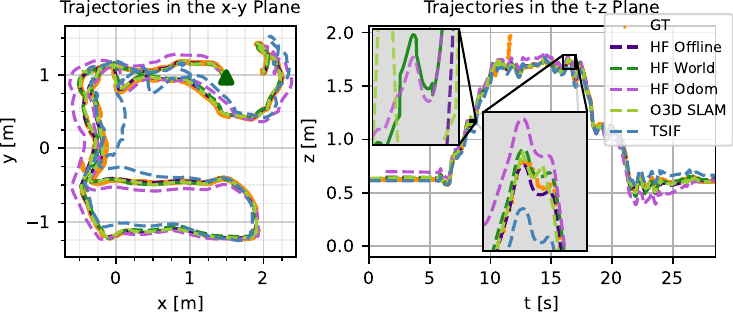}
\centering
\caption{\textbf{Top:} Visualization of the \textbf{ANYmal parkour} experiment. The mesh of the environment is generated using the offline estimate of \ac{HF}. \textbf{Bottom:} Top-down view and z-coordinate estimate of the different estimates. While \ac{HF} World is more accurate, \ac{HF} Odom leads to significantly smoother estimates.\looseness=-1}
\label{fig:exp_anymal_parkour_estimated_map_and_traj}
\vspace{-3ex}
\end{figure}

ANYmal parkour~\cite{hoeller2024anymal} constitutes one of the most dynamic and complex autonomous robot motion controllers on a quadrupedal robot to date.
Fast vertical motions (up to \SI{2.07}{\meter\per\second} total speed and \SI{59.43}{\meter\second^{-2}} effective acceleration) characterize its deployment, including jumping over gaps and hard-to-detect-and-model multi-contact interactions between the environment and the robot body, knees, and shanks. 
We focus on smooth, low-latency trajectories and minimal drift to aid the \ac{RL} controller.
\Cref{fig:exp_anymal_parkour_estimated_map_and_traj} shows estimated paths of \textbf{\textit{i)}} offline and \textbf{\textit{ii)}} online $\World$/$\Odom$ \ac{HF}, \textbf{\textit{iii})} \ac{TSIF} (ANYbotics’ default leg-inertial odometry), and \textbf{\textit{iv)}} raw low-rate Open3D SLAM poses.
The mesh shown at the top of \Cref{fig:exp_anymal_parkour_estimated_map_and_traj} is generated from the accumulated point-cloud map of the Velodyne VLP-16 LiDAR using the provided offline optimized poses of \ac{HF}.\looseness=-1
\begin{table}[b]
\centering
\vspace{-2ex}
\caption{\ac{ATE}, \ac{ARE}, \ac{RTE} and \ac{RRE} mean and standard deviation of the different methods for the \textbf{ANYmal parkour} experiment. Best result for all high rate measurements in \textbf{bold}.}
\vspace{-2ex}
\begin{tabularx}{\columnwidth}{l|>{\raggedright\arraybackslash}X >{\raggedright\arraybackslash}X|>{\raggedright\arraybackslash}X >{\raggedright\arraybackslash}X |>{\centering\arraybackslash}X >{\centering\arraybackslash}X |>{\centering\arraybackslash}X >{\centering\arraybackslash}X}
    \hline
    \rowcolor{CaptionColor} 
    Method & \multicolumn{2}{c|}{ATE [\si{\centi\meter}]} & \multicolumn{2}{c|}{ARE [\si{\degree}]} & \multicolumn{2}{c|}{RTE [\%]} & \multicolumn{2}{c}{RRE [\si{\degree}/\si{\meter}]} \\
    \rowcolor{CaptionColor} 
                                    & $\mu$ & $\sigma$ & $\mu$ & $\sigma$ & $\mu$ & $\sigma$ & $\mu$ & $\sigma$ \\
    \hline
    \rowcolor{gray!12} 
    \multicolumn{9}{c}{\ac{LR} Only (\textbf{low rate})} \\
    \hline
    Open3D-SLAM                    &  4.2 & 8.3     & 4.3 & 12.0      & 3.8  & 8.3     & 4.5  & 2.3     \\
    \hline
    \rowcolor{gray!12} 
    \multicolumn{9}{c}{\ac{IMU} + leg kinematic tightly fused (\textbf{high rate})} \\
    \hline
    \ac{TSIF}~\cite{bloesch2017two} &  24.9 & 15.2     & 7.5  & \textbf{13.9}    & 15.3 & 14.6    & \textbf{8.0}  & 26.7     \\
    \ac{HF} World                   &  33.2 & 19.9     & 9.4  & 14.3    & 13.1 & 13.6    & 11.0  & 31.5     \\
    \ac{HF} Odom                    &  43.2 & 23.8     & 11.1  & 14.8   & 14.3  & 16.1   & 10.1  & 28.9     \\
    \hline
    \rowcolor{gray!12} 
    \multicolumn{9}{c}{\ac{IMU} + \ac{LR} + leg kinematic tightly fused (\textbf{high rate})} \\
    \hline
    \ac{HF} World &  \textbf{7.0} & \textbf{10.5}     & \textbf{4.2}  & 14.6     & 8.3  & 11.1    & 8.8  & \textbf{26.4}     \\
    \ac{HF} Odom  &  17.4 & 12.6    & 5.0  & 14.6     & \textbf{7.5}  & \textbf{10.8}    & 10.5  & 29.3     \\
    \hline
    \hline
    \ac{HF} Offline &  4.9 & 10.5     & 3.7  & 15.5     & 3.9  & 10.1    & 8.5  & 28.0     \\
    \hline
    \rowcolor{gray!12} 
    \multicolumn{9}{c}{\ac{IMU} +\ac{LR} + leg kinematic tightly fused + robust norm (\textbf{high rate})} \\
    \hline
    \ac{HF} World & \textbf{7.0} & 10.6     & \textbf{4.2}   & 14.5    & 8.4  & 11.1    & 9.0  & 27.0    \\
    \ac{HF} Odom  & 16.3 & 11.9    & 4.6   & 14.7    & 8.0  & 10.9    & 11.2  & 28.3     \\
    \hline
    \hline
    \ac{HF} Offline & 4.8 & 10.6   & 3.7   & 15.6    & 3.9  & 10.1    & 8.2  & 27.1     \\
    \hline
\end{tabularx}
\label{table:exp_anymal_parkour_quantitative}
\end{table}

Quantitative results are reported in \Cref{table:exp_anymal_parkour_quantitative}. 
While the much simpler leg-only estimation of \ac{HF} is not as accurate as the more engineered and optimized commercial TSIF estimator based on~\cite{bloesch2017two}, \ac{HF} can significantly reduce all error metrics as soon as \ac{LR} is fused in.
Moreover, the estimate is smoother in $\Odom$ and achieves a lower \ac{RTE}.
The introduction of \ac{LR} reduces not only \ac{ATE} and \ac{ARE}, as expected, but also the more critical local quantities \ac{RTE} and \ac{RRE}, which matter most during highly dynamic motions, as present here. Moreover, the benefit of tightly fusing the leg odometry also becomes apparent, both for \ac{IMU}+leg fusion and for \ac{IMU}+\ac{LR}+Leg fusion.
The benefit of using robust norms on noisy and deficient foothold positions to reject outliers is unclear despite the high amount of slip and multi-contact.\looseness=-1

\pdfpxdimen=\dimexpr 1 in/72\relax
\begin{table}[b]
\vspace{-3ex}
\centering
\caption{Comparison of methods in terms of \ac{ATE}, \ac{ARE}, \ac{RTE} and \ac{RRE} for the \textbf{five indoor evaluation sequences} against \ac{mocap} \ac{GT}.}
\vspace{-2ex}
\resizebox{1.0\columnwidth}{!}{
\begin{tabularx}{\columnwidth}{l|>{\raggedright\arraybackslash}X >{\raggedright\arraybackslash}X|>{\raggedright\arraybackslash}X >{\raggedright\arraybackslash}X |>{\centering\arraybackslash}X >{\centering\arraybackslash}X |>{\centering\arraybackslash}X >{\centering\arraybackslash}X}
    \hline
    \rowcolor{CaptionColor} 
    Method & \multicolumn{2}{c|}{ATE [\si{\centi\meter}]} & \multicolumn{2}{c|}{ARE [\si{\degree}]} & \multicolumn{2}{c|}{RTE [\%]} & \multicolumn{2}{c}{RRE [\si{\degree}/\si{\meter}]} \\
    \rowcolor{CaptionColor} 
                                    & $\mu$ & $\sigma$ & $\mu$ & $\sigma$ & $\mu$ & $\sigma$ & $\mu$ & $\sigma$ \\
    \hline
    \rowcolor{gray!12} 
    \multicolumn{9}{c}{\ac{LR} Only (\textbf{low rate})} \\
    \hline
    Open3D-SLAM                    &  3.0 & 1.5     & 2.70 & 0.64      & 2.59  & 1.92     & 1.40  & 1.12     \\
    \hline
    \rowcolor{gray!12} 
    \multicolumn{9}{c}{\ac{IMU} + leg odometry velocity loosely fused (\textbf{high rate})} \\
    \hline
    \ac{HF} World                      &  13.7 & 6.3     & 5.69 & 2.73      & 2.72  & 2.49     & 1.39  & 1.02     \\
    \hline
    \rowcolor{gray!12} 
    \multicolumn{9}{c}{\ac{IMU} + leg kinematics tightly fused (\textbf{high rate})} \\
    \hline
    \ac{TSIF}~\cite{bloesch2017two} &  7.9 & 2.9   & 3.73 & 0.75 & 2.17 & 2.26 & \textbf{1.24} & \textbf{0.90}  \\
    \ac{HF} World                   & 10.1 & 4.0   & 4.97 & 1.36 & 2.60 & 2.58 & 1.52 & 1.11  \\
    \hline
    \rowcolor{gray!12} 
    \multicolumn{9}{c}{\ac{IMU} + \ac{LR} + leg odometry velocity loosely fused (\textbf{high rate})} \\
    \hline
    \ac{HF} World                   & 4.0 & 2.4 & \textbf{2.68} & 0.59 & 2.61 & 2.41 & 1.36  & 1.01     \\
    \hline
    \rowcolor{gray!12} 
    \multicolumn{9}{c}{\ac{IMU} + \ac{LR} + leg kinematics tightly fused (\textbf{high rate})} \\
    \hline
    \ac{HF} World &  \textbf{2.8}  & \textbf{1.4}    & 2.75  & \textbf{0.54}    & \textbf{2.04}  & \textbf{1.67}  & 1.32  & 0.96     \\
    \hline
    \hline
    \ac{HF} Offline                 &  2.8  & 1.4    & 2.58  & 0.52    & 2.14  & 1.63     & 1.29  & 0.98     \\
    \hline
\end{tabularx}
}
\label{table:exp_anymal_indoor_quantitative}
\end{table}

\subsubsection{Indoor Locomotion Dataset}
\label{sec:exp_anymal_indoor_mocap}
\begin{figure}[t]
\includegraphics[width=\columnwidth, clip, trim={0 0 0 {.065\columnwidth}}]{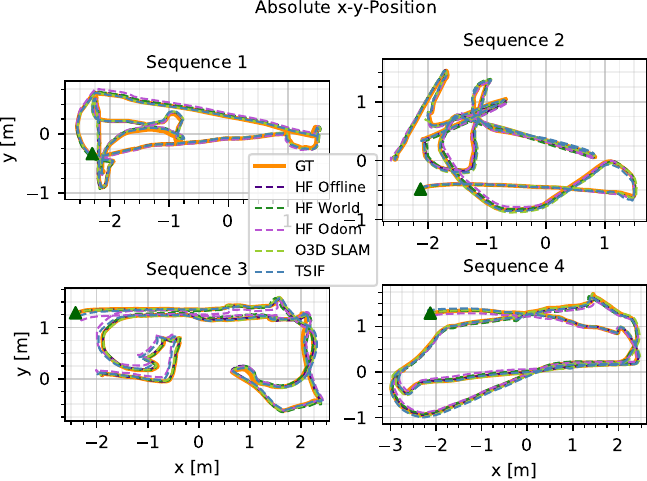}
\centering
\caption{Top-down view of the first four sequences of the \textbf{indoor locomotion} dataset. The methods are compared against the \ac{GT} \ac{mocap} trajectory.\looseness=-1}
\label{fig:exp_anymal_indoor}
\vspace{-3ex}
\end{figure}

This section provides additional quantitative evaluations on ANYmal against the \textit{Qualisys} \ac{mocap} \ac{GT} on five sequences, being a more controlled indoor evaluation of walking at various speeds on different obstacles. In addition, the robot steps on unstable, moving ground, rendering contact estimation difficult. Lastly, the robot traverses short \SI{45}{\deg} stairs, posing challenges through rapid height and attitude change.
The sensors used in this experiment are \ac{IMU}, \ac{LiDAR} and leg kinematics.
We focus on the impact of tightly including the leg kinematics vs.\ loosely fusing the velocity estimate of the external leg odometry estimator \ac{TSIF}~\cite{bloesch2017two}.
\Cref{fig:exp_anymal_indoor} shows a top-down view of the first four sequences.
\Cref{table:exp_anymal_indoor_quantitative} provides the averaged results over all five sequences.
The first two subcategories show the benefit of tight vs.\ loose kinematic fusion (velocity of \ac{TSIF}).
As in \Cref{sec:exp_anymal_parkour}, the estimation performance of \ac{TSIF}~\cite{bloesch2017two} is slightly better than the simpler formulation of \ac{HF}.
\pdfpxdimen=\dimexpr 1 in/72\relax
\pdfpxdimen=\dimexpr 1 in/72\relax
\begin{figure}[b]
\vspace{-3ex}
\includegraphics[width=\columnwidth]{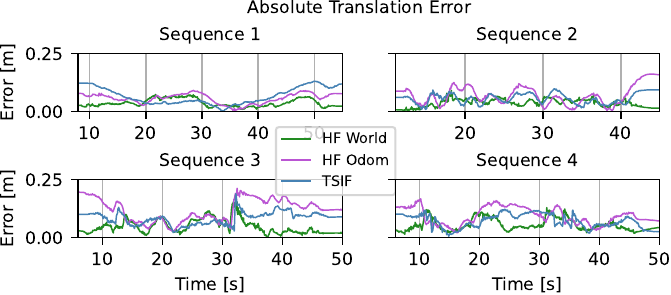}
\includegraphics[width=\columnwidth]{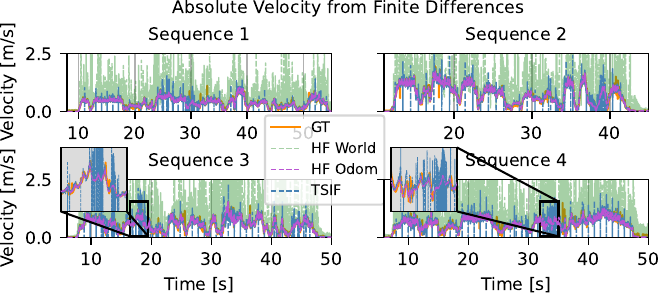}
\centering
\caption{\textbf{Two top rows:} absolute translation error of the different methods. The estimate in world has a smaller absolute error than the odometry solutions. \textbf{Two bottom rows:} absolute velocity computed via finite differences from the estimated position. The \ac{HF} Odom estimate is by far the smoothest, even more than the fully local \ac{TSIF}, in particular during slip (zoomed).\looseness=-1}
\label{fig:exp_anymal_indoor_eval}
\end{figure}
Yet, when including \ac{LR}, the results significantly reduce drift. Interestingly, in the presence of \ac{LR}, the tightly-fused leg kinematic formulation of \Cref{sec:local_landmark} performs better than loosely fusing the non-perfect velocity estimates of \ac{TSIF}, reducing the \ac{ATE} and \ac{ARE}.
\ac{HF} works well for the investigated dynamic-but-controlled motions, with a few obstacles on the ground, highlighted by the smaller errors than in \Cref{table:exp_anymal_parkour_quantitative}.
The two top rows of \Cref{fig:exp_anymal_indoor_eval} show the \ac{ATE} over the first four sequences, demonstrating the slightly bigger overall error of \ac{HF} Odom when compared to \ac{HF} World. 
Yet, the two bottom rows of \Cref{fig:exp_anymal_indoor_eval} highlight the increased smoothness of the odometry estimate; the finite differences of \ac{HF} World (green) cause spikes due to the measurement updates at \ac{LR} rate. Moreover, also \ac{TSIF} (blue) experiences some spikes, particularly in the presence of slip, although fewer and smaller. The trajectory of \ac{HF} Odom (purple) is perfectly smooth, i.e., even smoother than the finite-differenced \ac{GT}.\looseness=-1  

\vspace{-1ex}
\subsection{\anontext{\acs{RACER}}{Vehicle} -- Fast off-road Navigation}
\label{sec:exp_racer}
\begin{figure}[t]
\centering
\includegraphics[width=\columnwidth]{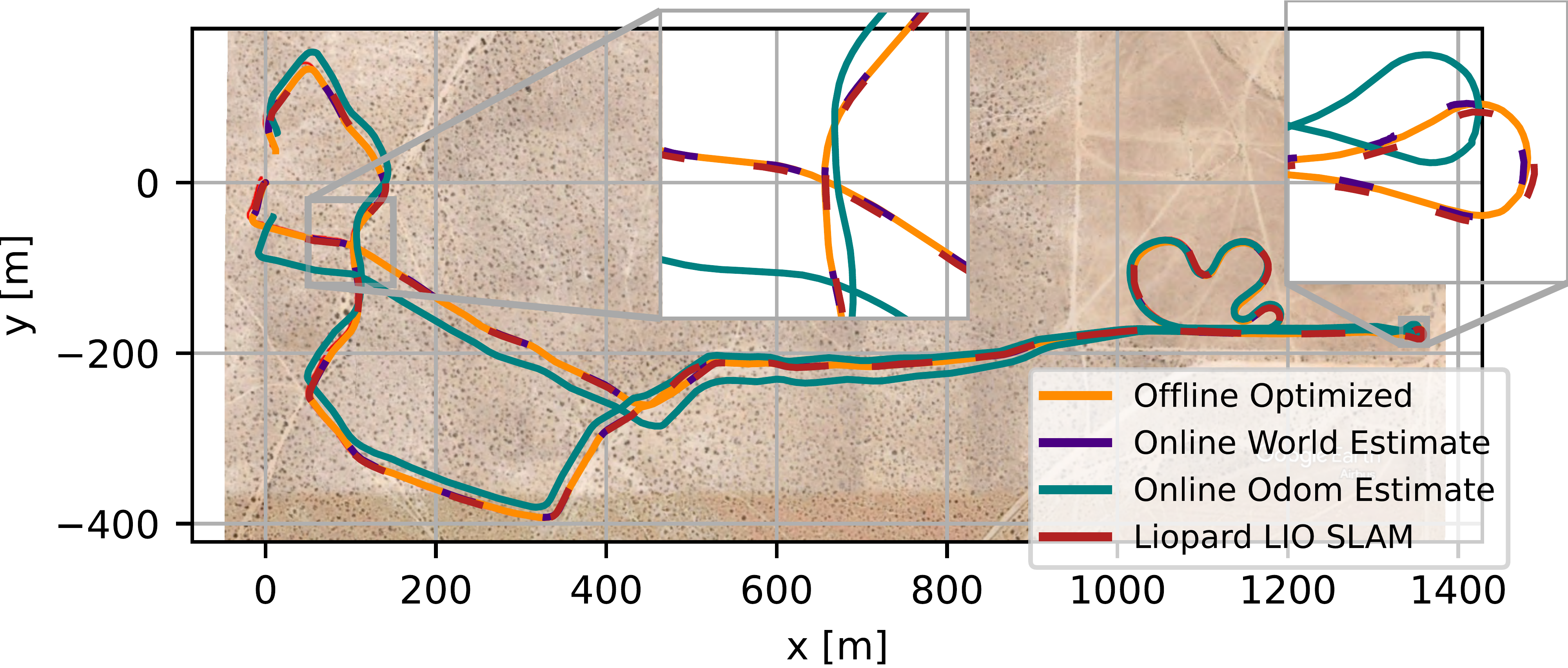}
\centering
\caption{Full run in \protect\anontext{the \textit{\textbf{Helendale}} desert, USA}{place}, in preparation of the \anontext{\ac{RACER}}{anonymous} challenge. The fusion is performed using \ac{IMU}, \ac{GNSS}, \ac{LR}, wheel encoders, and \ac{RADAR}.\looseness=-1}
\label{fig:exp_racer_driven_path}
\vspace{-3ex}
\end{figure}

The customized Polaris RZR S4 1000 Turbo is a fast off-road vehicle with three \acp{LiDAR}, a \si{\milli\meter}-wave \ac{RADAR}, a single \ac{GNSS} antenna and wheel encoder. In the investigated dataset, it traverses \SI{4.1}{\kilo\meter} over uneven terrain at high speeds (up to \SI{9.66}{\meter\per\second}). It requires fast and smooth velocity estimates for the planner and tracking controller. This scenario is challenging, as the traversed environments are desert-like with few geometrical features and high wheel spin.
\ac{HF} has been integrated on the \anontext{\textit{NASA JPL} vehicle}{vehicle} and used as the primary estimator as part of the \anontext{\ac{RACER}}{anonymous} challenge, e.g., reported in~\anontext{\cite{frey2024roadrunner}}{anonymous reference}. The vehicle is shown in the bottom left of \Cref{fig:exp_robotic_platforms}.
Nissov \textit{et al.} used an early version of~\cite{nubert2022graph} to integrate \ac{RADAR} by including the full velocity vector~\cite{nissov2024robust}, similar to \ac{HF}'s implementation in \Cref{equ:method_linear_velocity_factor}.
Because~\cite{nissov2024robust} conducts similar evaluations, we keep the validation short. Still, we present some results as both other works based their code on~\cite{nubert2022graph} instead of the \ac{HF} formulation here.
The same data from the \anontext{\textit{Helendale} desert, USA}{desert place} is used as in~\cite{nissov2024robust}, but with the full \ac{HF} formulation, including frame alignment and smooth odometry. 
Moreover, the single-wheel encoder is newly integrated using the local velocity factor of \Cref{equ:method_linear_velocity_factor}.
\ac{HF} fuses \textit{five} different sensor modalities: \ac{IMU}, \ac{GNSS}, \ac{LiDAR}, \ac{RADAR}, and the wheel encoder.
The estimated and offline-optimized trajectories are shown in \Cref{fig:exp_racer_driven_path}.
As also shown in~\cite{nissov2024robust}, the addition of \ac{RADAR} and/or wheel encoders helps during highly dynamic driving in featureless areas.\looseness=-1

\pdfpxdimen=\dimexpr 1 in/72\relax
\pdfpxdimen=\dimexpr 1 in/72\relax
\begin{figure}[b]
\vspace{-3ex}
\includegraphics[width=\columnwidth]{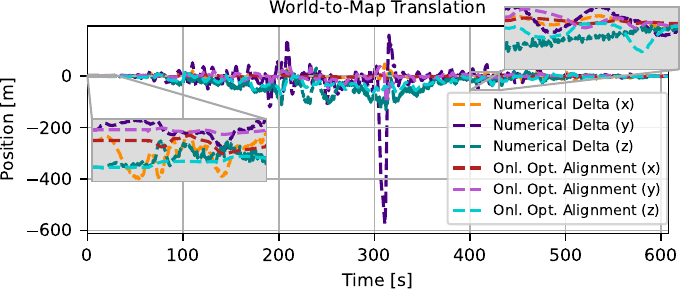}
\centering
\caption{Visualization of $\T_{\World \Liopard}$ at full \ac{IMU} rate through \textit{\textbf{i)}} simple numerical delta computation and \textit{\textbf{ii)}} the optimized estimate from \ac{HF}. The latter is smoother and removes the jumps due to synchronization and numeric stability. The correctness of the large delta values was manually verified.\looseness=-1}
\label{fig:exp_racer_localization_manager_vs_optimized}
\end{figure}

\paragraph*{\acs{HF} as Localization Manager}
In many robotic systems, the robot state is relevant w.r.t. not only $\World$ but also other frames, e.g., for navigation purposes, the map frame of the \ac{SLAM} solution $\Liopard$ builds the robot-centric local map. 
Thus, practitioners often use \textit{localization managers}, which look up the robot state in each frame, synchronize them, and add a new node to the transformation manager by simply computing their delta, e.g., $\T_{\World \Liopard} = \T_{\World \Imu} \T_{\Liopard \Imu}^{-1}$.
This approach can be problematic, as these two transformations are not synchronized, causing noise or jumps in $\T_{\World \Liopard}$, in particular, if $\T_{\World \Imu}$ and/or $\T_{\Liopard \Imu}$ are far from the origin.
In contrast, \ac{HF} provides these transformations synchronized and smoothed, as they are jointly optimized with the robot state.
\Cref{fig:exp_racer_localization_manager_vs_optimized} compares the online-optimized version of $\T_{\Liopard \Imu}$ at full rate to the manual delta computation at synchronized timestamps.
Looking up the state with the synchronized \ac{HF} leads to significantly smoother estimates and less noise.
\replace{}{This comparison is particularly relevant for the Vehicle dataset, as the long distances traveled amplify synchronization artifacts, making this the key result to highlight over standard trajectory-error baselines.}\looseness=-1

\pdfpxdimen=\dimexpr 1 in/72\relax
\begin{figure}[t]
\includegraphics[width=\columnwidth, clip, trim=0cm 4.0cm 0cm 0cm]{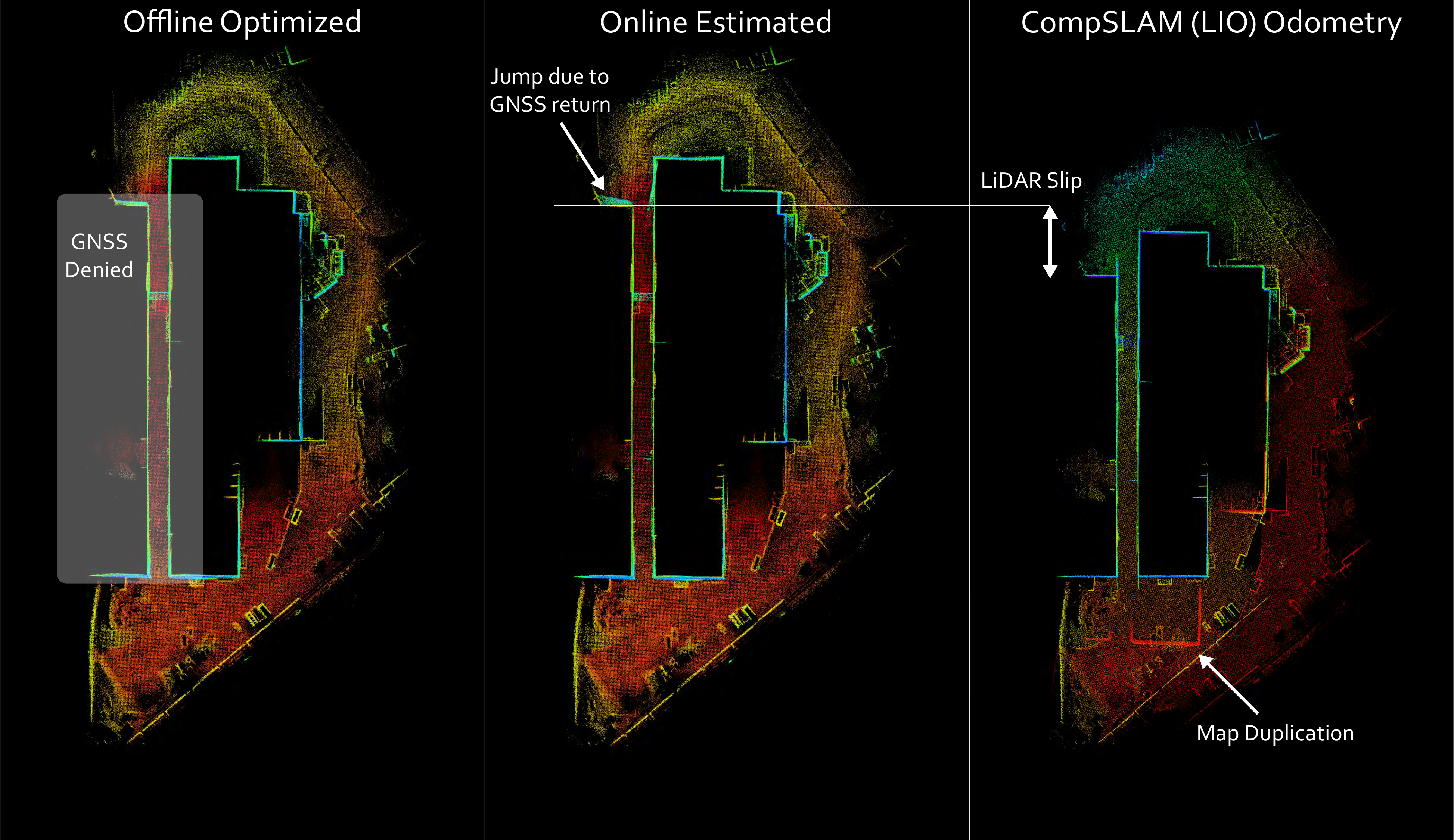}
\centering
\caption{Created maps for the \protect\anontext{\textbf{Oberglatt}, Switzerland}{construction place} experiment using the offline, online and raw \ac{LR} estimates. The drift occurs due to \ac{LiDAR} degeneracy.}
\label{fig:exp_heap_oberglatt}
\vspace{-3ex}
\end{figure}

\subsection{\protect\anontext{\acs{HEAP}}{Excavator}: Long-Term Global \& Local Estimation}
\label{sec:exp_heap}
The \anontext{\ac{HEAP}~\cite{heap2021}}{excavator name} hydraulic walking excavator has an \ac{IMU}, two GNSS antennas, an Ouster OS0-128 LiDAR sensor, and a rotary encoder between the cabin and the chassis. Two \ac{GNSS} antennas make yaw fully observable, even during extended static operation, as \anontext{\ac{HEAP}}{excavator name} operates for hours or days. 
The robot must work even without \ac{GNSS} while maintaining global pose accuracy to perform construction tasks in the real world. This goal is challenging, as \ac{GNSS} signal can often disappear, usually leading to drifting estimates due to the lack of a global reference.
This motivated~\cite{nubert2022graph}, where a dual factor-graph formulation was used to switch the context depending on \ac{GNSS} availability. This section demonstrates that this can be achieved using a single generic graph with the \ac{HF} formulation.\looseness=-1

\subsubsection{Simultaneous Geometric Degeneracy and GNSS Dropout}
\begin{figure}[t]
\includegraphics[width=\columnwidth]{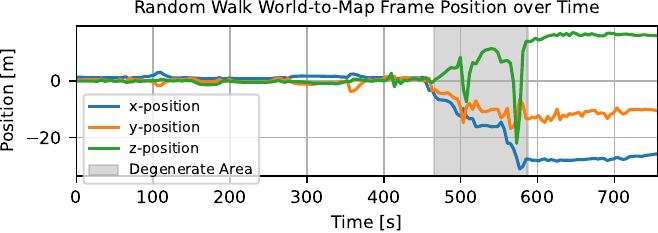}
\centering
\caption{Estimated translation drift for the \protect\anontext{\textbf{Oberglatt}, Switzerland}{construction} experiment. As expected, the framework estimates larger drift during geometric degeneracy.\looseness=-1}
\label{fig:exp_heap_oberglatt_drift}
\vspace{-3ex}
\end{figure}

The investigated dataset constitutes a typical industrial construction site task in \anontext{Oberglatt, Switzerland}{construction place}. 
The dataset contains static and driving parts. At the same time, the corridor between the two industrial buildings (cf. \Cref{fig:exp_heap_oberglatt}) is both \ac{GNSS}-denied and geometrically degenerate, as also shown in~\cite{tuna2024informed}, affecting the \ac{LiDAR} scan-to-map registration of CompSLAM~\cite{khattak2020complementary}.
By fusing the scan-to-map registration as an absolute pose and including the feature-based Coin-LIO~\cite{pfreundschuh2024coin} output of the dense image-like point cloud of the \ac{LiDAR}, the entire mission can be conducted despite the loss of two core measurements (\ac{GNSS} in $\World$ and \ac{LiDAR} path in $\Compslam$), as shown in the maps in \Cref{fig:exp_heap_oberglatt}.
Thus, three reference frames are present: $\World$, $\Compslam$, and $\Coinlio$.
The location of the drifting $\Compslam$-frame over time is visualized in \Cref{fig:exp_heap_oberglatt_drift}, highlighting the effectiveness of drift modeling during degeneracy.\looseness=-1

\pdfpxdimen=\dimexpr 1 in/72\relax

\subsubsection{Alignment with Local Keyframes}
\label{sec:exp_local_keyframes}
\begin{figure}[b]
\vspace{-3ex}
\includegraphics[width=0.41\columnwidth, clip, trim=0 0 0 0.39\columnwidth]{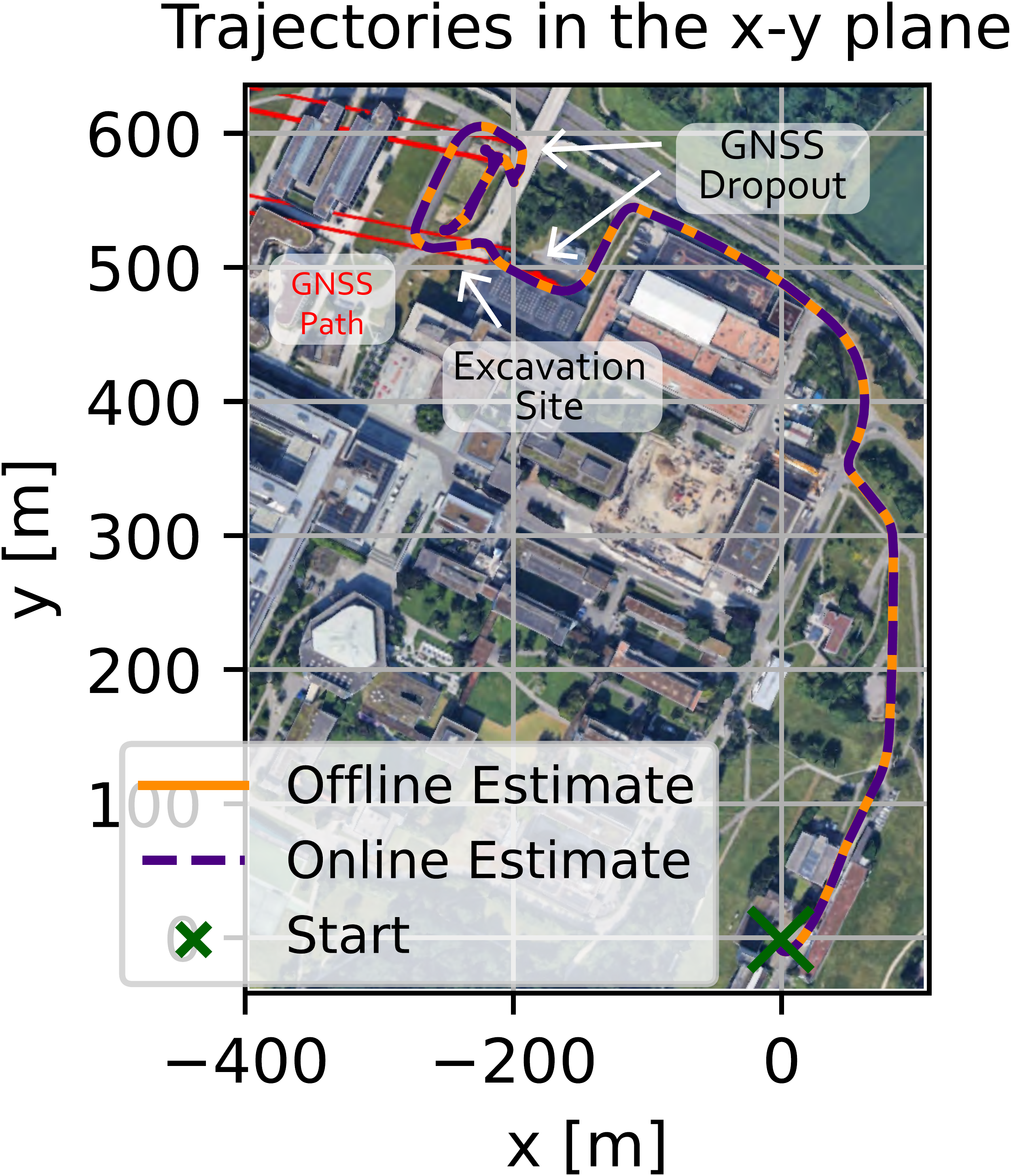}
\includegraphics[width=0.575\columnwidth]{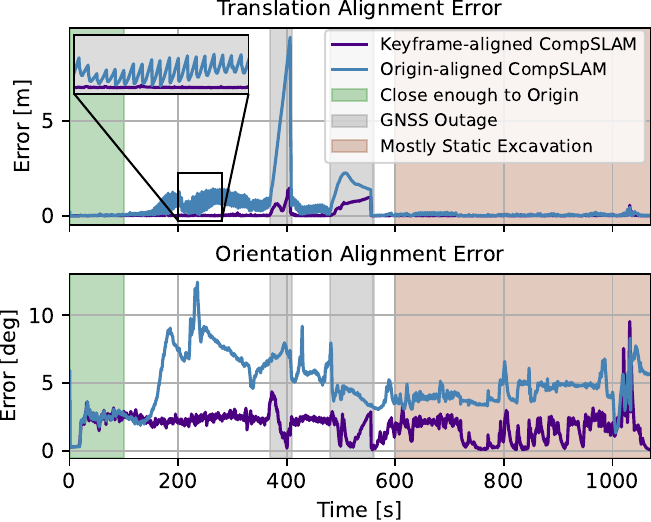}
\centering
\caption{Translation/orientation alignment error for the \protect\anontext{\textbf{ETH Zurich campus}}{urban place} for \textit{i)} alignment around the global $\World$ origin and \textit{ii)} around the local keyframe. When far away from the origin, the alignment error is significantly larger for case \textit{ii)}.\looseness=-1}
\label{fig:exp_heap_origin_alignment}
\end{figure}

Another key component for aligning long trajectories is the alignment around local keyframes, as introduced in \Cref{sec:local_keyframe_alignment}. 
\Cref{fig:exp_heap_origin_alignment} showcases the result on a second dataset collected on \anontext{ETH Zurich campus}{urban place} with a more extensive total distance.  Here, the reference-frame alignment is performed \textit{including} drift modeling, but \textit{with and without} alignment around the $\World$-frame origin instead of the local keyframe.
\pdfpxdimen=\dimexpr 1 in/72\relax
As visible in the figure, the online-aligned \ac{LR} does not correctly align with the estimated robot position, eventually leading to oscillations after around \SI{150}{\second} due to linearly increasing sensitivity (cf. \Cref{sec:local_keyframe_alignment}).
\looseness=-1


%% file: sections/7-Discussion.tex
\subsection{Analysis}
The real-world experiments above show that \ac{HF} benefits diverse robotic applications. Aligning reference frames is crucial to simplify the setup and fuse measurements directly, \textit{as is}, without transforming them to a common frame, using dual factors~\cite{nubert2022graph}, or converting them to local measurements.
In practice, performing the alignment around a local keyframe rather than the $\World$ origin reduces the distance-dependent rotational sensitivity (\Cref{sec:exp_local_keyframes}). Most measurement types also require a random-walk model on these keyframes to compensate for drift. As illustrated in \Cref{fig:exp_heap_oberglatt_drift}, this step makes state estimation robust to geometric degeneracies without active prevention.
Having access to an offline-optimized trajectory at full \ac{IMU} rate is useful for assessing the quality and smoothness of the real-time estimates. Likewise, the synchronized reference-frame relationships from the optimization are beneficial and remove the need for a localization manager.\looseness=-1

\vspace{-2ex}
\subsection{Limitations \& Lessons Learned}
To ensure good practical performance, the graph’s initialization must be handled carefully, especially \replace{online}{during online operation}. Without an initial state prior, the problem can become ill-posed. Assuming the robot is static at startup is helpful but often too strict when initializing in motion. A practical compromise is to constrain the initial state with high uncertainty so it can quickly converge.
If initial values are far from the true state, the optimizer may fall into an incorrect local minimum. \replace{Such errors are critical w}{W}hen estimating the robot’s \replace{position}{state} in the \ac{GNSS} world frame using a single antenna\replace{}{, such errors are hard to recover from}: since yaw is initially unknown, the system may need to wait until sufficient motion provides observability. This strategy is used in the outdoor ANYmal experiments to initialize heading via \replace{}{explicit} Umeyama alignment.
Real robots also face computational limits, and \Cref{tab:exp_anymal_hike_computational_complexity} shows the tradeoff between allocating more states (higher accuracy) and fewer states (lower computation). 
Finally, tuning the framework can be challenging for new users, as greater flexibility introduces more parameters, e.g., drift levels, initial uncertainty of alignment variables. Nevertheless, the default settings work well for most practical scenarios.
\looseness=-1

%% file: sections/8-Conclusions.tex
This work presented \acl{HF}, a general and flexible sensor-fusion framework designed to work well on a wide set of platforms and systems. Unlike earlier works, \acs{HF} allows the user to add measurements without heuristic preprocessing or alignment by explicitly considering the corresponding reference $\Reference_i$ and sensor frame $\Sensor_i$ as part of a single optimization.
The presented large-scale real-world deployments and dynamic indoor applications demonstrate the flexibility and broad applicability of the problem formulation. These experiments highlight the importance of the contributions from \Cref{sec:intro} across different platforms and use cases.
The corresponding framework, \acl{HF}, is the new default state estimator on multiple platforms and has been released to the community as a well-documented open-source framework with multiple examples.
Future work might include introducing a random process for modeling the evolution of global states, \ac{CT} estimation methods into the backend, and/or an adaptive scheme for state creation based on the robot dynamics and/or the anticipated measurement arrival time.
Moreover, the injection of \ac{ML}, either as a learned probabilistic prior in ill-posed situations or to filter measurement outliers, holds significant promise.\looseness=-1

%% file: biographies.tex
\begin{IEEEbiography}
[{\includegraphics[trim={236 0 236 0},clip,width=1in,height=1.25in,keepaspectratio]{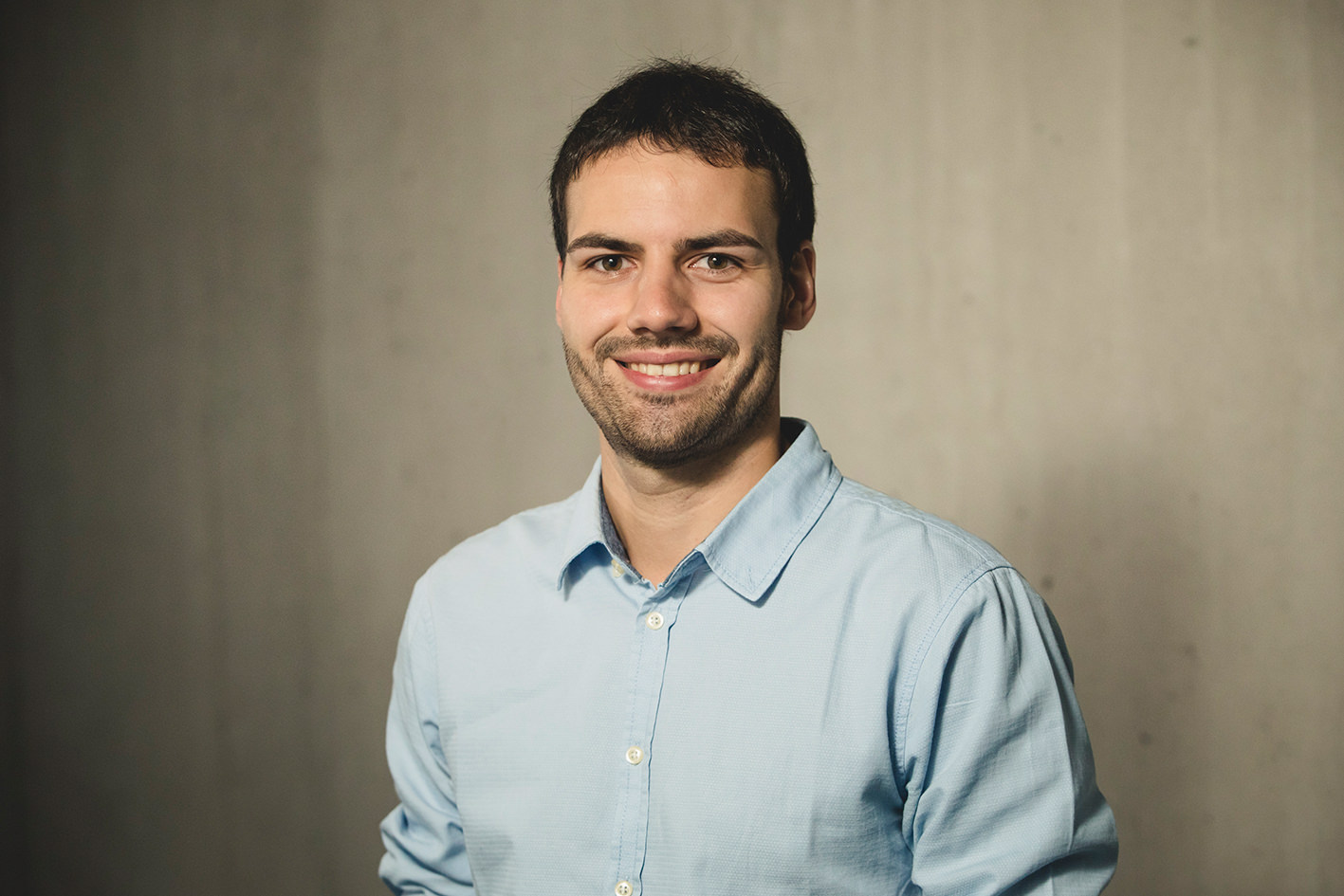}}]%
{Julian Nubert}~(Student Member, IEEE)~is a PhD student in the Robotic Systems Lab at ETH Zurich. He received his M.Sc. in Robotics, Systems \& Control in 2020 from ETH Zurich. He is affiliated with the Max Planck Institute through the MPI ETH Center for Learning Systems. His research interests lie in robust robot perception and how it can be used to deploy mobile robotic systems. Julian received the ETH silver medal and was awarded the Willi Studer Prize for his accomplishments during his master's studies.\looseness=-1
\end{IEEEbiography}

\begin{IEEEbiography}
[{\includegraphics[trim={18 0 18 0},clip,width=1in,height=1.25in,keepaspectratio]{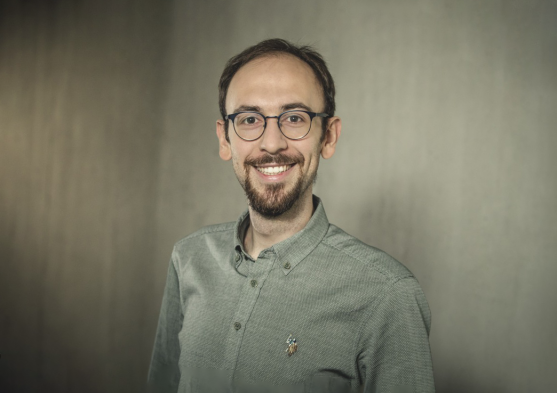}}]%
{Turcan Tuna}~(Student Member, IEEE)~is a PhD student in the Robotic Systems Lab at ETH Zurich. He received his M.Sc. in Robotics, Systems \& Control in 2022 from ETH Zurich. Previously, he completed a double major, B.Sc in Mechanical Engineering and Control \& Automation Engineering, at Istanbul Technical University. He graduated with distinction from both of his B.Sc majors. His research interests include developing and deploying robust localization, perception, and mapping frameworks on robotic systems.\looseness=-1
\end{IEEEbiography}

\begin{IEEEbiography}
[{\includegraphics[trim={236 0 236 0},clip,width=1in,height=1.25in,keepaspectratio]{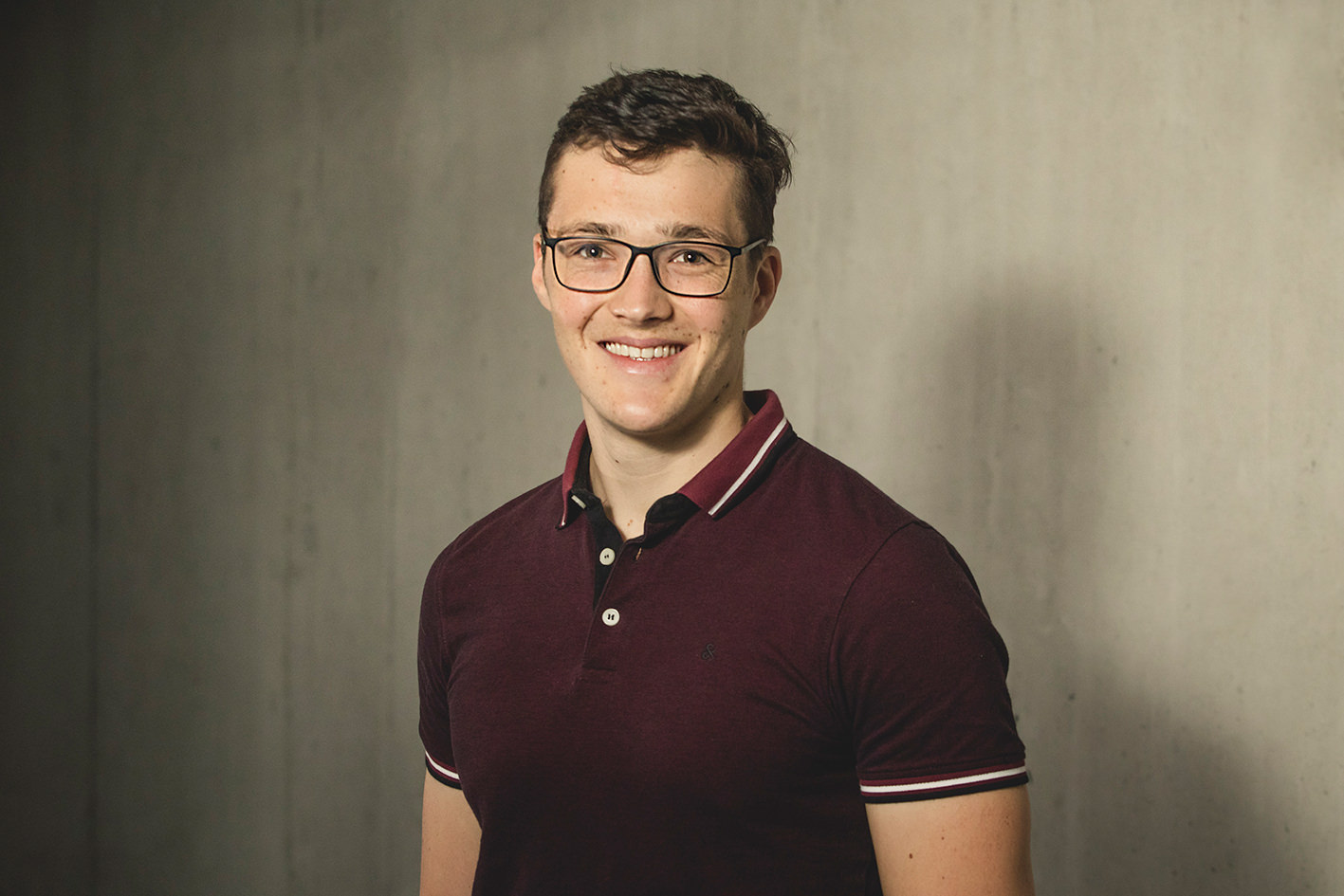}}]%
{Jonas Frey}~(Student Member, IEEE)~is a PhD student in the Robotic Systems Lab at ETH Zurich. He received his M.Sc. in Robotics, Systems \& Control in 2021 from ETH Zurich. He is affiliated with the Max Planck Institute through the MPI ETH Center for Learning Systems. His research interests lie in perception, navigation, locomotion, and how it can be used to deploy mobile robotic systems.\looseness=-1
\end{IEEEbiography} 

\begin{IEEEbiography}
[{\includegraphics[trim={236 0 236 0},clip,width=1in,height=1.25in,keepaspectratio]{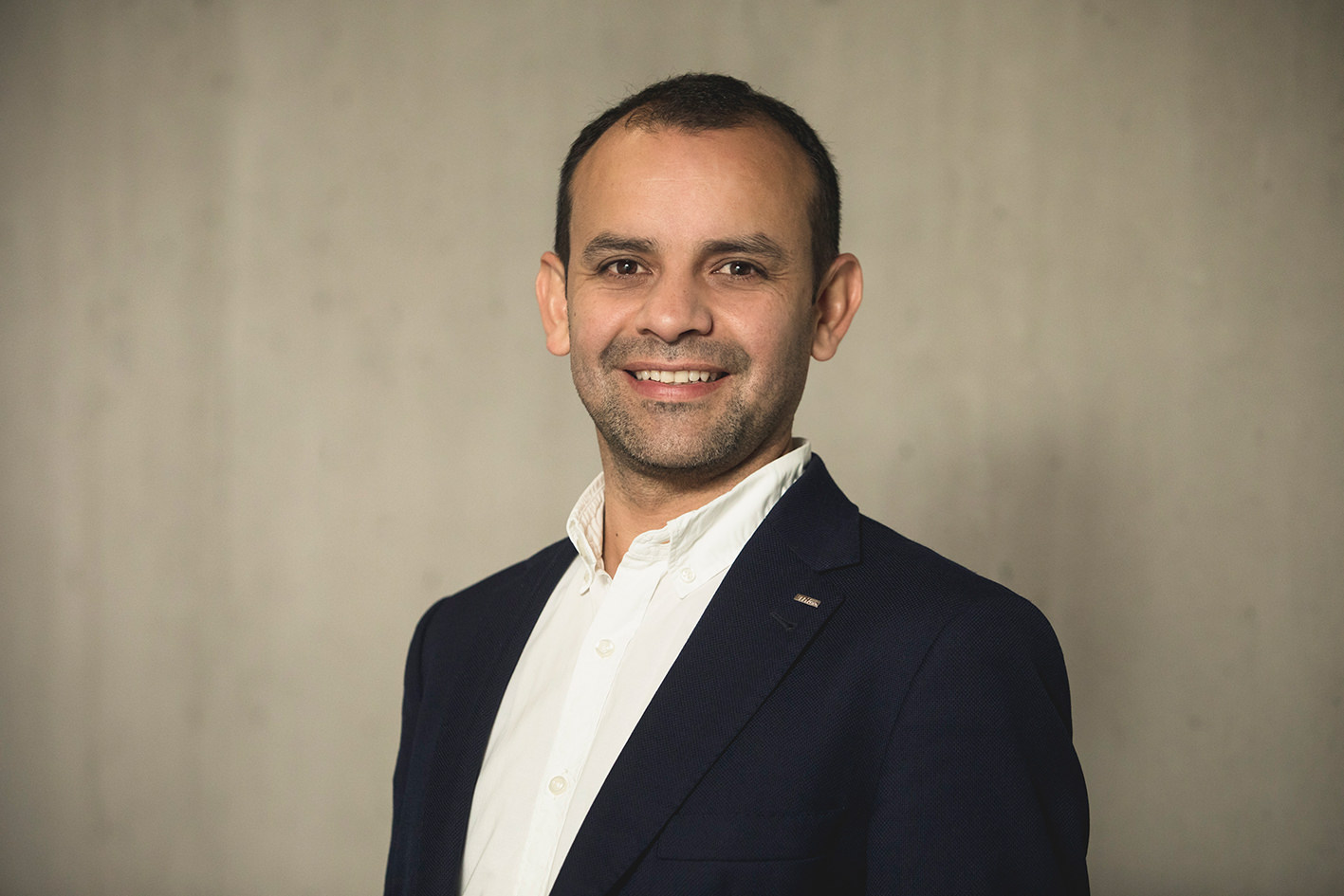}}]%
{Cesar Cadena}~(Member, IEEE)~received his PhD (2011) in computer science from the University of Zaragoza, Spain. He is a Senior Scientist at ETH Zurich. His research interests are in providing machines the ability to understand and interact with our ever-changing world. His work covers robotic scene understanding, both geometry and semantics, covering semantic mapping, data association, and place recognition tasks, simultaneous localization and mapping, as well as persistent mapping in dynamic environments and robot navigation.\looseness=-1
\end{IEEEbiography}

\begin{IEEEbiography}
[{\includegraphics[trim={130 0 130 0},clip,width=1in,height=1.25in,keepaspectratio]{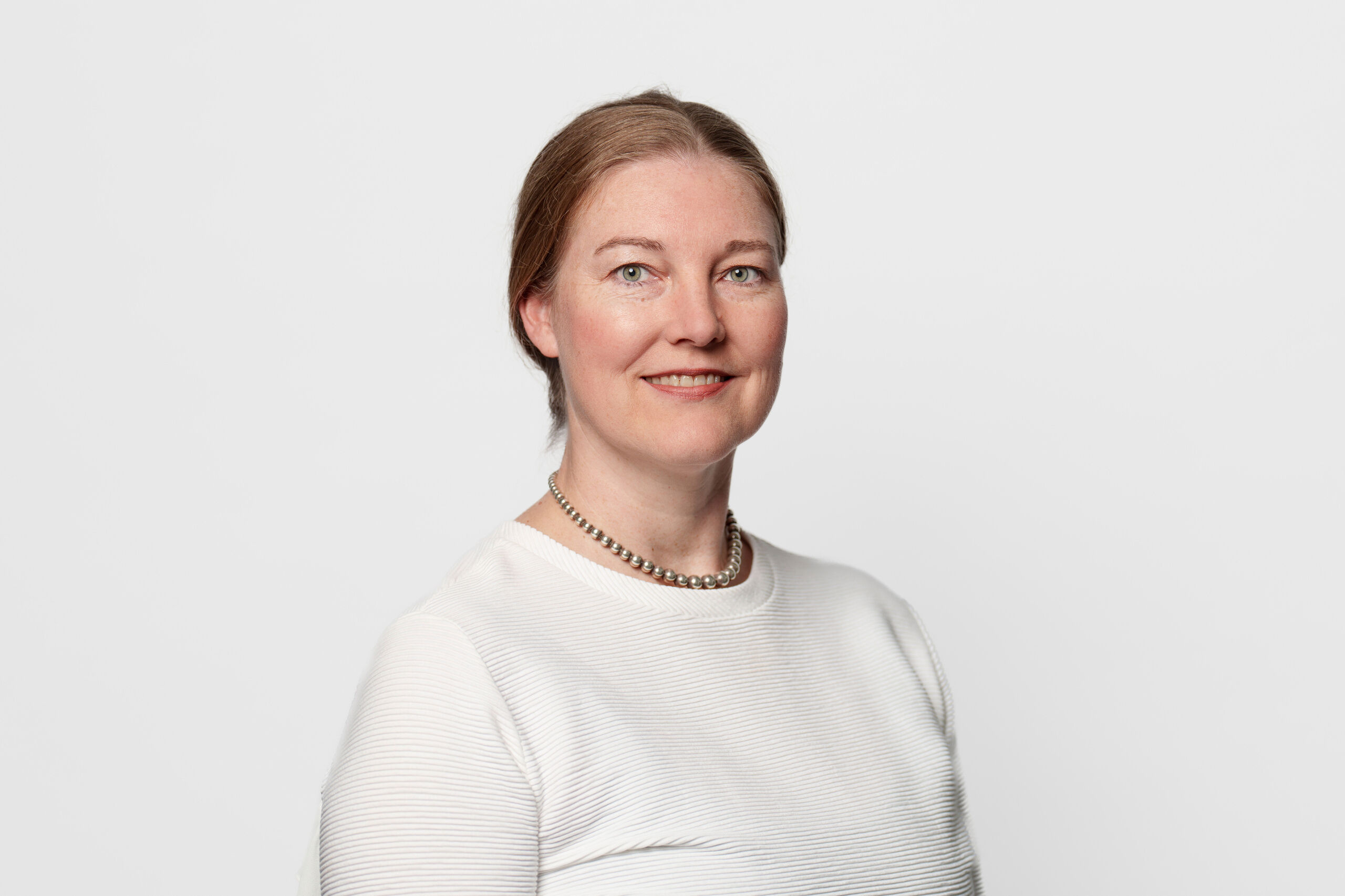}}]%
{Katherine J. Kuchenbecker}~(Fellow, IEEE)~earned B.S., M.S., and Ph.D. degrees in Mechanical Engineering from Stanford University, USA, in 2000, 2002, and 2006, respectively. She was a postdoctoral researcher at Johns Hopkins University and an Assistant and Associate Professor in the GRASP Lab at the University of Pennsylvania from 2007 to 2016. Since 2017, she has been a Director at the Max Planck Institute for Intelligent Systems in Stuttgart, Germany. Her research blends robotics and human-computer interaction, including work in haptics, teleoperation, physical human-robot interaction, tactile sensing, and medical applications. She has been honored with a 2009 NSF CAREER Award, the 2012 IEEE RAS Academic Early Career Award, a 2014 Penn Lindback Award for Distinguished Teaching, and elevation to IEEE Fellow in 2022.\looseness=-1
\end{IEEEbiography}

\begin{IEEEbiography}
[{\includegraphics[trim={236 0 236 0}, clip,width=1in,height=1.25in,keepaspectratio]{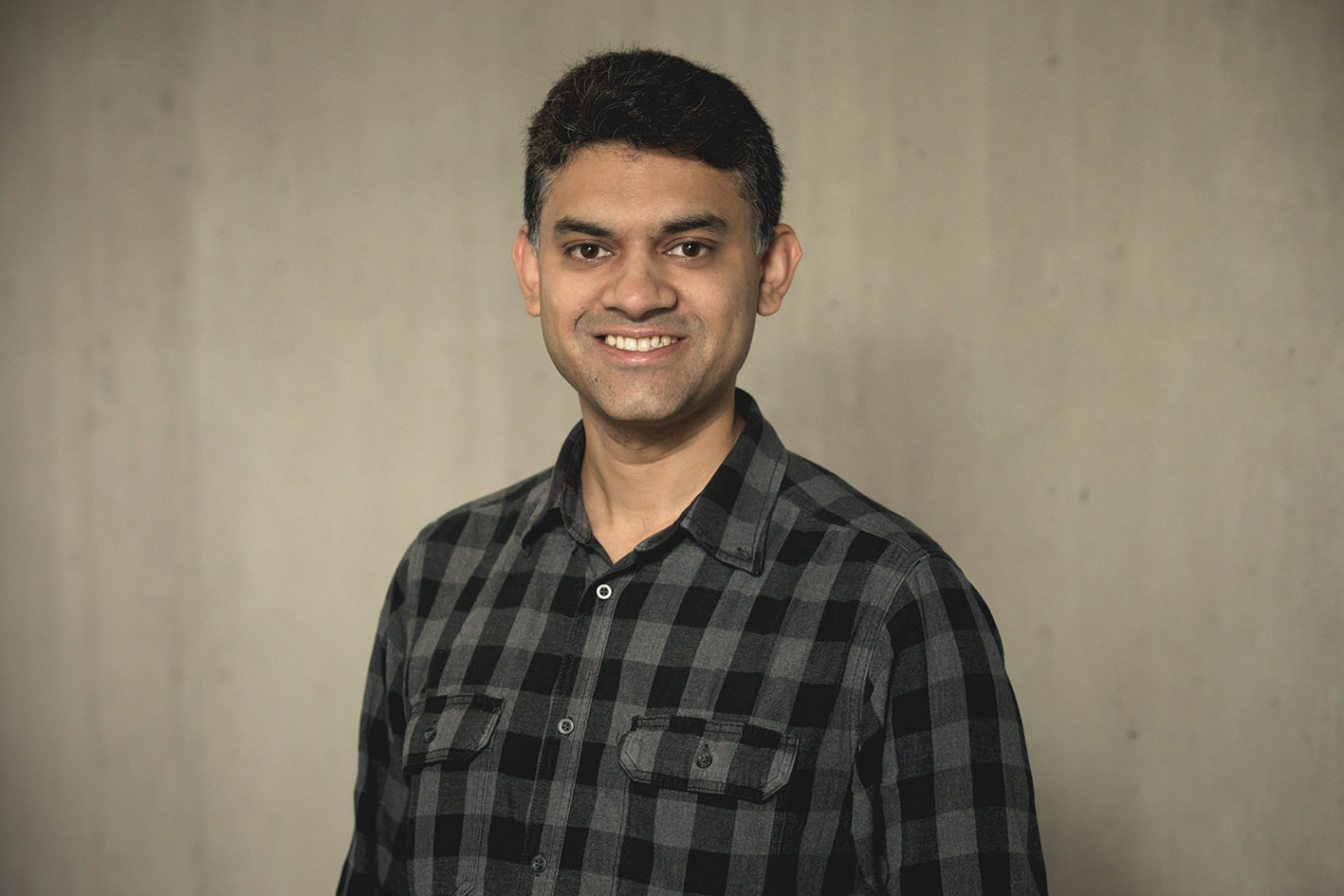}}]%
{Shehryar Khattak}~(Member, IEEE)~ is a Robotics Technologist in the Perception Systems Group at NASA's Jet Propulsion Laboratory (JPL). His work focuses on enabling resilient robot autonomy in complex environments through multi-sensor information fusion. Currently, he is the Principal Investigator (PI) for the Multi-robot Autonomous Intelligent Search and Rescue project at JPL. He previously served as the perception lead for JPL's team in the DARPA RACER project and for Team CERBERUS, winners of the DARPA Subterranean Challenge. Before joining JPL, Shehryar was a postdoctoral researcher at ETH Zurich. He earned his Ph.D. (2019) and M.S. (2017) in Computer Science from the University of Nevada, Reno. Additionally, he holds an M.S. in Aerospace Engineering from KAIST (2012) and a B.S. in Mechanical Engineering from GIKI (2009).\looseness=-1
\end{IEEEbiography}

\begin{IEEEbiography}
[{\includegraphics[trim={236 0 236 0}, clip,width=1in,height=1.25in,keepaspectratio]{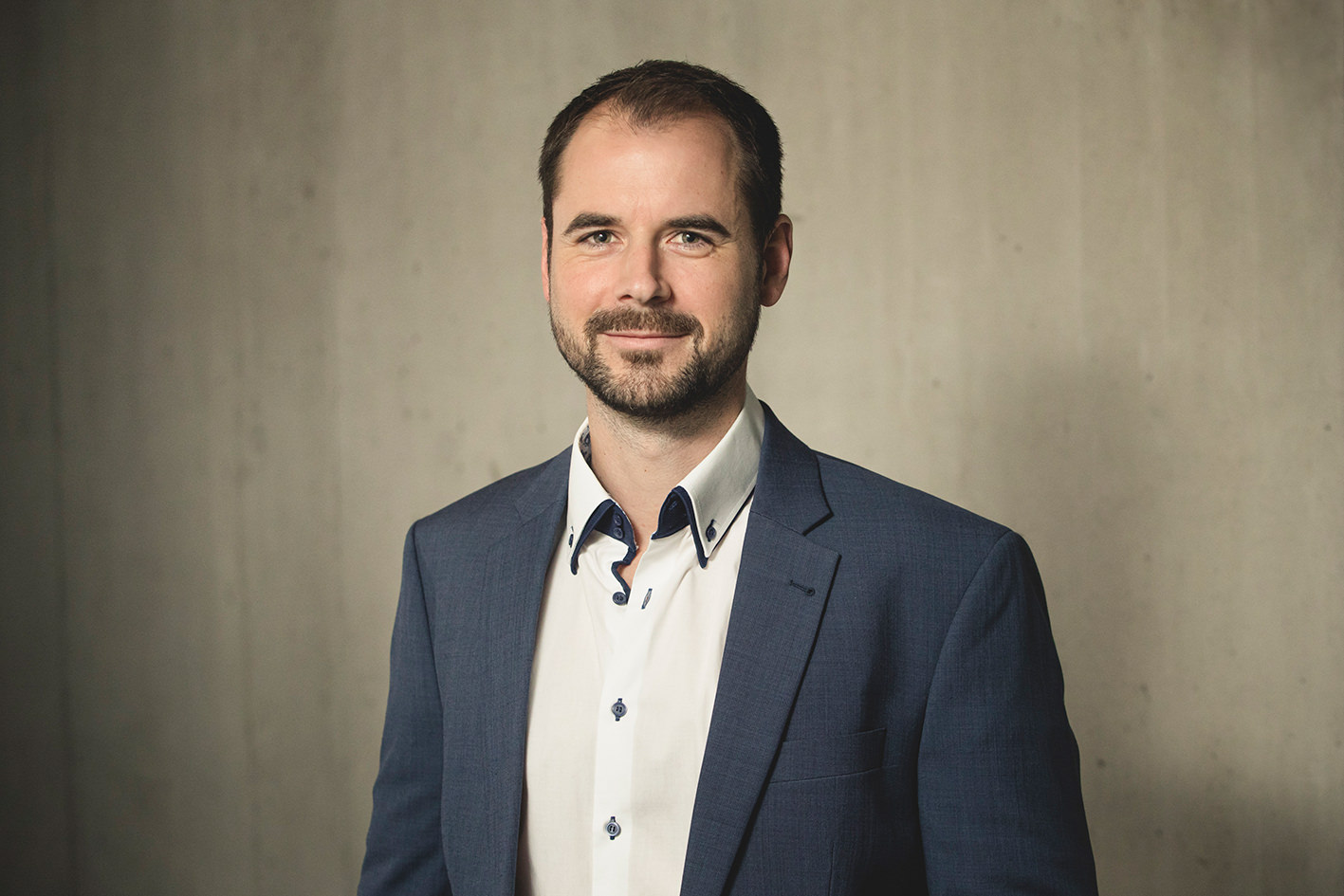}}]%
{Marco Hutter}~(Member, IEEE)~is an Associate Professor for Robotic Systems and Director of the Center for Robotics at ETH Zurich. He received his M.Sc. and Ph.D. from ETH Zurich in 2009 and 2013 in design, actuation, and control of legged robots. His research interests are in the development of novel machines and actuation concepts together with the underlying control, planning, and machine learning algorithms for locomotion and manipulation. Marco is the recipient of an ERC Starting Grant, PI of the NCCRs robotics, automation, and digital fabrication, the DARPA SubT Challenge winner, and a co-founder of several ETH Startups such as ANYbotics and Gravis Robotics. He is also the Director of the Robotics and AI Institute (RAI) Zurich.\looseness=-1
\end{IEEEbiography}